%% file: bong-neurips24-final.tex
\documentclass{article}

\usepackage[final]{neurips_2024}

\usepackage[utf8]{inputenc} 
\usepackage[T1]{fontenc}    
\usepackage{hyperref}       
\usepackage{url}            
\usepackage{booktabs}       
\usepackage{amsfonts}       
\usepackage{nicefrac}       
\usepackage{microtype}      
\usepackage{xcolor}         

\usepackage{xspace}
\usepackage{amsmath}
\usepackage{amssymb}
\usepackage{amsthm}
\usepackage{bm}
\usepackage[capitalize]{cleveref}
\Crefname{algocf}{Algorithm}{Algorithms}
\usepackage{accents}
\usepackage{multirow}
\usepackage{dsfont}
\usepackage{todonotes}
\usepackage{algorithm2e}
\usepackage{wrapfig,booktabs}
\usepackage{enumitem}

\usepackage[export]{adjustbox}
\usepackage{graphicx}
\usepackage{caption}
\usepackage{subcaption}

\theoremstyle{plain}
\newtheorem{theorem}{Theorem}[section]
\newtheorem{proposition}[theorem]{Proposition}

\theoremstyle{definition}

\theoremstyle{remark}

\newcommand{\bong}{\textsc{bong}\xspace}
\newcommand{\Bong}{\textsc{Bong}\xspace}
\newcommand{\mcbong}{\textsc{bong-mc}\xspace}

\newcommand{\lbong}{\textsc{bong-lin}\xspace}

\newcommand{\bog}{\textsc{bog}\xspace}
\newcommand{\Bog}{\textsc{Bog}\xspace}

\newcommand{\blr}{\textsc{blr}\xspace}
\newcommand{\Blr}{\textsc{Blr}\xspace}

\newcommand{\bbb}{\textsc{bbb}\xspace}
\newcommand{\BBB}{\textsc{Bbb}\xspace} 

\newcommand{\fc}{\textsc{fc}\xspace}

\newcommand{\dg}{\textsc{diag}\xspace}
\newcommand{\dlr}{\textsc{dlr}\xspace}

\newcommand{\fcmom}{\textsc{fc{\un}mom}\xspace}
\newcommand{\diagmom}{\textsc{diag{\un}mom}\xspace}

\newcommand{\mc}{\textsc{mc}\xspace}
\newcommand{\ef}{\textsc{ef}\xspace}

\newcommand{\un}{\fontsize{5}{5}\selectfont\textunderscore}

\newcommand{\hessmc}{\textsc{mc-hess}\xspace}
\newcommand{\hesslin}{\textsc{lin-hess}\xspace}
\newcommand{\efmc}{\textsc{mc-ef}\xspace}
\newcommand{\eflin}{\textsc{lin-ef}\xspace}

\newcommand{\mchess}{\hessmc}
\newcommand{\linhess}{\hesslin}
\newcommand{\mcef}{\efmc}
\newcommand{\linef}{\eflin}

\newcommand{\gMC}{\vg^{\textsc{mc}}}
\newcommand{\gL}{\vg^{\textsc{lin}}}
\newcommand{\gLEF}{\vg^{\linef}}

\newcommand{\GMCH}{\vG^{\mchess}}
\newcommand{\GLH}{\vG^{\linhess}}
\newcommand{\GMCEF}{\vG^{\mcef}}
\newcommand{\GLEF}{\vG^{\linef}}

\newcommand{\gradmat}{\hat{\vG}^{(1:\nsample)}}
\newcommand{\gradmatT}{\hat{\vG}^{(1:\nsample)^\trans}}

\newcommand{\SARCOS}{\textsc{sarcos}\xspace}
\newcommand{\MNIST}{\textsc{mnist}\xspace}

\newcommand{\CMEKF}{\textsc{cm-ekf}\xspace}
\newcommand{\VDEKF}{\textsc{vd-ekf}\xspace}
\newcommand{\LOFI}{\textsc{lo-fi}\xspace} 
\newcommand{\VON}{\textsc{von}\xspace}
\newcommand{\VOGN}{\textsc{vogn}\xspace}
\newcommand{\SLANG}{\textsc{slang}\xspace}
\newcommand{\Slang}{\textsc{Slang}\xspace}
\newcommand{\RVGA}{\textsc{rvga}\xspace}

\newcommand{\nparam}{P}
\newcommand{\nsample}{M}
\newcommand{\rank}{R}
\newcommand{\nhutchinson}{N}

\newcommand{\sizeA}{\footnotesize}
\newcommand{\sizeB}{\scriptsize}
\newcommand{\sizeC}{\tiny}

\newcommand{\kpm}[1]{#1}
\newcommand{\mj}[1]{#1}
\newcommand{\rev}[1]{#1}

\input{macros.tex}

\title{Bayesian Online Natural Gradient  (\bong)}

\author{%
  Matt Jones\\
  University of Colorado\\
  \texttt{mcjones@colorado.edu}\\
    \And
  Peter Chang\\
  MIT \\
  \texttt{gyuyoung@mit.edu} \\
  \And
  Kevin Murphy  \\
    Google DeepMind \\
  \texttt{kpmurphy@google.com} 
}

\begin{document}

\maketitle

\input{sections/abstract}
\input{sections/intro}
\input{sections/related}
\input{sections/background}
\input{sections/methods}

\input{sections/experiments}

\input{sections/concl}
\newpage
\begin{ack}
Thanks to Gerardo Duràn-Martìn and Alex Shestopaloff for helpful input. MJ was supported by NSF grant 2020-906.

\end{ack}

\bibliography{refs}
\bibliographystyle{plainnat}

\newpage
\appendix

\input{sections/pseudocode}
\input{sections/appx-experiments}
\input{sections/proof}
\input{sections/MD}
\input{sections/derivations}

\end{document}

%% file: macros.tex
\newcommand{\KLpq}[2]{D_{\mathbb{KL}}\!\left({#1}| {#2}\right)}

\newcommand{\eat}[1]{}

\newcommand{\nparams}{P}
\newcommand{\nout}{C}

\newcommand{\niter}{I}

\newcommand{\diag}{\mathrm{diag}}
\newcommand{\Diag}{\mathrm{Diag}}

\newcommand{\data}{\mathcal{D}}
\newcommand{\real}{\sR}

\newcommand{\loss}{\mathcal{L}}

\newcommand{\T}{\intercal}
\newcommand{\trans}{\T}

\newcommand{\myvec}[1]{\mathbf{#1}}
\newcommand{\myvecsym}[1]{\boldsymbol{#1}}

\makeatletter
\newcommand{\algorithmfootnote}[2][\footnotesize]{%
  \let\old@algocf@finish\@algocf@finish
  \def\@algocf@finish{\old@algocf@finish
    \leavevmode\rlap{\begin{minipage}{\linewidth}
    #1#2
    \end{minipage}}%
  }%
}
\makeatother

\def\1{\bm{1}}

\newcommand{\gauss}{\mathcal{N}}

\newcommand{\const}{\mathrm{const}}

\newcommand{\expect}[1]{\mathbb{E}\left[{#1}\right]} 
\newcommand{\expectQ}[2]{\mathbb{E}_{{#2}}\!\left[ {#1} \right]} 

\newcommand{\var}[1]{\mathbb{V}\left[ {#1}\right]}

\newcommand{\vchi}{\myvecsym{\chi}}

\newcommand{\vell}{\myvecsym{\ell}}
\newcommand{\veta}{\myvecsym{\eta}}

\newcommand{\vmu}{\myvecsym{\mu}}

\newcommand{\vlambda}{\myvecsym{\lambda}}
\newcommand{\vLambda}{\myvecsym{\Lambda}}

\newcommand{\vpsi}{\myvecsym{\psi}}

\newcommand{\vrho}{\myvecsym{\rho}}
\newcommand{\vtheta}{\myvecsym{\theta}}

\newcommand{\vsigma}{\myvecsym{\sigma}}
\newcommand{\vSigma}{\myvecsym{\Sigma}}

\newcommand{\vUpsilon}{\myvecsym{\Upsilon}}

\newcommand{\vb}{\myvec{b}}

\newcommand{\vf}{\myvec{f}}
\newcommand{\vg}{\myvec{g}}

\newcommand{\vv}{\myvec{v}}

\newcommand{\vx}{\myvec{x}}

\newcommand{\vy}{\myvec{y}}

\def\vtheta{{\bm{\theta}}}

\def\vb{{\bm{b}}}

\def\vf{{\bm{f}}}
\def\vg{{\bm{g}}}

\def\vv{{\bm{v}}}

\def\vx{{\bm{x}}}
\def\vy{{\bm{y}}}

\newcommand{\vA}{\myvec{A}}
\newcommand{\vB}{\myvec{B}}

\newcommand{\vF}{\myvec{F}}
\newcommand{\vG}{\myvec{G}}
\newcommand{\vH}{\myvec{H}}
\newcommand{\vI}{\myvec{I}}

\newcommand{\vK}{\myvec{K}}

\newcommand{\vQ}{\myvec{Q}}
\newcommand{\vR}{\myvec{R}}
\newcommand{\vS}{\myvec{S}}

\newcommand{\vU}{\myvec{U}}
\newcommand{\vV}{\myvec{V}}
\newcommand{\vW}{\myvec{W}}

\def\sR{{\mathbb{R}}}

%% file: sections/abstract.tex
\begin{abstract}
  We propose a novel approach to  sequential Bayesian inference based on variational Bayes (VB).
  The key insight is that,
  in the online setting,
  we do not need to add the KL term to regularize to the prior (which comes from the posterior at the previous timestep);
  instead we can optimize just the expected log-likelihood,
  performing a single step of natural gradient descent
  starting at the prior predictive.
  We prove this method
 recovers exact Bayesian inference 
  if the model is conjugate.
\rev{We also  show how to compute an
 efficient deterministic 
 approximation to the VB objective,
  as well as our simplified objective,
 when the variational distribution is
 Gaussian or a sub-family, including the case of
 a diagonal plus low-rank
precision matrix.}
We show empirically that our
method outperforms other online VB methods
  in the  non-conjugate setting,
  such as online learning for neural networks,
  especially when controlling for computational costs.
\end{abstract}

%% file: sections/intro.tex
\section{Introduction}
\label{sec:intro}

Bayesian methods for neural network (NN) training aim to minimize the
Kullback-Leibler divergence between true and estimated posterior distributions.
This is equivalent to minimizing the variational loss (or negative
ELBO)
\begin{equation}
    \loss(\vpsi) = -\expectQ{\log p(\data\vert\vtheta)}{\vtheta\sim q_{\vpsi}} 
    + \KLpq{q_{\vpsi}}{p_{0}}
    \label{eq:nELBO}
\end{equation}
Here $\vtheta$ are the network parameters, 
$\vpsi$ are the variational parameters of the approximate posterior 
$q_{\vpsi}(\vtheta)$,
$\data$ is the training dataset, 
and $p_{0}(\vtheta)$ is the prior. 
The two terms in the variational loss correspond to data fit and regularization to the prior, 
the latter being analogous to a regularizer 
$r(\vtheta) = -\log p_{0}(\vtheta)$ 
in traditional point estimation methods like SGD.

An important set of approaches learns the variational parameters by gradient descent on $\loss(\vpsi)$ \citep{BBB}. 
More recently Khan and colleagues \citep{VON,BLR,shen2024IVON} have proposed using the natural gradient
$\vF_{\vpsi}^{-1} \nabla_{\vpsi} \loss (\vpsi)$
where $\vF_{\vpsi}$ is the Fisher information matrix of the variational family evaluated at $q_{\vpsi}$. 
Natural gradient descent (NGD) is often more efficient than vanilla GD because it accounts for the intrinsic geometry of the variational family \citep{Amari98}. 
\cite{BLR} call this approach the "Bayesian Learning Rule" or BLR. Using various choices for the variational distribution, 
generalized losses replacing negative log-likelihood,
and other approximations, they reproduce many standard optimization methods such as Adam,
and derive new ones.

We study Bayesian NN optimization in online learning, where the data are observed in sequence, 
$\data_{t} = \{ (\vx_{k},\vy_{k})_{k=1}^{t} \}$,
and the algorithm maintains an approximate posterior 
$q_{\vpsi_{t}}(\vtheta_{t}) \approx  p(\vtheta_{t}\vert\data_{t})$,
which it updates at each step.
Fast updates (in terms of both computational speed and statistical efficiency) are critical for many online learning applications
\citep{Zhang2024CL}.
To allow for nonstationarity in the datastream,
we include a time index on $\vtheta_{t}$,
to represent that the parameters may change over time,
as is standard for  approaches based on state-space models and the extended Kalman filter
(see e.g., \citep{Sarkka23}). 
The belief state is updated recursively using the prior 
$q_{\vpsi_{t\vert t-1}}$
derived from the previous time step so that the variational loss becomes
\begin{equation}
    \loss(\vpsi_{t}) = 
    -\expectQ
        {\log p(\vy_{t}\vert\vx_{t},\vtheta_{t})}
        {\vtheta_{t}\sim q_{\vpsi_{t}}}
    + \KLpq
        {q_{\vpsi_{t}}}
        {q_{\vpsi_{t\vert t-1}}}
    \label{eq:online-variational-loss}
\end{equation}
One option for this online learning problem is to apply
NGD on $\loss(\vpsi_t)$
at each time step, iterating until $\vpsi_{t}$ converges before consuming the next observation. 
Our first contribution is a proposal for skipping this inner loop by 
(a) performing a single natural gradient step with unit learning rate and 
(b) omitting the $D_{\mathbb{KL}}$ term in \cref{eq:online-variational-loss} so that learning is based only on expected loglikelihood:
\begin{equation}
    \vpsi_{t} =
    \vpsi_{t\vert t-1}
    + \vF_{\vpsi_{t|t-1}}^{-1} 
    \nabla_{\vpsi_{t|t-1}} 
    \expectQ
        {\log p\left(\vy_{t}\vert\vx_{t},\vtheta_{t}\right)}
        {\vtheta_{t}\sim q_{\vpsi_{t\vert t-1}}}
    \label{eq:bong}
\end{equation}
These two modifications work together: instead of regularizing toward the prior explicitly using 
$\KLpq{q_{\vpsi_{t}}}{q_{\vpsi_{t\vert t-1}}}$,
we do so implicitly by using $\vpsi_{t\vert t-1}$ as the starting point of our single natural gradient step. 
This may appear as a heuristic but we prove in \cref{thm:exact-when-conjugate} that it yields exact Bayesian inference when 
$q_{\vpsi}$ and $p\left(\vy\vert\vx,\vtheta\right)$ are conjugate 
and $q_{\vpsi}$ is an exponential family with natural parameter $\vpsi$. 
Thus our proposed update can be viewed as a relaxation of the Bayesian update to the non-conjugate variational case. 
\rev{As is common in work on variational inference, we view the result for the conjugate case as a motivating foundation that ensures our method is exact in certain simple settings.
The experiments reported in \cref{sec:experiments} and \cref{sec:appx-expts} complement the theory by showing our method also works well in more general settings.}
We call \cref{eq:bong} the Bayesian online natural gradient (\bong). 

Our second contribution concerns ways of computing
the expectation in \cref{eq:nELBO,eq:online-variational-loss,eq:bong}.
This is intractable for NNs,
even for variational distributions that are easy to compute,
since
the likelihood takes the form
$p(\vy_{t}\vert\vx_{t},\vtheta_{t}) = p(\vy_{t}\vert f(\vx_{t},\vtheta_{t}))$
with $f(\vx_{t},\vtheta_{t})$, representing the function computed by the network,
is a complex, nonlinear function of $\vtheta_t$.
Many previous approaches have approximated the expected loglikelihood by sampling methods which add variance and computation time depending on the number of samples \citep{BBB,shen2024IVON}. 
We propose a deterministic, closed-form update that applies when the variational distribution is Gaussian (or a sub-family) and the likelihood is an exponential family with natural parameter
$f(\vx_{t},\vtheta_{t})$
and mean parameter
$h(\vx_{t},\vtheta_{t})$
(e.g., for classification, $f$ returns the vector of class logits, $h$ returns class probabilities, and $h={\rm softmax}(f)$).
This update can be derived in two equivalent ways. 
First, we use a local linear approximation of the network 
$h(\vx_{t},\vtheta_{t}) \approx \bar{h}_{t}(\vtheta_{t})$
\citep{Immer2021linear} and a Gaussian approximation of the likelihood 
$\gauss(\vy_{t} | \bar{h}_{t}(\vtheta_{t}), \vR_t)$
\citep{Ollivier2018,Tronarp2018}.
Under these assumptions the expectation in \cref{eq:bong} can be calculated analytically.
Alternatively, we use a different linear approximation 
$f(\vx_{t},\vtheta_{t}) \approx \bar{f}_{t}(\vtheta_{t})$
and a delta approximation
$q_{\vpsi_{t|t-1}} (\vtheta_{t}) \approx \delta_{\vmu_{t|t-1}}(\vtheta_{t})$
where 
$\vmu_{t|t-1} = \expectQ{\vtheta_{t}}{q_{\vpsi_{t\vert t-1}}}$
is the prior mean, so that the expectation in \cref{eq:bong} is replaced by a plugin prediction. 
\rev{The linear($h$)-Gaussian approximation is previously known but the linear($f$)-delta approximation is new, and}
we prove in \cref{thm:lbong} that they yield the same update, 
which we call linearized \bong, or \lbong.
\kpm{
Finally, we discuss different ways of approximating the Hessian of the objective, which is needed for NGD.
}

Our \bong framework unifies several existing methods for Bayesian online learning, 
and it offers new algorithms based on alternative variational families or parameterizations.
We define a large space of methods by
\kpm{
combining 4 different update rules
with 4 different ways of computing the relevant
expected gradients and Hessians
and 3 different variational families
(Gaussians with full, diagonal, and
diagonal-plus-low rank precision matrices).
}
We  conduct experiments systematically testing how these factors affect  performance.
We find support for all three principles of our approach---
NGD, implicit regularization to the prior, and linearization---
in terms of both statistical and computational efficiency.
\rev{Code for our experiments is available at https://github.com/petergchang/bong/.}

%% file: sections/related.tex
\section{Related work}
\label{sec:related}

Variational inference approximates the Bayesian posterior from within some suitable family in a way that bypasses the normalization term \citep{zellner1988optimal,jordan1999introduction}.
A common choice for the variational family is a Gaussian.
For online learning, the exact update equations for Gaussian variational filtering are given by the \RVGA method of \citep{RVGA}. 
This update is implicit but can be approximated by an explicit \RVGA update which we show arises as a special case of \bong.
Most applications of Gaussian VI use a mean-field approximation defined by diagonal covariance, which scales linearly with model size. More expressive but still linear in the model size are methods that express the covariance \citep{Tomczak2020} or precision \citep{SLANG,LRVGA,lofi} as a sum of diagonal and low rank matrices (DLR).
In this paper, we consider variational families defined by full covariance,
diagonal covariance, and DLR covariance.

For NNs and other complicated models, even the variational approximation can be intractable, so methods have been developed to approximately minimize the VI loss.
Bayes by backprop (\bbb) \citep{BBB} learns a variational distribution on NN weights by iterated GD on the VI loss of \cref{eq:nELBO}. They focus on mean-field Gaussian approximations but the approach also applies to other variational families. Here we adapt \bbb to online learning to compare to our methods.

The Bayesian learning rule (\blr) replaces \bbb's GD with NGD \citep{BLR}. 
Several variants of \blr have been developed including \VON and \VOGN for a mean-field Gaussian prior \citep{VON} and \SLANG for a \dlr Gaussian \citep{SLANG}.
\Blr has also been used to derive versions of many classic optimizers
including SGD, RMSprop and Adam \citep{khan2018fast,BLR,lin2024sqrtfree,shen2024IVON}.
Although \blr has been applied to online learning,
we are particularly interested in Bayesian filtering
including in nonstationary environments,
where observations must be processed one at time 
and updates are based on the posterior from the previous step,
often in conjunction with parameter dynamics.
We therefore develop filtering versions of \blr to compare to \bong,
some of which reduce to \VON, \VOGN and \SLANG in the batch setting,
while others are novel.
We also note \blr is a mature theory including several clever tricks we have not yet incorporated into our framework.

\rev{
\cite{BLR} observe that conjugate updating is equivalent to one step of \blr with learning rate 1. This is similar to our \cref{thm:exact-when-conjugate} except that \blr retains the KL term in the variational loss. \Blr and \bong agree in this case because the gradient of the KL is zero on \blr's first iteration: $\nabla_{\vpsi=\vpsi_{t|t-1}}\KLpq{q_{\vpsi}}{q_{\vpsi_{t|t-1}}}=0$. Therefore \bong can be seen as a special case of \blr with one update step per observation and learning rate 1. Our contribution is to recognize that doing a single update step allows the KL term to be dropped entirely, yielding a substantially simpler algorithm which our experiments show also performs better.
}

While \blr allows alternative losses in place of the NLL in \cref{eq:online-variational-loss},
we can also  replacing the KL term with other divergences \citep{knoblauch2022optimization}. Our approach fits within that "generalized VB"
framework in that it drops the divergence altogether.
Our approach of implicitly regularizing to the prior using a single NGD step is also similar to the implicit MAP filter of \citep{bencomo2023implicit} which performs truncated GD from the prior mode. The principal difference is they perform GD on model parameters ($\vtheta_t$) while we do NGD on the variational parameters ($\vpsi_t$). Thus \bong maintains a full prior and posterior while IMAP is more concerned with how the choice of optimizer can substitute for explicit tracking of covariance.

\rev{
We show two other ways to derive the \bong update in \cref{sec:MD-formulation}, one of which is to replace the expected NLL 
in \cref{eq:online-variational-loss} with a linear approximation and solve the resulting equation exactly.
Several past works have taken this approach, arriving at updates similar to ours.
\cite{cherief2019generalization} study streaming variational Bayes and propose solving \cref{eq:online-variational-loss} with a linearized expected NLL. When the variational family is an exponential family their update becomes NGD \citep{khan2017conjugate} and matches the \bong update.
\cite{hoeven2018many} show how mirror descent can be derived as a special case of Exponential Weights \citep{littlestone1994weighted}, which is closely related to Bayesian updating. The resulting algorithm is similar to \bong and follows from linearizing the NLL instead of expected NLL, with an additional delta assumption at the prior mean.
\cite{lyu2021black} study relaxed block-box optimization where the objective is $\arg\min_{\vpsi}\expectQ{f(\vx)}{\vx\sim q_{\vpsi}}$ for some target function $f$. They use a mirror descent formulation with linearized expected loss and KL regularizer and show the resulting update is NGD on expected loss, formally equivalent to our BONG update.
From the perspective of this prior work, our contribution is to express the \bong update simply as NGD on the expected NLL, motivated by replacing the KL with implicit regularization, and to show how this yields a variety of known and novel algorithms for Bayesian filtering.
}

EKF applications to NNs apply Bayesian filtering using a local linear approximation of the network, leading to simple closed form updates \citep{Singhal1988,Puskorius1991}.
The classic EKF assumes a Gaussian observation distribution but it has been extended to other exponential families (e.g. for classification) by matching the mean and covariance in what we call the conditional moments EKF (\CMEKF) \citep{Ollivier2018,Tronarp2018}.
Applying a KL projection to diagonal covariance yields the variational diagonal EKF (\VDEKF) \citep{diaglofi}. Alternatively, projecting to diagonal plus low rank 
\kpm{precision}
using SVD gives \LOFI \citep{lofi}. We derive all these methods as special cases of \lbong.
Further developments in this direction include the method of \citep{titsias2023kalman} which does Bayesian filtering on only the final weight layer, and WoLF \citep{duran2024outlier} which achieves robustness to outliers through data-dependent weighting of the loglikelihood.

%% file: sections/background.tex
\section{Background}
\label{sec:background}

We study online supervised learning where the agent receives input $\vx_{t}\in\real^{D}$ and observation $\vy_{t}\in\real^{\nout}$ on each time step,
which it aims to model with a function $f_t(\vtheta_t) = f(\vx_{t},\vtheta_{t})$ such as a NN with weights $\vtheta_{t}\in\real^{\nparam}$. 
The predictions for $\vy_{t}$ are given by some observation distribution
$p(\vy_t|f_t(\vtheta_t))$.
For example, $f$ may compute the mean for regression or the class logits for classification.

We work in a Bayesian framework where the agent maintains an approximate posterior distribution over $\vtheta_{t}$ after observing data 
$\data_{t} = \left\{ (\vx_{k},\vy_{k}\right)_{k=1}^{t}\} $.
The filtering posterior 
    $q_{\vpsi_{t}}(\vtheta_{t}) 
    \approx p(\vtheta_{t}\vert\data_{t})$
is approximated within some parametric family indexed by the variational parameter $\vpsi_{t}$.
We allow for nonstationarity by assuming $\vtheta$ changes over time according to some dynamic model $p(\vtheta_{t}\vert\vtheta_{t-1})$.
By pushing the posterior from step $t-1$ through the dynamics we obtain a prior for step $t$ given by 
$q_{\vpsi_{t|t-1}}(\vtheta_{t}) \approx p(\vtheta_{t}\vert\data_{t-1})$.
For example suppose the variational posterior
from the previous step is Gaussian,
$q_{\vpsi_{t-1}}(\vtheta_{t-1})
= \gauss(\vtheta_{t-1}|\vmu_{t-1},\vSigma_{t-1})$,
and the dynamics model is an
Ornstein-Uhlenbeck process,
as proposed in prior work
\citep{Kurle2020,titsias2023kalman}
to handle non-stationarity,
i.e., the dynamics model has the form
$\vtheta_{t} \sim \gauss(
\gamma_t \vtheta_{t-1} + (1-\gamma_t) \vmu_0,
\vQ_t)$,
where $\vQ_t=(1-\gamma_t^2) \vSigma_0$ is the covariance of the noise process,
$0 \leq \gamma_t \leq 1$ is the degree of drift,
and $p(\vtheta_0)=\gauss(\vmu_0,\vSigma_0)$ is the prior.
In this case, 
the parameters of the prior
predictive distribution are 
$\vmu_{t|t-1} = \gamma_t \vmu_{t-1} + (1-\gamma_t) \vmu_0$
and
$\vSigma_{t|t-1} = \gamma_t^2 \vSigma_{t-1}+\vQ_t$. 
In general the predict step may require approximation to stay in the variational family
(e.g., if the dynamics are nonlinear).
In this paper,
our focus is the update step from $\vpsi_{t|t-1}$ to $\vpsi_{t}$ upon observing $(\vx_{t},\vy_{t})$,
so for simplicity we assume
constant (static) parameters,
i.e., $p(\vtheta_t|\vtheta_{-1})=\delta(\vtheta_t - \vtheta_{t-1})$ (equivalently $\gamma_t=1$), so $\vpsi_{t|t-1} = \vpsi_{t-1}$;
however, our method can trivially handle non-stationary parameters.

Variational inference seeks an approximate posterior that minimizes the KL divergence from the exact Bayesian update from the prior. 
In the online setting this becomes
\begin{align}
    \vpsi^*_t
    = \underset{\vpsi}{\arg\min} \,
    \KLpq
        {q_{\vpsi}(\vtheta_{t})} 
        {Z_t^{-1} q_{\vpsi_{t|t-1}}(\vtheta_{t}) \, p(\vy_{t}| f_t(\vtheta_{t}))}
    = \underset{\vpsi}{\arg\min} \, \loss_t (\vpsi)
    \label{eq:online-VB}
\end{align}
where $\loss_t$ is the online VI loss defined in \cref{eq:online-variational-loss},
and the normalization term $Z_t$ (which depends on $\vx_t$) drops out as an additive constant.
Our goal is an efficient approximate solution to this variational optimization problem.

We will sometimes assume the variational
posterior $q_{\vpsi}$ is an exponential family 
distribution with natural parameter $\vpsi$ so that
$
q_{\vpsi_{t}}(\vtheta_{t})
    = \exp\left(\vpsi_{t}^{\trans} T(\vtheta_{t}) -\Phi(\vpsi_{t}\right))
$,
with log-partition function $\Phi$ and sufficient statistics $T(\vtheta_{t})$.
Assuming $\Phi$ is strictly convex (which holds in the cases we study) there is a bijection between $\vpsi\rev{_t}$ and the dual (or expectation) parameter
$\vrho_{t} = \expectQ {T(\vtheta_{t})} {\vtheta_{t}\sim q_{\vpsi_{t}}}$.
Classical thermodynamic identities imply 
that the Fisher information matrix has the form
$\vF_{\vpsi_{t}} = \partial\vrho_{t} / \partial\vpsi_{t}$.
This has important implications for NGD on exponential families \citep{BLR}
because it implies that for any function $\ell$ defined on the variational parameter space the natural gradient wrt natural parameters
$\vpsi_t$
is the regular gradient wrt the dual parameters
$\vrho_t$,
i.e., 
 $
    \vF_{\vpsi_{t}}^{-1} \nabla_{\vpsi_{t}} \ell
    = \nabla_{\vrho_{t}} \ell.
$

\eat{
and satisfies the classical thermodynamic identity
\begin{equation}
    \nabla_{\vpsi_{t}}\Phi(\vpsi_{t})
    = \vrho_{t}
    \label{eq:mirror-map}
\end{equation}
We will assume $q_{\vpsi}$ is a minimal exponential family, meaning $\Phi$ is strictly convex and therefore $\vpsi\mapsto\vrho$ is invertible,
so that $\vrho$ can be treated as a function of $\vpsi$ and vice versa. 
\Cref{eq:mirror-map} implies the Fisher can be written as
\begin{align}
    \vF_{\vpsi_{t}} 
    &= -\expectQ
        {\nabla_{\vpsi_{t}}^{2}\log q_{\vpsi_{t}}(\vtheta_{t})}
        {\vtheta_{t}\sim q_{\vpsi_{t}}}
    = \frac{\partial\vrho_{t}}{\partial\vpsi_{t}}
\end{align}
}

%% file: sections/methods.tex
\section{Methods}

We propose to approximate the variational optimization problem in \cref{eq:online-VB} using the \bong update in \cref{eq:bong}. 
When $q_{\vpsi}$ is an exponential family,
the fact that the
natural gradient wrt the natural parameters
$\vpsi_t$
is the regular gradient wrt the dual parameters
$\vrho_t$
implies an equivalent mirror descent form
(see \cref{sec:MD-formulation} for further analysis of \bong from the MD perspective):
\begin{equation}
    \vpsi_{t} 
    = \vpsi_{t\vert t-1} 
    + \nabla_{\vrho_{t|t-1}} 
    \expectQ
        {\log p(\vy_{t}\vert\vx_{t},\vtheta_{t})}
        {\vtheta_{t}\sim q_{\vpsi_{t\vert t-1}}}
    \label{eq:bong-MD}
\end{equation}
This is NGD with unit learning rate on the variational loss in \cref{eq:online-variational-loss} but ignoring the 
$\KLpq {q_{\vpsi}} {q_{\vpsi_{t|t-1}}}$
term. 
In this section we first prove this method is optimal when the model is conjugate
and then describe extensions to more complex cases of practical interest.

\input{sections/methods-conjugate}
\input{sections/gaussian}
\input{sections/MC}
\input{sections/methods-lbong}
\input{sections/EF}
\input{sections/method-space}

%% file: sections/methods-conjugate.tex
\subsection{Conjugate case}
\label{sec:conjugate}

Our approach is motivated by the following result which states that \bong matches exact Bayesian inference when the variational distribution and the likelihood are conjugate exponential families:
\begin{proposition} \label{thm:exact-when-conjugate}
    Let the observation distribution (likelihood) be an exponential family with natural parameter $\vtheta_{t}$ 
    (where  
    \kpm{$T_l(\vy_t)=\vy_t$
    is the sufficient statistics for the likelihood
    and $A(\vtheta_t)$ is the log-partition function)
    }
    \begin{align}
        p_{t}(\vy_{t}\vert\vtheta_{t}) 
        & \kpm{=} \exp \left(
            \vtheta_{t}^{\trans} \vy_{t}
            - A(\vtheta_{t})
            - b(\vy_{t})
        \right)
        \label{eq:conjugate-likelihood}
    \end{align}
    and let the prior be the conjugate
    exponential family
    \begin{align}
        q_{\vpsi_{t\vert t-1}}(\vtheta_{t}) 
        &= \exp \left(
            \vpsi_{t|t-1}^{\trans} T(\vtheta_{t})
            - \Phi(\vpsi_{t|t-1})
        \right)
\end{align}
with $T(\vtheta_{t}) = \left[\vtheta_{t}; -A(\vtheta_{t})\right]$.
Then the exact Bayesian update agrees with \cref{eq:bong-MD}.
\end{proposition}

The proof is in \cref{sec:proofs}. 
Writing the natural parameters of the prior as
$\vpsi_{t\vert t-1} = [\vchi_{t\vert t-1};\nu_{t\vert t-1}]$,
we show the Bayesian update and \bong both yield
$\vchi_t = \vchi_{t|t-1} + \vy_t$ and 
$\nu_t = \nu_{t|t-1} + 1$.
Intuitively, we are just accumulating a sum of the observed sufficient statistics, and a counter of
the sample size (number of observations seen so far).

%% file: sections/gaussian.tex
\subsection{Variational case}

In practical settings the conjugacy assumption of \cref{thm:exact-when-conjugate} will not be met, so \cref{eq:bong,eq:bong-MD} will only approximate the Bayesian update.
In this paper we restrict to Gaussian variational families.
We refer to the unrestricted case as FC (full covariance),
defined by the variational distribution
\begin{align}
    q_{\vpsi_{t|t-1}}(\vtheta_{t}) 
    &= \gauss \left(\vtheta_{t} | \vmu_{t|t-1}, \vSigma_{t|t-1}\right)
\end{align}
where $\vSigma_{t|t-1}$ can be any \rev{positive semi-definite (}PSD\rev{)} matrix. The natural and dual parameters are
$\vpsi = (\vSigma^{-1} \vmu, -\tfrac{1}{2}{\rm vec}(\vSigma^{-1}))$ and
$\vrho = (\vmu, {\rm vec}(\vmu\vmu^\trans + \vSigma))$.
\Cref{sec:FC-Nat-bong} shows that \cref{eq:bong-MD} 
translated back to $(\vmu,\vSigma)$
gives the following \bong update for the FC case:
\begin{align}
    \vmu_{t} 
    &= \vmu_{t|t-1} 
    + \vSigma_{t} 
    \underbrace{\expectQ
        {\nabla_{\vtheta_{t}} 
        \log p(\vy_{t}| f_{t}(\vtheta_{t}))}
        {\vtheta_{t} \sim q_{\vpsi_{t|t-1}}}
        }_{\kpm{\vg_t}}
    \label{eq:RVGA-explicit-mean} \\
    \vSigma_{t}^{-1} 
    &= \vSigma_{t|t-1}^{-1} - 
    \underbrace{
    \expectQ
        {\nabla_{\vtheta_{t}}^{2} \log p(\vy_{t} |  f_{t}(\vtheta_{t}))}
        {\vtheta_{t} \sim q_{\vpsi_{t|t-1}}}
        }_{\kpm{\vG_t}}
    \label{eq:RVGA-explicit-cov}
\end{align}
which matches the explicit update in the RVGA method of \citep{RVGA}.

%% file: sections/MC.tex
\subsection{Monte Carlo approximation}
\label{sec:MC}

The integrals over the prior $q_{\vpsi_{t|t-1}}$ in 
\cref{eq:RVGA-explicit-mean,eq:RVGA-explicit-cov} 
are generally intractable and must be approximated.
One option is to use Monte Carlo, in what we call \mcbong.
Given $\nsample$ independent samples 
$\hat{\vtheta}_t^{(m)} \sim q_{\vpsi_{t|t-1}}$,
we estimate the expected gradient
$\vg_t=\expectQ
        {\nabla_{\vtheta_{t}}\log p(\vy_{t}\vert f_{t}(\vtheta_{t}))}
        {\vtheta_{t}\sim q_{\vpsi_{t|t-1}}}$
and expected Hessian
$\vG_t = \expectQ
        {\nabla_{\vtheta_{t}}^{2}\log p\left(\vy_{t}\vert f_{t}\left(\vtheta_{t}\right)\right)}{\vtheta_{t}\sim q_{\vpsi_{t|t-1}}}$
as the empirical means
\begin{alignat}{3}
    \gMC_t 
    &= \frac{1}{\nsample} \sum_{m=1}^\nsample \hat{\vg}_t^{(m)},
    \quad &
     \hat{\vg}_t^{(m)}
    &= \nabla_{\vtheta_{t}=\hat{\vtheta}_{t}^{(m)}} \log p(\vy_{t}\vert f_{t}(\vtheta_{t}))
        \label{eq:gMC} \\
    \GMCH_t &= 
         \frac{1}{\nsample} \sum_{m=1}^\nsample \hat{\vG}_t^{(m)},
         \quad &
         \hat{\vG}_t^{(m)}  
         &= \nabla^2_{\underset{}{\vtheta_{t} = \hat{\vtheta}_t^{(m)}}} \log p(\vy_{t}\vert f_{t}(\vtheta_{t}))
    \label{eq:GMCH} 
\end{alignat}
We use $\GMCH$ only for small models.
Otherwise we use empirical Fisher (\cref{sec:EF}).

\eat{
\footnote{For algorithms that iterate over an inner loop on each time step indexed by $i$ (BLR and BBB) we use the same approach with modified notation (subscripts):
$\hat{\vtheta}^{(m)}_{t,i} \sim q_{\vpsi_{t,i}}$,
$\hat{\vG}_{t,i} = [\hat{\vg}^{(m)}_{t,i}]_{m\in[\nsample]}$,
$\gMC_{t,i} \approx \expectQ{\nabla_{\vtheta_t} \log p}{q_{\vpsi_{t,i}}}$, and
$\GMCH_{t,i} \approx \expectQ{\nabla^2_{\vtheta_t} \log p}{q_{\vpsi_{t,i}}}$.
}
}

%% file: sections/methods-lbong.tex
\subsection{Linearized \bong}
\label{sec:linearized}

As an alternative to \mcbong, we propose a linear approximation 
we call \lbong 
that yields a deterministic and closed-form update.
Assume the likelihood is an exponential family as in \cref{thm:exact-when-conjugate} 
but with natural parameter predicted by some function 
$f_t(\vtheta_t) = f(\vx_{t},\vtheta_{t})$:
\begin{align}
    p(\vy_t | \vx_t, \vtheta_t)
    \rev{=}
    \exp\left(
        f_t\!\left(\vtheta_t\right)^\trans \vy_t 
        - A(f_t(\vtheta_t))
        - b(\vy_t)
    \right)
    \label{eq:exfam-likelihood-general}
\end{align}
We also define the dual (moment) parameter of the likelihood as
$h_t(\vtheta_t) = \expect{\vy_t | f_t(\vtheta_t)}$.
In a NN, $f_t$ and $h_t$ are related by the final response layer.
For example in classification $f_t$ and $h_t$ give the class logits and probabilities, 
with $h_t(\vtheta_t) = {\rm softmax}(f_t(\vtheta_t))$,
with $\vy_t$ being the one-hot encoding.

We now define two methods for approximating the expected gradient $\vg_t$ and expected Hessian $\vG_t$,
based on linearizing the predictive model at the prior mean $\vmu_{t|t-1}$
in terms of either $f_t(\vtheta_t)$ or $h_t(\vtheta_t)$,
and then prove their equivalence.

The {\bf linear($h$)-Gaussian} approximation  \citep{Ollivier2018,Tronarp2018}
linearizes $h_t(\vtheta_t)$
\begin{align}
    \bar{h}_{t}(\vtheta_{t}) 
    &= \hat{\vy}_t + \vH_{t}(\vtheta_{t}-\vmu_{t|t-1})
    \label{eq:hbar} \\
    \hat{\vy}_t 
    &= h_t(\vmu_{t|t-1}) \\
    \vH_{t} 
    &= \rev{\frac{\partial h_t}{\partial \vtheta_t}_{\vert \vtheta_t=\vmu_{t|t-1}}} 
    \label{eqn:jacH}
\end{align}
and approximates the likelihood by a Gaussian with variance based at $\vmu_{t|t-1}$
\begin{align}
    \bar{p}_t^{\rm LG}(\vy_t|\vtheta_t)
    = \gauss(\vy_t | \bar{h}_t(\vtheta_t), \vR_t), 
    \quad \vR_t 
    = \var{\vy_t | \vtheta_t = \vmu_{t|t-1}} 
    \label{eq:gaussian-likelihood}
\end{align}

The {\bf linear($f$)-delta approximation}
linearizes $f_t(\vtheta_t)$ and maintains the original exponential family likelihood distribution in \cref{eq:exfam-likelihood-general}
\begin{align}
    \bar{f}_{t}(\vtheta_{t}) 
    &= f_t(\vmu_{t|t-1}) + \vF_{t}(\vtheta_{t}-\vmu_{t|t-1})
    \label{eq:fbar} \\
    \vF_{t}
    &= \rev{\frac{\partial f_t}{\partial\vtheta_t}_{\vert\vtheta_t=\vmu_{t|t-1}}}
    \label{eqn:jacF} \\
    \bar{p}^{\rm LD}_t(\vy_t|\vtheta_t) 
    &\propto
    \exp\left(
        \bar{f}_t(\vtheta_t)^\trans \vy_t 
        - A(\rev{\bar{f}}_t(\vtheta_t))
        - b(\vy_t)
    \right)
\end{align}
It also uses a plug-in approximation that replaces 
$q_{\vpsi_{t|t-1}}(\vtheta_t)$ 
with a point mass 
$\delta_{\vmu_{t|t-1}}(\vtheta_t)$ 
so that the expected gradient and Hessian are approximated by their values at the prior mean, i.e., 
$\nabla_{\vtheta_t = \vmu_{t|t-1}} \log \bar{p}_t^{\rm LD} (\vy_{t}| \vtheta_t)$ 
and
$\nabla^2_{\vtheta_t = \vmu_{t|t-1}} \log \bar{p}_t^{\rm LD} (\vy_{t}| \vtheta_t)$,
\kpm{rather than being sampled}.
    
\begin{proposition}
\label{thm:lbong}
    Under a Gaussian variational distribution,
    the linear($h$)-Gaussian and linear($f$)-delta approximations 
    yield the same values for the expected gradient and Hessian
    \begin{align}
        \gL_t
        &= \vH_t^\trans \vR_t^{-1} (\vy_t - \hat{\vy}_t) 
        \label{eq:gL}\\
        \GLH_t
        &= - \vH_t^\trans \vR_t^{-1} \vH_t
        \label{eq:GL}
    \end{align}
\end{proposition}

See \cref{sec:proofs} for the proof.
The main idea for the $\gL_t$ part is that
the linear-Gaussian assumptions make the gradient linear in $\vtheta_t$
so the expected gradient equals the gradient at the mean.
The main idea for the $\GLH_t$ part is that
eliminating the Hessian of the NN requires different linearizing assumptions for the Gaussian and delta approximations,
and the remaining nonlinear terms (from the log-likelihood in \cref{eq:exfam-likelihood-general}) agree because of the property of exponential families
that the Hessian of the log-partition $A$ equals the conditional variance $\vR_t$.

Applying \cref{thm:lbong} to \cref{eq:RVGA-explicit-mean,eq:RVGA-explicit-cov} gives the \lbong update for a FC Gaussian prior
\begin{align}
\vmu_{t} & =\vmu_{t|t-1}+\vK_{t}(\vy_{t}-\hat{\vy}_{t})\\
\vSigma_{t} & =\vSigma_{t|t-1}-\vK_{t}\vH_{t}\vSigma_{t|t-1}\\
\vK_{t} & =\vSigma_{t|t-1}\vH_{t}^{\trans}\left(\vR_{t}+\vH_{t}\vSigma_{t|t-1}\vH_{t}^{\trans}\right)^{-1} 
\end{align}
where $\vK_t$ is the Kalman gain matrix
(see \cref{sec:lbong-fc}). This matches the \CMEKF \citep{Tronarp2018,Ollivier2018}.

\eat{
\begin{align}
    \vmu_{t} 
    &= \vmu_{t|t-1} + \vSigma_{t} \vH_{t}^{\trans} \vR_{t}^{-1} (\vy_{t}-\hat{\vy}_{t})
    \label{eq:EFEKF-update-mean}\\
    \vSigma_{t}^{-1} 
    &= \vSigma_{t|t-1}^{-1} + \vH_{t}^{\trans} \vR_{t}^{-1} \vH_{t}
    \label{eq:EFEKF-update-cov}
\end{align}
    Alternatively, we can write the result in terms
    of the posterior covariance instead of the posterior precision, yielding the \CMEKF \citep{Tronarp2018,Ollivier2018}
     \begin{align}
\vSigma_{t} 
&= \vSigma_{t|t-1} - \vK_t \vH_{t} \vSigma_{t|t-1}
         = \vSigma_{t|t-1} - \vK_t \vS_t \vK_t^\trans 
          \label{eq:EFEKF-update-cov2}
         \\
\vK_t &= \vSigma_{t|t-1} \vH_t^\trans \vS_t^{-1} \\
\vS_t &= \vH_t \vSigma_{t|t-1} \vH_t^\trans+ \vR_t
    \end{align}
where $\vK_t$
is the Kalman gain matrix
and $\vS_t$ is the covariance matrix
of the prior predictive in observation space.
The conversion from precision form
    to covariance form
    follows from the matrix inversion lemma,
    which gives
    $\vK_t = \vSigma_{t|t-1} \vH_t^\trans 
(\vH_t \vSigma_{t|t-1} \vH_t^\trans+ \vR_t)^{-1}
=(\vSigma_{t|t-1}^{-1}
+ \vH_t^\trans \vR_t^{-1} \vH_t)^{-1}
\vH_t^\trans \vR_t^{-1}
$.

In \citep{Tronarp2018} the quantities  $\bar{h}_{t}(\vtheta_{t})$ and $\vR_t$ are called conditional moments, so we shall call this method \CMEKF.
}

%% file: sections/EF.tex
\subsection{Empirical Fisher}
\label{sec:EF}

\setlength\intextsep{-7pt}
\begin{wraptable}{r}{3.3cm}
    \centering
    \footnotesize
    \begin{tabular}{cc}
    \hline
    Name & Eqs. \\ 
    \hline
    \hessmc & \eqref{eq:gMC}, \eqref{eq:GMCH} \\
    \hesslin & \eqref{eq:gL}, \eqref{eq:GL} \\
    \efmc & \eqref{eq:gMC}, \eqref{eq:GMCEF} \\
    \eflin & \eqref{eq:gL-cheap}, \eqref{eq:GLEF} \\
    \hline
    \end{tabular}
    \caption{The 4   Hessian approximations.
    }
    \label{tab:hessian}
\end{wraptable}

The methods in \cref{sec:MC,sec:linearized}
require explicitly computing the Hessian of the loss (\hessmc)
or the Jacobian of the network (\hesslin).
These are too expensive for large models or high-dimensional observations.
Instead we can use an empirical Fisher approximation that 
replaces the Hessian with the outer product of the gradient
(see e.g, \citep{martens2020new}).

For the \efmc variant, 
 we make the following approximation:
\begin{align}
  \GMCEF_t &=  -\frac{1}{\nsample} 
  \gradmat_t \gradmatT_t
   \label{eq:GMCEF}
\end{align}
where
$\gradmat_t = [\hat{\vg}_t^{(1)},\dots,\hat{\vg}_t^{(\nsample)}]$
is the $\nparam\times\nsample$ matrix of gradients from the MC samples.

We can also consider a similar approach for the \eflin variant that is Jacobian-free and sampling-free. 
Note that if $\hat{\vy}_t$ were the true value of $\expect{\vy_t | \vx_t}$ 
(i.e., if the model were correct)
then we would have 
$\expect{(\vy_t - \hat{\vy}_t) (\vy_t - \hat{\vy}_t)^\trans} = \vR_t$,
implying
$\expect{\gL_t \left(\gL_t\right)^\trans} = -\GLH_t$.
This suggests using
\begin{align}
    \gLEF_t
        &= \nabla_{\vtheta_t = \vmu_{t|t-1}} 
        \left[ -\tfrac{1}{2}
            \left(\vy_t - h_t(\vtheta_t)\right)^\trans \vR_t^{-1} (\vy_t - h_t(\vtheta_t))
        \right] \\
        &= \kpm{\left(
        \frac{\partial h_t(\vtheta_t)}
        {\partial \vtheta_t}
        \right)_{\vtheta_t=\vmu_{t|t-1}}^\trans
        \vR_t^{-1} 
        (\vy_t - h_t(\vmu_{t|t-1}))
        = \gL_t
        }
    \label{eq:gL-cheap} \\
    \GLEF_t
        &= - \gL_t \left(\gL_t\right)^\trans
        \label{eq:GLEF}
\end{align}
where
\cref{eq:GLEF} is the EF approximation to \cref{eq:GL}.
%

A more accurate EF approximation is possible by sampling virtual observations $\tilde{\vy}_t$ from $p(\cdot | f_t(\hat{\vtheta_{t}}^{(m)}))$ or $p(\cdot | f_t(\vmu_{t|t-1}))$ and using them for the gradients in \cref{eq:GMCEF} or \cref{eq:GLEF} (respectively) \citep{martens2020new,kunstner2020limitations}.
However, in our experiments we use the actual observations $\vy_t$ which is faster and follows previous work (e.g., \citep{VON}).

%% file: sections/method-space.tex

\input{sections/variants}
\input{sections/families}
\input{sections/algo-overview}

%% file: sections/variants.tex
\subsection{\kpm{Update rules}}

\label{sec:variants}

\setlength\intextsep{-15pt}
\begin{wraptable}{r}{5cm}
    \centering
    \footnotesize
    \begin{tabular}{ccc}
    \hline
    Name & Loss & Update \\ 
    \hline
    \bong & $\expect{\rm NLL}$ & NGD($\niter=1$) \\
    \bog & $\expect{\rm NLL}$ & GD($\niter=1$) \\
    \blr & ${\rm VI}$ & NGD($\niter \geq 1$) \\
    \bbb & ${\rm VI}$ & GD($\niter \geq 1$) \\
    \hline
    \end{tabular}
    \caption{The 4 update algorithms.
        }
    \label{tab:space}
\end{wraptable}

In addition to the four ways of approximating
the expected Hessian (summarized in \cref{tab:hessian}),
we also consider four variants of \bong,
based on what kind of loss we optimize
and what kind of update we perform,
as 
 we describe below.
See \cref{tab:space} for a summary.



\kpm{
{\bf \bong} (Bayesian online natural gradient)
performs one step of  NGD
on the expected log-likelihood.
We set learning rate to $\alpha_t=1$ since this is optimal for conjugate models. The update (for an exponential variational family) is as in \cref{eq:bong-MD}:
\begin{equation}
    \vpsi_{t} 
    = \vpsi_{t\vert t-1} 
    + \nabla_{\vrho_{t|t-1}} 
    \expectQ
        {\log p(\vy_{t}\vert\vx_{t},\vtheta_{t})}
        {\vtheta_{t}\sim q_{\vpsi_{t\vert t-1}}}
    \label{eq:bong-MD2}
\end{equation}
}

{\bf \bog} (Bayesian online gradient)
performs one step of GD (instead of NGD)
on the expected log-likelihood. We include a learning rate $\alpha$ because GD does not have the scale-invariance of NGD:
\begin{equation}
    \vpsi_{t} 
    = \vpsi_{t} 
    + \alpha_t \nabla_{\vpsi_{t}} 
    \expectQ
        {\log p(\vy_{t}|f_t(\vtheta_{t}))}
        {\vtheta_{t}\sim q_{\vpsi_{t}}}
    \label{eq:bog}
\end{equation}

{\bf \blr} (Bayesian learning rule, \citep{BLR}) 
uses NGD (like \bong) but optimizes
the VI loss using multiple iterations,
instead of optimizing the expected NLL with a single step.
When modified to the online setting,
\blr starts an inner loop at each time step with
$\vpsi_{t,0} = \vpsi_{t\vert t-1}$
and iterates
\begin{equation}
    \vpsi_{t,i} 
    = \vpsi_{t,i-1} 
    + \alpha_t \vF_{\vpsi_{t,i-1}} \nabla_{\vpsi_{t,i-1}} \!\! 
    \left(
        \expectQ
            {\log p(\vy_{t}|f_t(\vtheta_{t}))}
            {\vtheta_{t}\sim q_{\vpsi_{t,i-1}}}
        - \KLpq{q_{\vpsi_{t,i-1}}}{q_{\vpsi_{t|t-1}}}
    \right)
    \label{eq:blr}
\end{equation}

For an exponential variational family this can be written in mirror descent form 
\begin{equation}
    \vpsi_{t,i} 
    = \vpsi_{t,i-1} 
    + \alpha_t \nabla_{\vrho_{t,i-1}} \!\!
    \left(
        \expectQ
            {\log p(\vy_{t}|f_t(\vtheta_{t}))}
            {\vtheta_{t}\sim q_{\vpsi_{t,i-1}}}
        - \KLpq{q_{\vpsi{t,i-1}}}{q_{\vpsi_{t|t-1}}}
    \right)
    \label{eq:blr-MD}
\end{equation}

{\bf \bbb} 
(Bayes By Backprop, \citep{BBB})
is like \blr but uses GD instead of NGD.
When 
adapted to online learning, it starts each time step at
$\vpsi_{t,0} = \vpsi_{t\vert t-1}$
and iterates with GD:
\begin{equation}
    \vpsi_{t,i} 
    = \vpsi_{t,i-1} 
    + \alpha_t \nabla_{\vpsi_{t,i-1}} \!\!
    \left(
        \expectQ
            {\log p(\vy_{t}|f_t(\vtheta_{t}))}
            {\vtheta_{t}\sim q_{\vpsi_{t,i-1}}}
        - \KLpq{q_{\vpsi_{t,i-1}}}{q_{\vpsi_{t|t-1}}}
    \right)
    \label{eq:bbb}
\end{equation}


%% file: sections/families.tex
\subsection{Variational families and their parameterizations}
\label{sec:families}

We investigate five variational families for the posterior distribution: 
(1) FC Gaussian using natural parameters $\vpsi = (\vSigma^{-1} \vmu, -\tfrac{1}{2}\vSigma^{-1})$,
(2) FC Gaussian using central moment parameters $\vpsi = (\vmu, \vSigma)$,
(3) diagonal Gaussian using natural parameters $\vpsi = (\vsigma^{-2} \vmu, -\tfrac{1}{2} \vsigma^{-2})$ \rev{(using elementwise exponents and products)},
(4) diagonal Gaussian using central moment parameters $\vpsi = (\vmu, \vsigma^2)$, and
(5) DLR Gaussian with parameters $\vpsi = (\vmu, \vUpsilon, \vW)$ 
and precision $\vSigma^{-1} = \vUpsilon + \vW \vW^\trans$
where 
$\vUpsilon \in \real^{\nparam\times\nparam}$ is diagonal and
$\vW \in \real^{\nparam\times\rank}$ with $\rank\ll\nparam$.
The moment parameterizations are included to test the importance of using natural parameters per \cref{thm:exact-when-conjugate}.
The diagonal family allows learning of large models because it scales linearly in the model size $\nparam$.
DLR also scales linearly but is more expressive than diagonal, maintaining some of the correlation information between parameters that is lost in the mean field (diagonal) approximation \citep{LRVGA,SLANG,lofi}.

Optimizing the \bong objective
wrt $(\vmu,\vUpsilon, \vW)$
using NGD methods is challenging 
because the Fisher information matrix in this parameterization cannot be efficiently inverted.
Instead we first derive the update
wrt the FC natural parameters
(leveraging the fact that the prior
$\vSigma_{t|t-1}^{-1}$ is DLR
to make this efficient),
and then use SVD to project the posterior
precision back to low-rank form,
following our prior \LOFI work
\citep{lofi}.
However, if we omit the Fisher preconditioning matrix and use GD as in \bog and \bbb,
we can directly optimize the objective
wrt $(\vmu,\vUpsilon, \vW)$
(see \cref{sec:DLR-deriv}).

%% file: sections/algo-overview.tex
\subsection{Overall space of methods}
\label{sec:method-space}

\input{sections/table-algo}

\eat{
\Cref{tab:methods} lists the space of methods we consider defined by five binary contrasts: 
update via NGD vs.\ GD, 
loss defined by $\expect{\rm NLL}$ vs.\ VI, integration via MC vs.\ linearizing,
FC vs.\ DLR family,
and natural vs.\ source parameterization.
}

Crossing the four algorithms in \cref{tab:space}, 
\rev{the four methods of approximating the Hessian in \cref{tab:hessian}, and the five variational families yields 80 algorithms. \Cref{tab:methods} shows the 36 based on the three tractable Hessian approximations and the three variational families that use natural parameters.}
\eat{
\Cref{tab:methods} lists 40 of the methods
(focusing on the \efmc and \hesslin
Hessian approximations that we use in our experiments),
and states their time complexity, in terms of
$\nparam$, the dimensionality  of the parameter space;
$\nsample$, the number of MC samples (not relevant for the linearized variants);
$\niter$, the number of iterations per time step
(we fix $\niter=1$ for \bong and \bog variants);
$\rank$, the rank of the DLR approximation
(not relevant for full or diagonal covariance); and
$\nout$, the size of the output vector $\vy_t$
(relevant only for the \hesslin variants).
}
Update equations for all the algorithms are derived in \cref{sec:derivations}.
Pseudocode is given in \cref{sec:pseudocode}.

%% file: sections/table-algo.tex
\setlength\intextsep{12pt}
\begingroup
\setlength{\tabcolsep}{3pt}
\begin{table}[t]
\centering
 \footnotesize
\begin{tabular}{ccccc}
\hline 
\multicolumn{2}{l}{ } & \multicolumn{3}{c}{{Variational Family}}\tabularnewline
\cline{3-5} \cline{4-5} \cline{5-5} 
{Update} & {Hessian} & {Full} & {Diag} & {DLR}\tabularnewline
\hline 
    {BONG} & {MC-EF} & $O(MP^{2})$ [\RVGA\!] & $O(MP)$ & $\overset{}{O}((R+M)^{2}P)$ \tabularnewline
{BLR} & {MC-EF} & $O(IP^{3})$ & $O(IMP)$ [\VON\!] & $O(I(R+M)^{2}P)$ [\SLANG\!]\tabularnewline
    {BOG} & {MC-EF} & $O(P^{3})$ & $O(MP)$ & $O(RMP)$ \tabularnewline
{BBB} & {MC-EF} & $O(IP^{3})$ & $O(IMP)$ [\bbb\!] & $O(IR(R+M)P)$ \tabularnewline
\hline
{BONG} & {LIN-HESS} & $O(CP^{2})$ [\CMEKF\!] 
& $O(C^{2}P)$ [\VDEKF\!] & $O((R+C)^{2}P)$ [\LOFI\!] \tabularnewline
{BLR} & {LIN-HESS} & {{ $O(IP^{3})$}} & {{ $O(IC^{2}P)$}} & {{ $O(I(2R+C)^{2}P)$}}\tabularnewline
{BOG} & {LIN-HESS} & {{ $O(P^{3})$}} & {{ $O(C^{2}P)$}} & {{ $O(C(C+R)P)$}}\tabularnewline
{BBB} & {LIN-HESS} & {{ $O(IP^{3})$}} & {{ $O(IC^{2}P)$}} & {{ $O(I(C+R)RP)$}}\tabularnewline
\hline 
{BONG} & {LIN-EF} & $O(P^{2})$ 
& $O(P)$ & $O(R^{2}P)$ \tabularnewline
{BLR} & {LIN-EF} & {{ $O(IP^{3})$}} & {{ $O(IP)$}} & {{ $O(IR^{2}P)$}}\tabularnewline
{BOG} & {LIN-EF} & {{ $O(P^{3})$}} & {{ $O(P)$}} & {{ $O(RP)$}}\tabularnewline
{BBB} & {LIN-EF} & {{ $O(IP^{3})$}} & {{ $O(IP)$}} & {{ $O(IR^{2}P)$}}\tabularnewline
\hline 
\end{tabular}
\caption{
Time complexity of the algorithms.
$P$: params, $C$: observation dim, $M$: MC samples,
$I$: iterations, $R$: DLR rank.
We assume 
$\nparam \gg \{\niter,\nout,\rank,\nsample\}$ 
so display only the terms of leading order in $\nparam$.
\rev{
Time complexities for \mchess algorithms (not shown) are always at least as great as for the corresponding \mcef.
Full (full covariance) and Diag (diagonal covariance) columns indicate natural parameters; corresponding algorithms using moment parameters have the same complexities except \bog-\fcmom which is $O(MP^2)$ for \mcef, $O(CP^2)$ for \linhess, and $O(P^2)$ for \linef.
}
[Names] correspond to the following
existing methods (or variants thereof)
in the literature:
\RVGA: \citep{RVGA} (explicit update version);
\VON: \citep{VON} (modified for online);
\SLANG: \citep{SLANG} (modified for online);
\bbb: \citep{BBB} (modified for online \rev{and uses moment parameters});
\CMEKF: \citep{Ollivier2018,Tronarp2018};
\VDEKF: \citep{diaglofi};
\LOFI: \citep{lofi}.
}
    \label{tab:methods}
\end{table}
\endgroup

%% file: sections/experiments.tex
\section{Experiments}
\label{sec:experiments}

\newcommand{\Ntrain}{N_{\text{train}}}
\newcommand{\Ntest}{N_{\text{test}}}
\newcommand{\Dtrain}{\data^{\text{train}}}
\newcommand{\Dtest}{\data^{\text{test}}}

\newcommand{\MLPsmall}{\textsc{mlp}_{\text{S}}}
\newcommand{\MLPmed}{\textsc{mlp}_{\text{M}}}
\newcommand{\MLPlarge}{\textsc{mlp}_{\text{L}}}
\newcommand{\CNNsmall}{\textsc{cnn}_{\text{S}}}
\newcommand{\CNNlarge}{\textsc{cnn}_{\text{L}}}

\newcommand{\NLPD}{\text{NLPD}\xspace}
\newcommand{\NLL}{\text{NLL}\xspace}
\newcommand{\LL}{\text{LL}\xspace}
\newcommand{\LLPI}{\text{LL-PI}\xspace}
\newcommand{\LLMC}{\text{LL-MC}\xspace}
\newcommand{\LLLin}{\text{LL-Lin}\xspace}
\newcommand{\AccPI}{\text{Acc-PI}\xspace}
\newcommand{\AccMC}{\text{Acc-MC}\xspace}
\newcommand{\AccLin}{\text{Acc-Lin}\xspace}

This section presents our primary experimental results.
These are based on
\MNIST 
($D=784$, $\Ntrain=60$k, $\Ntest=10$k,
$C=10$ classes) \citep{lecun2010mnist}.
See \cref{sec:appx-expts} for more details
on these experiments, 
and more results on MNIST and other datasets.
We focus on training on a prefix of the first $T=2000$ examples from each dataset, since our main interest is in online learning from potentially nonstationary distributions, where rapid adaptation of a model in response to a small number of new data points 
is critical.

Our primary evaluation objective is the negative
log predictive density (NLPD)
of the  test set as a function of the number of  training points observed so far.\footnote{
We assume the training and test sets are drawn
from the same static distribution.
Alternatively, if there is only one stream of data
coming from a potential notstationary source,
we can use the prequential or one-step-ahead
log predictive density
\cite{Gama2013}.
We leave studying the non-stationary case
to future work.
}
It is defined as 
$    \NLPD_t = -\frac{1}{\Ntest}
    \sum_{i \in \Dtest}
    \log \left[
    \int p(\vy_i|f(\vx_i,\vtheta_t))
    q_{\vpsi_t}(\vtheta_t) {\rm d}\vtheta_t
    \right]
    $.
We approximate this integral in two main ways:
(1) using Monte Carlo sampling\footnote{
That is, we compute 
$S=100$ posterior samples
 $\vtheta_t^s \sim p(\vtheta_t|\Dtrain_{1:t})$
and then use
$p(\vy|\vx) \approx
\frac{1}{S} 
\sum_{s=1}^S p(\vy|\vx,\vtheta_t^s)$.
For efficiently sampling from a DLR Gaussian we follow \cite{SLANG}.
\kpm{Following
\cite{Immer21a}, we find it better to linearize
the model (i.e., replacing
 $h(\vx_i,\vtheta_t)$ with
 $\bar{h}(\vx_i, \vtheta_{t})$
 defined in \cref{eq:hbar})
 before pushing posterior samples through.
 We call this the Lin-MC approximation.
 }
 \label{ft:MC-predictive}
},%
or (2) using a plugin approximation,
where we replace the posterior $q_{\vpsi_t}(\vtheta_t)$
with a delta function centered at the mean,
$\delta(\vtheta_t - \vmu_t)$.

For methods that require a learning rate
(i.e., all methods except \bong),
we optimize it
wrt mid-way or
final performance on a holdout validation set,
using Bayesian optimization on \NLL.
All methods require specifying the prior
belief state, 
$p(\vtheta_0)=\gauss(\vmu_0, \vSigma_0=\sigma_0^2 \vI)$.
We optimize over $\sigma_0^2$ 
and sample $\vmu_0$ from a standard NN initializer.
As Hessian approximations, 
we  use \efmc with $\nsample=100$ samples,
 as well as the deterministic
 approximations \hesslin and \eflin.
(In the appendix, we also study \hessmc but find that it works very poorly, even with $\nsample=1000$.)

\eat{
We use $\niter=10$ iterations per step
for the iterative
methods (i.e., \blr, \bbb).
(In \cref{sec:iter}
we show that \blr performance deteriorates
significantly when using $\niter=1$,
whereas \bbb is robust to this parameter.)
}

\begin{figure}[hb!]
    \centering
       \includegraphics[width=.45\textwidth]{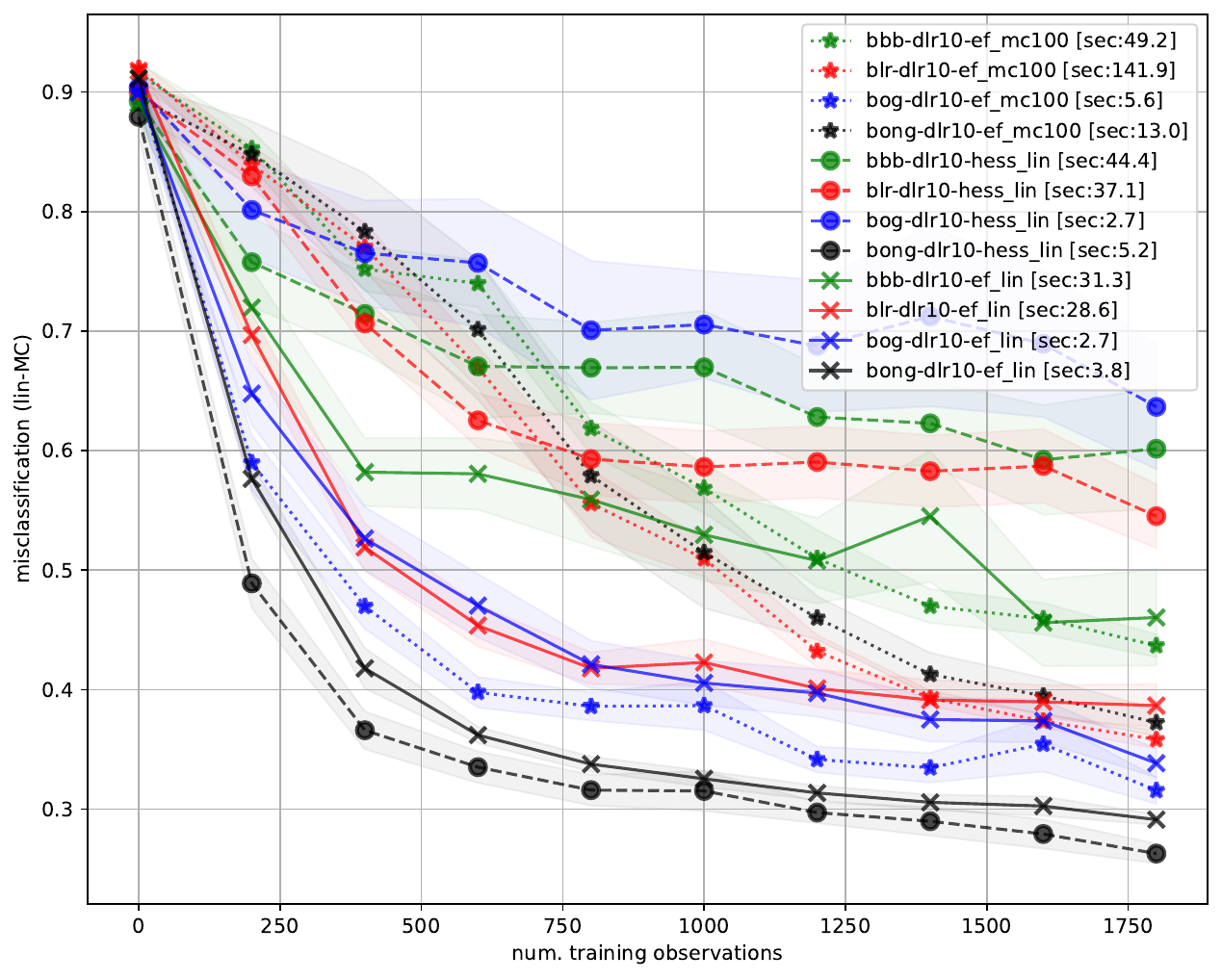}
    \includegraphics[width=.45\textwidth]{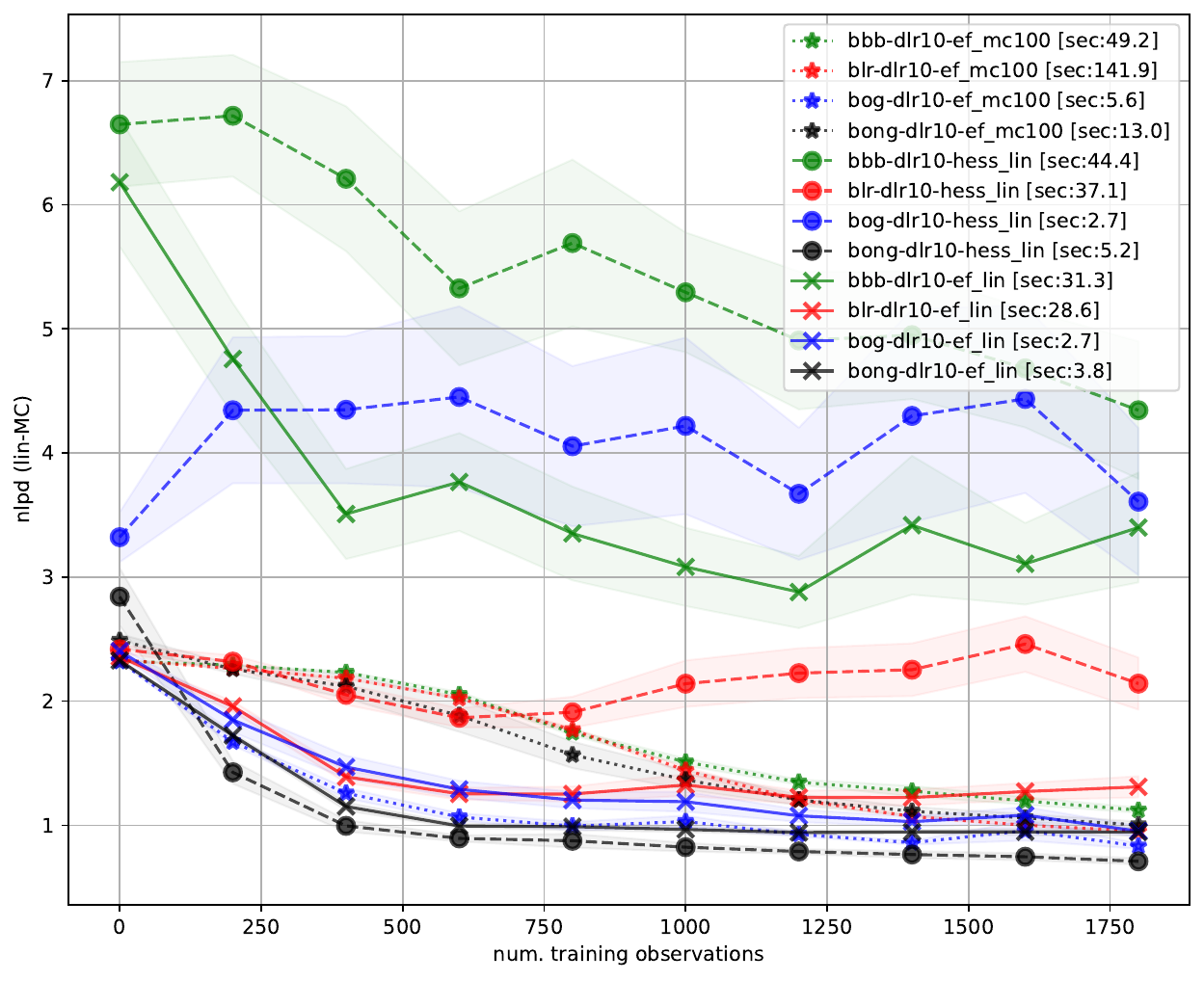}
    \caption{
        Performance on MNIST
    using Lin-MC 
    posterior predictive,
    where the posterior is 
    computed using \bong, \bog, \bbb and \blr 
    and the 3 tractable Hessian approximations
    with  \dlr-10 variational family.
}
    \label{fig:mnist-main-dlr}
\end{figure}

In \cref{fig:mnist-main-dlr} we compare the 4 main algorithms
using DLR family with rank $\rank=10$.
We apply these to
a CNN with two convolutional layers (each with 16 features and a (5,5) kernel), followed by two linear layers (one with 64 features and the final one with 10 features),
for a total of 57,722 parameters.
Shaded regions in these and all other plots indicate $\pm1$ SE, based on 5 independent trials randomly varying in the prior mean $\vmu_0$, data ordering, and MC sampling.
From this figure (and additional results in \cref{sec:appx-expts})
we conclude
\begin{itemize}[noitemsep,topsep=0pt,left=0pt]
    \item Linearization helps:
    \hesslin and \linef both
    outperform the MC variants.
    
    \item NGD helps: \bong outperforms \bog.
    
    \item Implicit regularization helps: \bong outperforms \blr.

    \item The \hesslin approximation outperforms \eflin, at least for \bong.

\item \kpm{\bbb generally does poorly.}

    \item \kpm{The \bong posterior predictive (using \hesslin) is slightly better  
    calibrated than  \bog, and both are much better than \blr and \bbb, at least for small sample sizes,
    as shown in \cref{fig:mnist-ece-dlr}.}

    \item \kpm{The plugin posterior predictive is similar to the Lin-MC predictive (see \cref{ft:MC-predictive}), and both are generally much better than the simple MC predictive, as shown in  \cref{fig:mnist-cnn}.}
\end{itemize}

\begin{figure}[hb!]
    \centering
        \includegraphics[width=.44\textwidth]{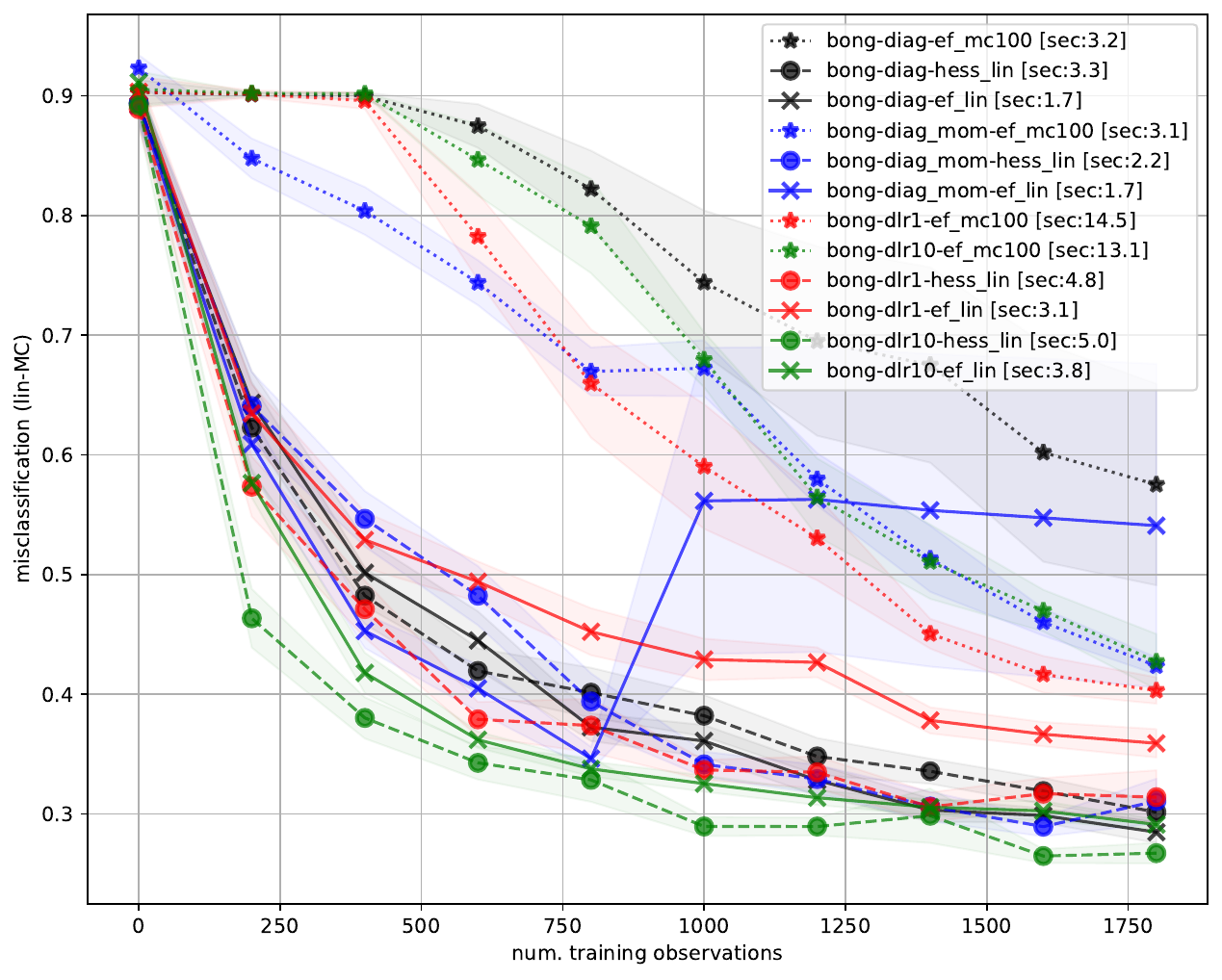}
    \includegraphics[width=.44\textwidth]{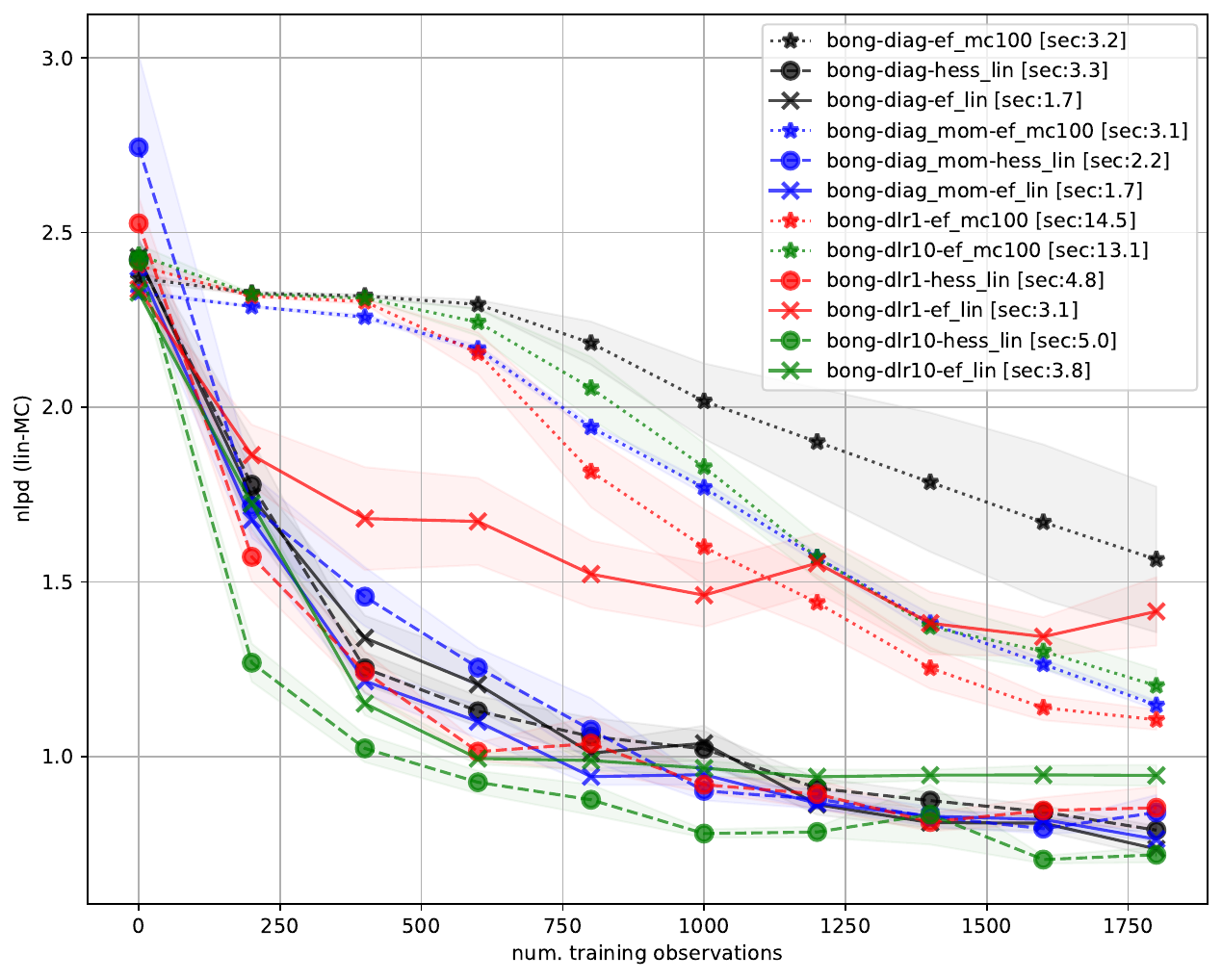}
    \caption{
    Performance on MNIST
    using Lin-MC 
    posterior predictive,
    where the posterior is 
    computed using \bong
    with different variational families,
    namely diagonal (natural and moment),
    \dlr-1, \dlr-10.
    }
    \label{fig:mnist-main-bong}
\end{figure}

\kpm{
In \cref{fig:mnist-main-bong} we 
compare \bong using different variational families, and conclude
}
\begin{itemize}[noitemsep,topsep=0pt,left=0pt]
    \item DLR-10 outperforms DLR-1, which is similar to diagonal (except when using \bong-\eflin, where DLR-1 is worse than diagonal). Also, we find (in results not reported here) that rank 5--10 often gives results as good as FC, but is much cheaper.

    \item Both natural and moment parameterizations for the diagonal representation perform comparably to each other,
    \kpm{although with \eflin, the moment parameterization can be numerically unstable}.\footnote{
    \kpm{
    This is likely due to the fact that
    in the moment update in \cref{eq:bong-diag-mom-var},
    we add $\diag(\vG_t)$ to the variance,
    whereas in the natural update in
    \cref{eq:bong-diag-prec2},
    we subtract $\diag(\vG_t)$ from the precision;
    the latter is more stable since $\vG_t$ is negative semi-definite for all 4 Hessian approximations.
    }
    }
\end{itemize}

\kpm{
Finally, in \cref{fig:mnist-main-runtime} we report the runtimes from these experiments and
conclude
}
\begin{itemize}[noitemsep,topsep=0pt,left=0pt]
   \item One-step methods (\bong and \bog) are faster than iterative methods (\blr and \bbb), as expected.

   \item Linearized methods (\hesslin and \eflin) are faster than MC methods (\efmc).

   \item EF methods (\eflin) are  a bit faster
   than methods that compute the Hessian exactly (\hesslin), especially for diagonal family. (This speedup is larger when the output dimensionality $C$ is big.)
\end{itemize}

%% file: sections/concl.tex
\section{Conclusions, limitations and future work}
\label{sec:conclusions}

Our experiment results show benefits of \bong's three main principles: 
NGD, 
implicit regularization to the prior,
and linearization.
The clear winner across datasets and variational families is \bong-\hesslin, which embodies all three principles.
\Blr-\hesslin nearly matches its performance but is much slower.
Several of the best-performing algorithms are previously known (notably \CMEKF and \LOFI) but we explain these results within a systematic theory that also offers new methods (including \blr-\hesslin).

\Bong is motivated by \cref{thm:exact-when-conjugate} which applies only in the idealized setting of a conjugate prior. Nevertheless we find it performs well in non-conjugate settings.
On the other hand our experiments are based on relatively small models and datasets. It will be important to test how our methods scale up, especially using the promising DLR representation.

%% file: sections/pseudocode.tex
\section{Abstract pseudocode}
\label{sec:pseudocode}

\Cref{algo:main,algo:predict,algo:update,algo:mc-hess-est,algo:mc-ef-est,algo:lin-hess-est,algo:lin-ef-est} give pseudocode for applying the methods we study.
For the predict step, we assume 
the dynamics model has the form
$\vtheta_t \sim \gauss(\vF_t \vtheta_{t-1} + \vb_t, \vQ_t)$.
The update step in \cref{algo:update} calls one of four grad functions (\cref{algo:lin-ef-est,algo:lin-hess-est,algo:mc-ef-est,algo:mc-hess-est}) that estimate the expected gradient and Hessian using either MC or linearization combined with either the direct Hessian (or Jacobian and observation covariance) or EF. 
The update step also calls an inner step function that implements \bong, \blr, \bog or \bbb on some variational family corresponding to the encoding of $\vpsi$ (not shown). 
In practice the grad-fn and inner-step-fn are not as cleanly separated because the full matrix $\bar{\vG}_{t,i}$ is not passed between them except when the variational family is FC. When the family is diagonal, grad-fn only needs to return $\diag(\bar{\vG}_{t,i})$. When grad-fn uses EF, it only needs to return $\gradmat_{t,i}$ (grad-\efmc) or $\gL_{t,i}$ (grad-\eflin) and inner-step-fn will implicitly use the outer product of this output as $\bar{\vG}_{t,1}$.
Finally, note the expressions for $\gL_{t,i}$ in \cref{algo:lin-ef-est,algo:lin-hess-est} are equivalent ways of computing the same quantity, as explained after \cref{eq:gL-cheap}.

\begin{algorithm}
\caption{Main  loop.}
\label{algo:main}
\For{$t=1:\infty$}
{
$\vpsi_{t|t-1} = \text{predict}(\vpsi_{t-1})$\\
$\vpsi_{t} = \text{update}(\vpsi_{t|t-1}, \vx_t, \vy_t)$\\
}
\end{algorithm}

\begin{algorithm}
\caption{Predict step.
}
\label{algo:predict}
$\text{def predict}(\vpsi_{t-1}=(\vmu_{t-1},\vSigma_{t-1}))$: \\
$\vmu_{t|t-1}=\vF_t \vmu_{t-1} + \vb_t$ \\
$\vSigma_{t|t-1}=\vF_t \vSigma_{t-1} \vF_t^\trans + \vQ_t$ \\
Return $\vpsi_{t|t-1}=(\vmu_{t|t-1}, \vSigma_{t|t-1})$
\end{algorithm}

\begin{algorithm}
\caption{Update step. The inner-step-fn is \bong, \blr, \bog or \bbb (not shown).
}
\label{algo:update}
$\text{def update}(\vpsi_{t|t-1}, \vx_t, \vy_t,
f(), A(), 
\text{grad-fn}, 
\text{inner-step-fn},
\alpha, \niter, \nsample)$: \\
$\vpsi_{t,0} = \vpsi_{t|t-1}$\\
$f_t(\vtheta_t)=f(\vx_t,\vtheta_t)$\\
$h_t(\vtheta_t)=E[\vy_t|\vx_t,\vtheta_t]
=\nabla_{\veta=f_t(\vtheta_t)} A(\veta)$ \\
$\vV_t(\vtheta_t)=\text{Cov}[\vy_t|\vx_t,\vtheta_t]
= \nabla^2_{\veta=f_t(\vtheta_t)} A(\veta)$\\
$\ell_t(\vtheta_t) = \log p(\vy_t|\vx_t, \vtheta_t)
=\log p(\vy_t|\vf_t(\vtheta_t))$\\
\For{$i=1:\niter$}
{
 $(\bar{\vg}_{t,i}, \bar{\vG}_{t,i}) 
 = \text{grad-fn}(\vpsi_{t,i-1}, \ell_t,  h_t, \vV_t,  \nsample)$ \\ 
$\vpsi_{t,i} = \text{inner-step-fn}(
\vpsi_{t|t-1}, \vpsi_{t,i-1}, \bar{\vg}_{t,i}, \bar{\vG}_{t,i},\alpha)$
}
Return $\vpsi_{t,\niter}$
\end{algorithm}

\begin{algorithm}
\caption{MC gradient/Hessian estimator}
\label{algo:mc-hess-est}
$\text{def grad-\hessmc}(\vpsi_{t,i-1}, 
 \ell_t, h_t=[], \vV_t=[],  \nsample)$: \\
\For{$m=1:\nsample$}
{
$\hat{\vtheta}_{t,i}^{(m)} \sim q_{\vpsi_{t,i-1}}(\vtheta)$ \\
$\hat{\vg}_{t,i}^{(m)}
    = \nabla_{\vtheta_{t}=\hat{\vtheta}_{t,i}^{(m)}} 
    \ell_t(\vtheta_t)$ \\
$\hat{\vG}_{t,i}^{(m)} = 
         \nabla^2_{\vtheta_{t} = \hat{\vtheta}_{t,i}^{(m)}}
         \ell_t(\vtheta_t)$\\
}
$\gMC_{t,i} 
    = \frac{1}{\nsample} \sum_{m=1}^\nsample \hat{\vg}_{t,i}^{(m)}$\\
$\GMCH_{t,i} = 
         \frac{1}{\nsample} \sum_{m=1}^\nsample \hat{\vG}_{t,i}^{(m)}$\\
Return $(\gMC_{t,i}, \GMCH_{t,i})$
\end{algorithm}

\begin{algorithm}
\caption{MC gradient/Hessian estimator with Empirical Fisher}
\label{algo:mc-ef-est}
$\text{def grad-\efmc}(\vpsi_{t,i-1}, 
 \ell_t, h_t=[], \vV_t=[],  \nsample)$: \\
\For{$m=1:\nsample$}
{
$\hat{\vtheta}_{t,i}^{(m)} \sim q_{\vpsi_{t,i-1}}(\vtheta)$ \\
$\hat{\vg}_{t,i}^{(m)}
    = \nabla_{\vtheta_{t}=\hat{\vtheta}_{t,i}^{(m)}} 
    \ell_t(\vtheta_t)$ \\
}
$\gMC_{t,i} 
    = \frac{1}{\nsample} \sum_{m=1}^\nsample \hat{\vg}_{t,i}^{(m)}$\\
$\gradmat_{t,i} = [\hat{\vg}_{t,i}^{(1)},\dots,\hat{\vg}_{t,i}^{(\nsample)}]$ \\
 $\GMCEF_{t,i} =  -\frac{1}{\nsample} \gradmat_{t,i} \gradmatT_{t,i}$ \\
Return $(\gMC_{t,i},  \GMCEF_{t,i})$
\end{algorithm}

\begin{algorithm}
\caption{Linearized gradient/Hessian estimator}
\label{algo:lin-hess-est}
$\text{def grad-\hesslin}(\vpsi_{t,i-1}, 
 \ell_t=[], h_t, \vV_t, \nsample=[])$: \\
 $\vmu_{t,i-1} = E[\vtheta_t|\vpsi_{t,i-1}]$\\
$\hat{\vy}_{t,i} 
= h_t(\vmu_{t,i-1})$ \\
$\vH_{t,i}
= \rev{\frac{\partial h_t}{\partial\vtheta_t}_{\vert\vtheta=\vmu_{t,i-1}}}$\\
$\vR_{t,i}  = \vV_t(\vmu_{t,i-1}) $\\
$\gL_{t,i} = \vH^\trans_{t,i} \vR^{-1}_{t,i} (\vy_t - \hat{\vy}_{t,i})$
\\
$\GLH_{t,i} = - \vH^\trans_{t,i} \vR^{-1}_{t,i} \vH_{t,i}$\\
Return $(\gL_{t,i},  \GLH_{t,i})$  
\end{algorithm}

\begin{algorithm}
\caption{Linearized gradient/Hessian estimator with empirical Fisher}
\label{algo:lin-ef-est}
$\text{def grad-\eflin}(\vpsi_{t,i-1}, 
 \ell_t=[], h_t, \vV_t, \nsample=[])$: \\
$\vmu_{t,i-1} = E[\vtheta_t|\vpsi_{t,i-1}]$\\
$\vR_{t,i}  = \vV_t(\vmu_{t,i-1}) $\\
$\gL_{t,i} = \nabla_{\vtheta_t = \vmu_{t,i-1}} 
        \left[ -\tfrac{1}{2}
            \left(\vy_t - h_t(\vtheta_t)\right)^\trans \vR_{t,i}^{-1} (\vy_t - h_t(\vtheta_t))
        \right]$ \\
$\GLEF_{t,i} = - \gL_{t,i} \left(\gL_{t,i}\right)^\trans$ \\
Return $(\gL_{t,i},  \GLEF_{t,i})$  
\end{algorithm}

%% file: sections/appx-experiments.tex
\section{Additional experiment results}
\label{sec:appx-expts}

In this section, we give a more thorough set of experimental results.

 \subsection{Running time measures}
\label{sec:speed}

\begin{figure}[h!]
    \centering
    \includegraphics[width=.48\textwidth,valign=t]{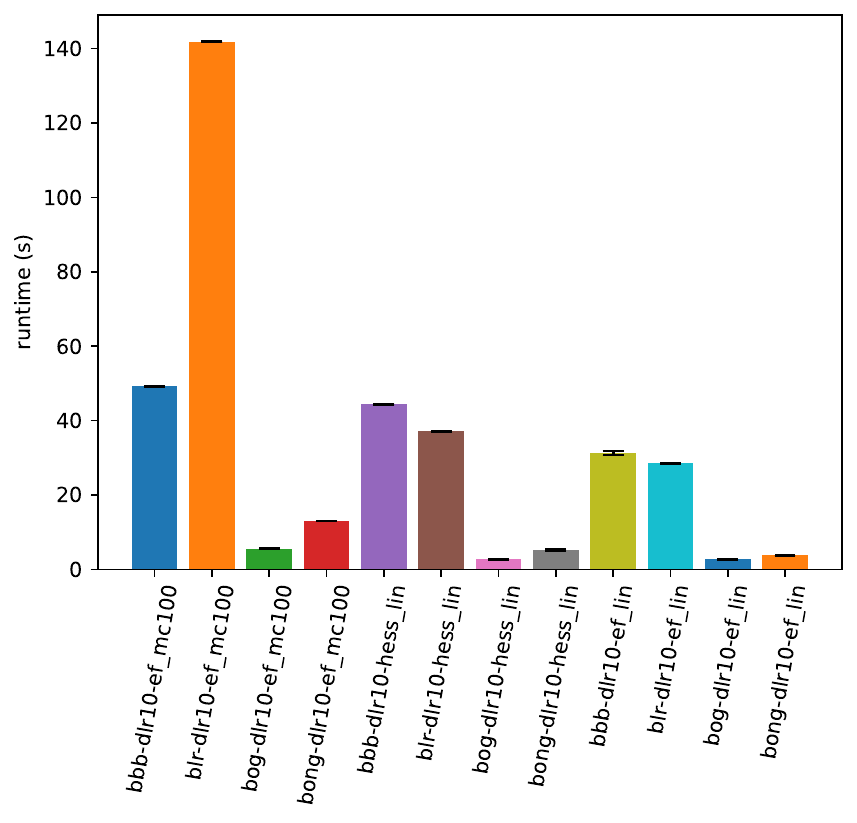}
    \includegraphics[width=.48\textwidth,valign=t]{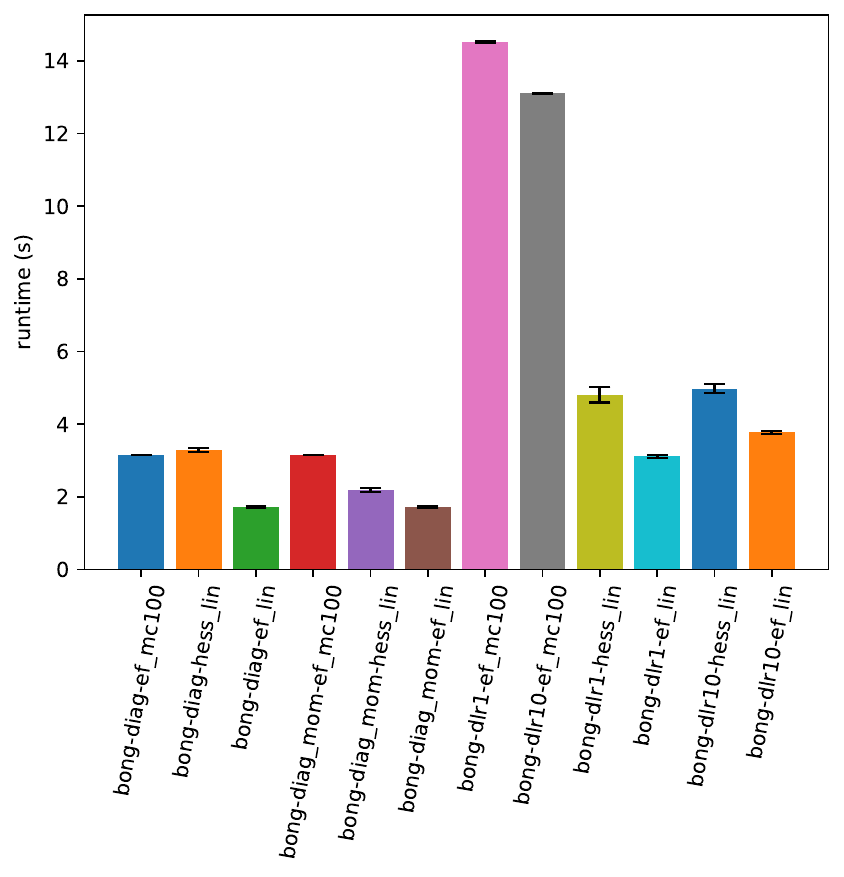}
    \caption{Runtimes for methods on MNIST.
    Left: Corresponding to \cref{fig:mnist-main-dlr} using different algorithms
    on \dlr-10 family.
    Right: Corresponding to 
    \cref{fig:mnist-main-bong},
    using \bong on different variational families.
    }
    \label{fig:mnist-main-runtime}
\end{figure}

The running times of the methods 
for the experiments in 
\cref{fig:mnist-main-dlr,fig:mnist-main-bong},
where we fit a CNN to MNIST, are shown in 
\cref{fig:mnist-main-runtime}.

\begin{figure}[h!]
    \centering
    \begin{tabular}{cc}
    \includegraphics[width=.48\textwidth]{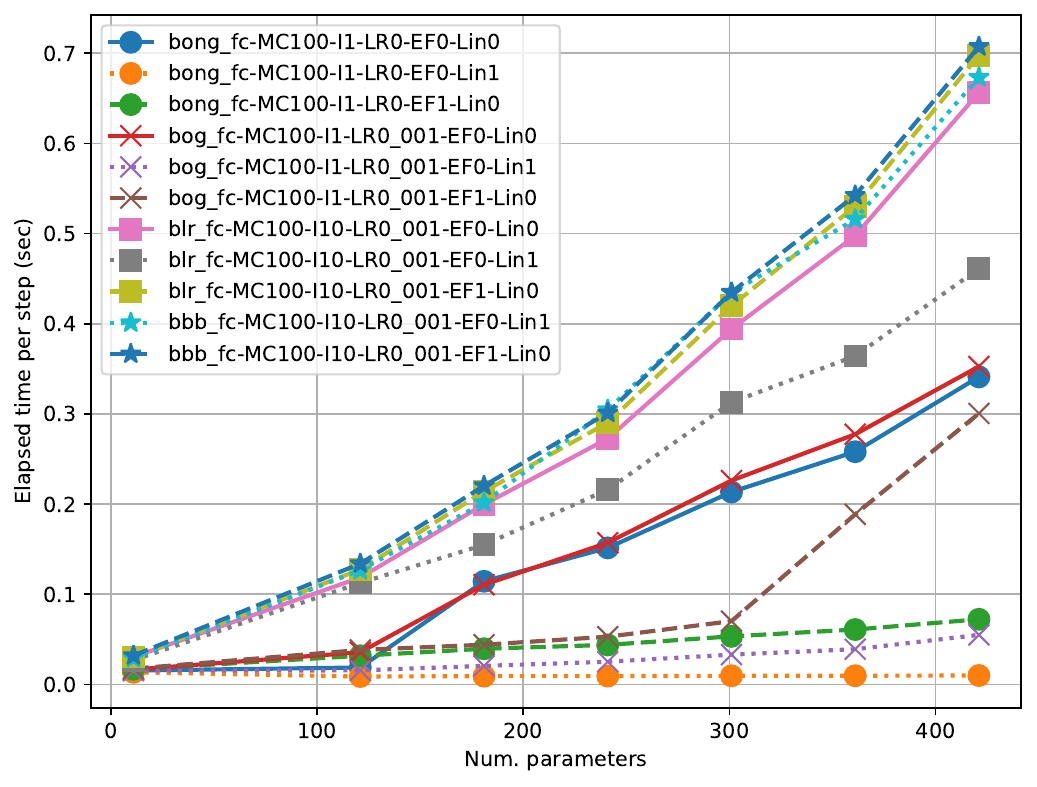}
    &
     \includegraphics[width=.48\textwidth]{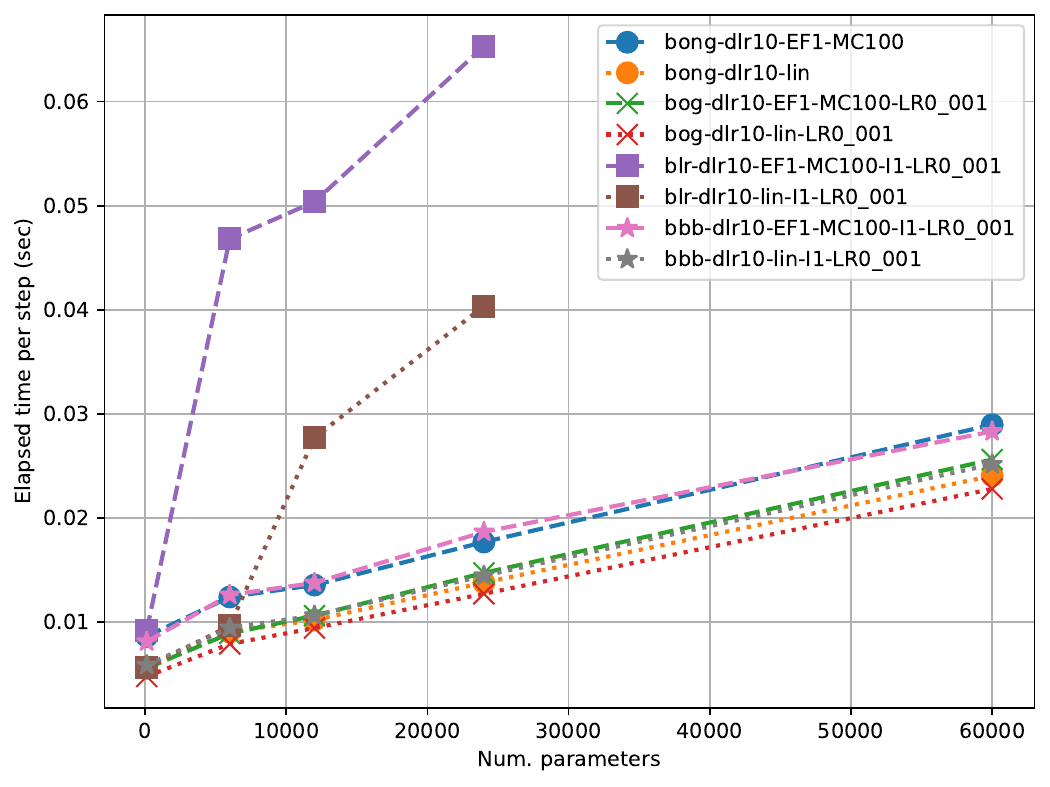}
     \\
     (a) & (b)
     \end{tabular}
    \caption{Running time (seconds)
    vs number of parameters $\nparam$
    (size of state space) on a synthetic regression problem.
For \bbb and \blr, we show results using $I=1$ and $I=10$ iterations per step.
Hessian approximations are denoted as follows:
EF0-Lin0 = MC-Hess,
EF1-Lin0 = EF-Hess,
EF0-Lin1 = Lin-Hess.
    (a) Full Covariance representation.
    (b) DLR representation.
    The BLR plot is truncated due to out of memory problem.
    }
    \label{fig:speed}
\end{figure}

The running times of the methods for the FC and DLR case,
where we fit an MLP to  a synthetic regression dataset,
are shown in \cref{fig:speed}.
The slower speed of \blr (even with $I=1$) relative to \bong
is at least partly attributable to the fact that
\blr must compute the SVD of a larger matrix
(see \cref{sec:BLR-DLR,sec:LBLR-DLR}).

\subsection{Detailed results for CNN on MNIST}
\label{sec:mnist-appx}

\begin{figure}[!th]
    \centering
    \includegraphics[width=0.4\textwidth]{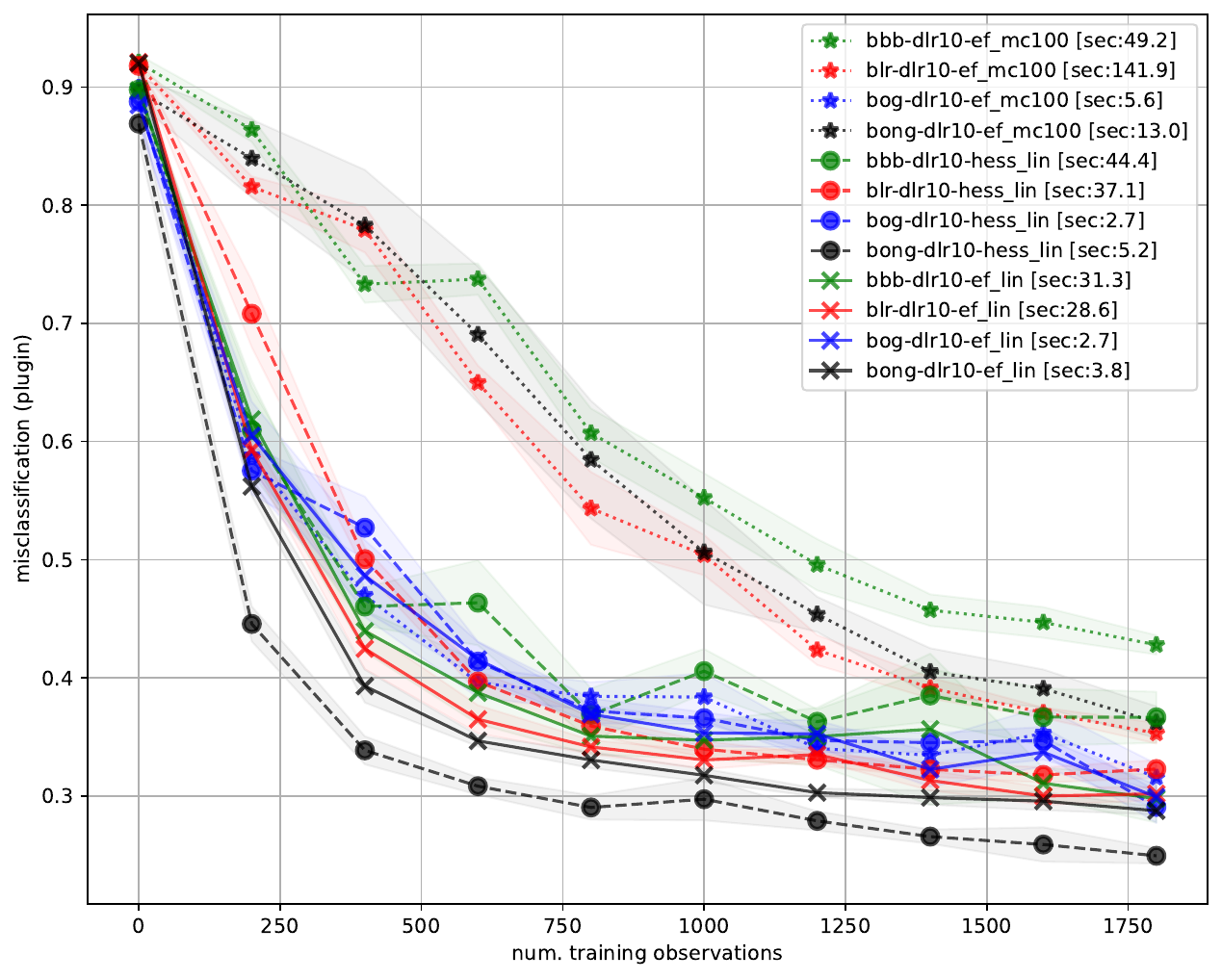}
    \includegraphics[width=0.4\textwidth]{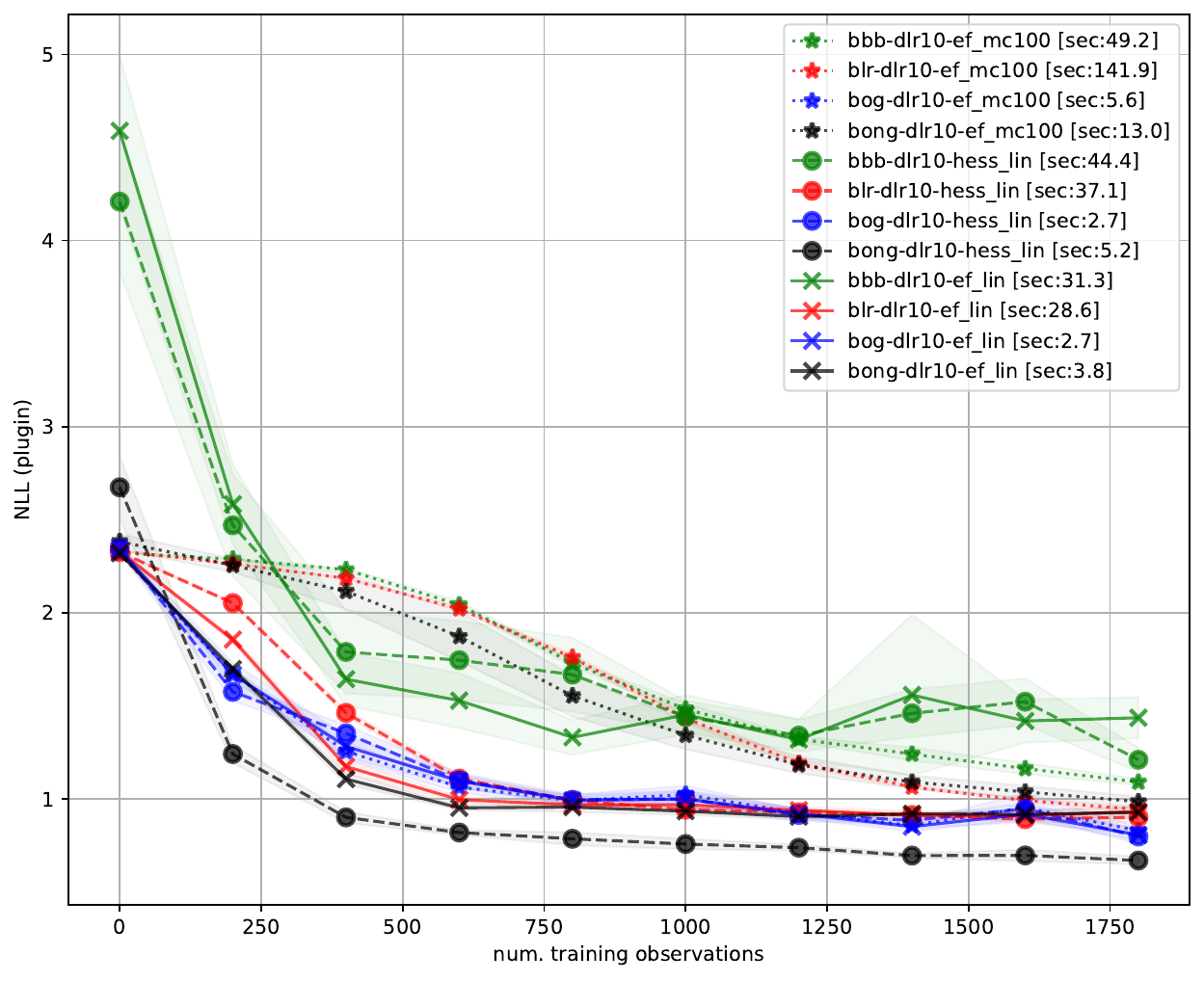}
    \\
    \medskip
    \includegraphics[width=0.4\textwidth]{figures/mnist-cnn/linmc_miscl_all.pdf}
    \includegraphics[width=0.4\textwidth]{figures/mnist-cnn/linmc_nlpd_all.pdf}
    \\
    \medskip
    \includegraphics[width=0.4\textwidth]{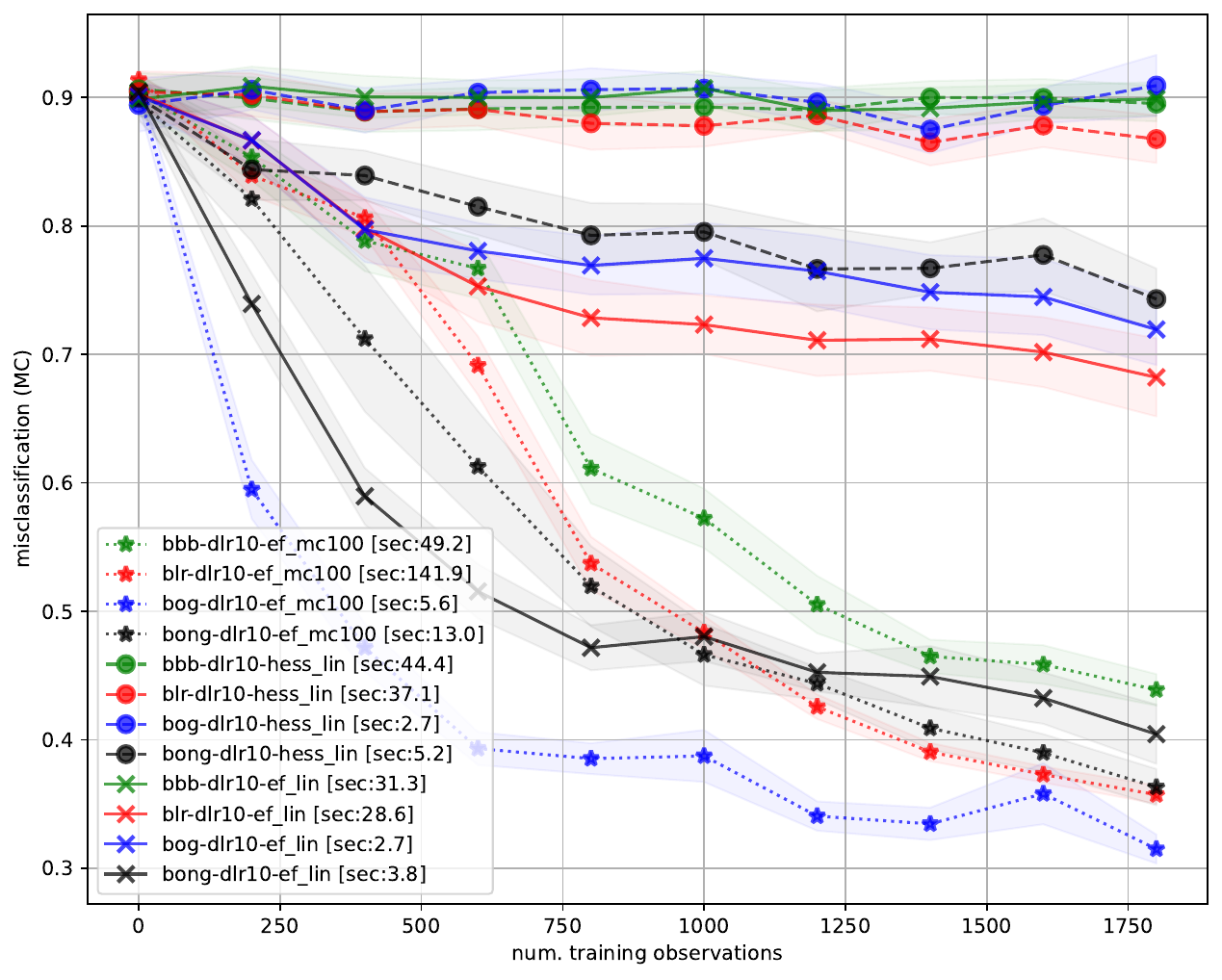}
    \includegraphics[width=0.4\textwidth]{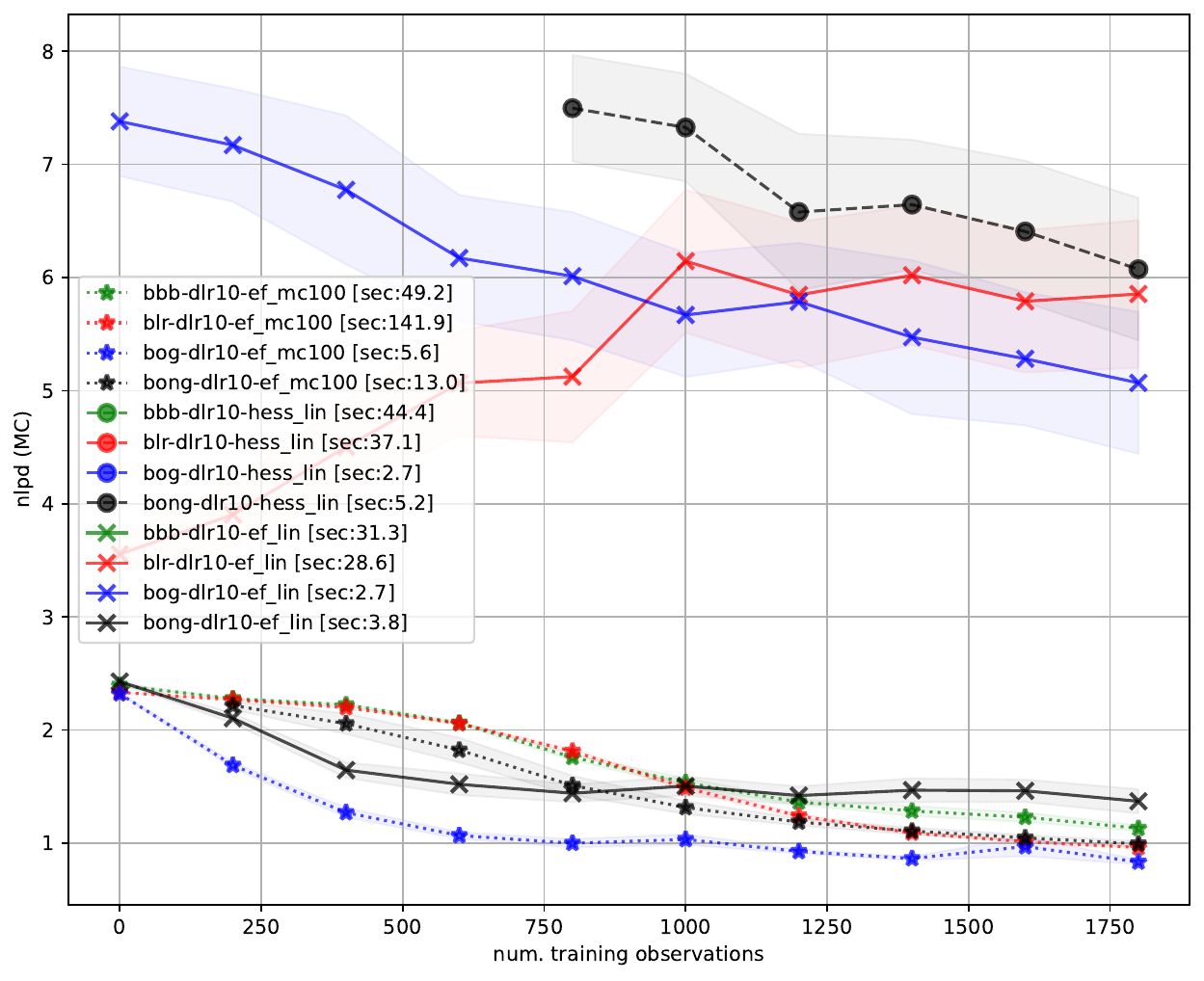}
    \\
    \includegraphics[width=0.4\textwidth]{figures/mnist-cnn/runtime_all.pdf}
    \caption{MNIST results for methods using DLR family.
    \kpm{
    Left column shows misclassification rate, right column showns NLL.
    First row uses plugin approximation to the posterior predictive,
    second row uses linearized MC approximation,
    and third row uses standard MC approximation.
    }
    }
    \label{fig:mnist-cnn}
\end{figure}

\begin{figure}[!th]
    \centering
    \includegraphics[width=0.4\textwidth]{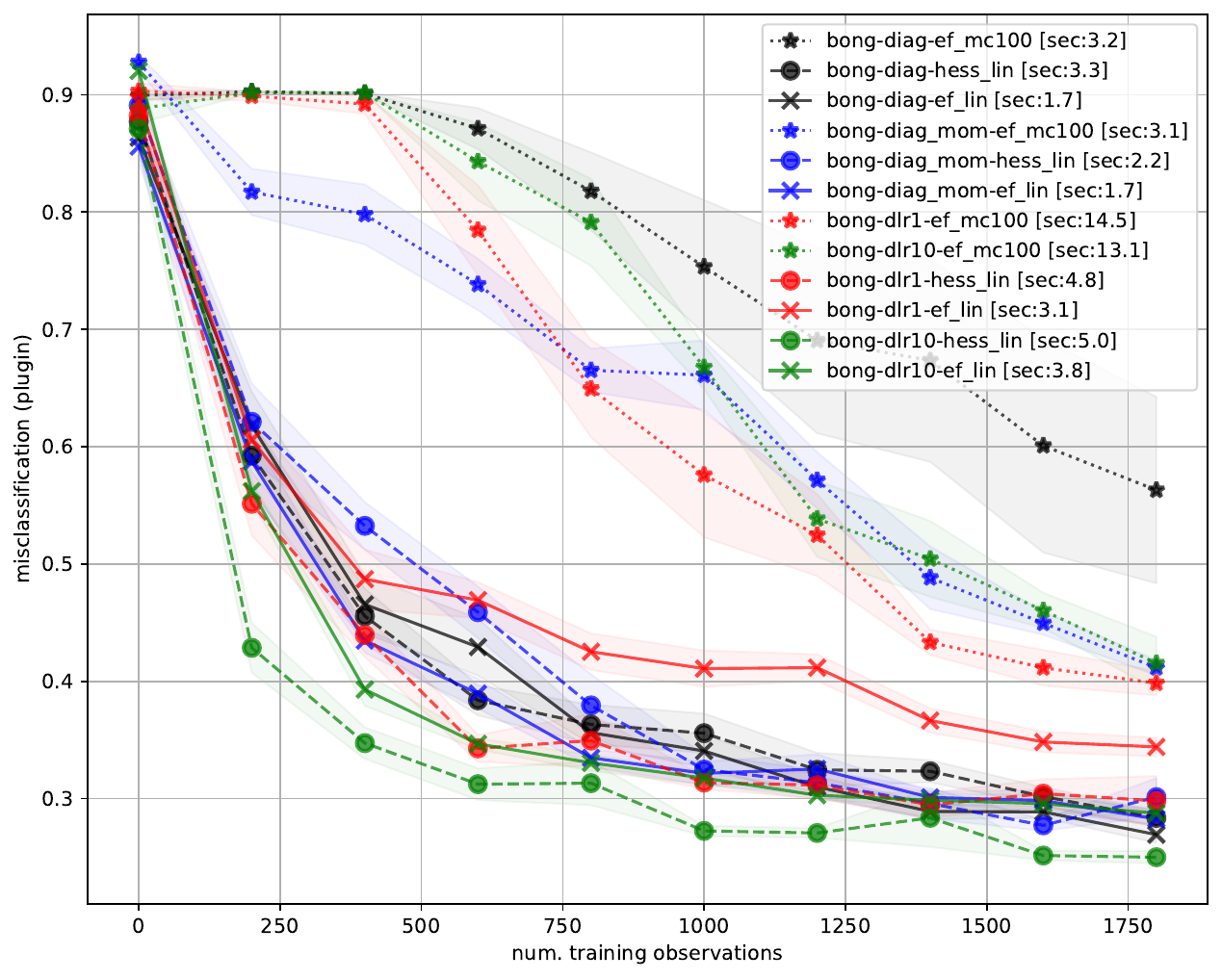}
    \includegraphics[width=0.4\textwidth]{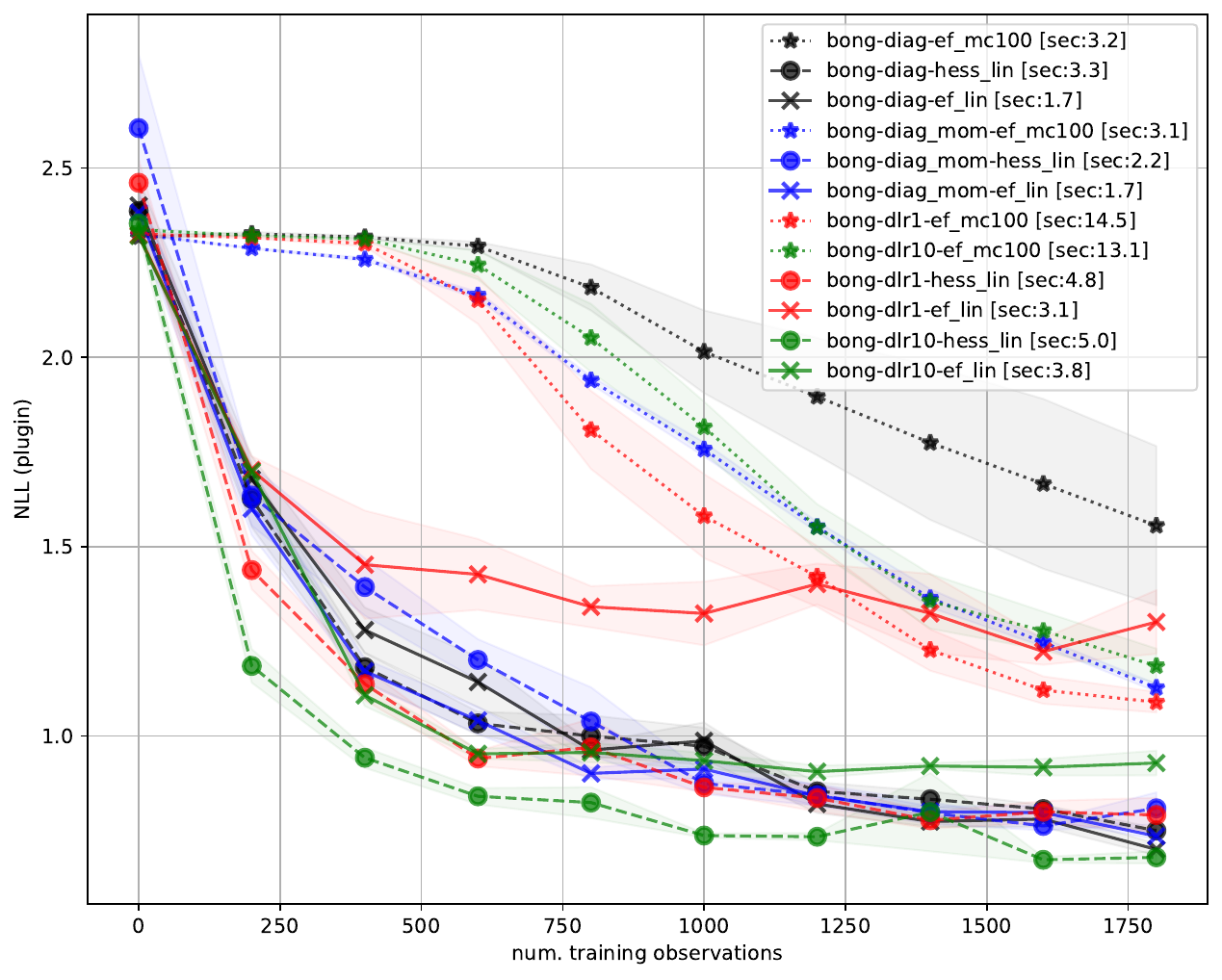}
    \\
    \medskip
    \includegraphics[width=0.4\textwidth]{figures/mnist-variants/linmc_miscl_bongs.pdf}
    \includegraphics[width=0.4\textwidth]{figures/mnist-variants/linmc_nlpd_bongs.pdf}
    \\
    \medskip
    \includegraphics[width=0.4\textwidth]{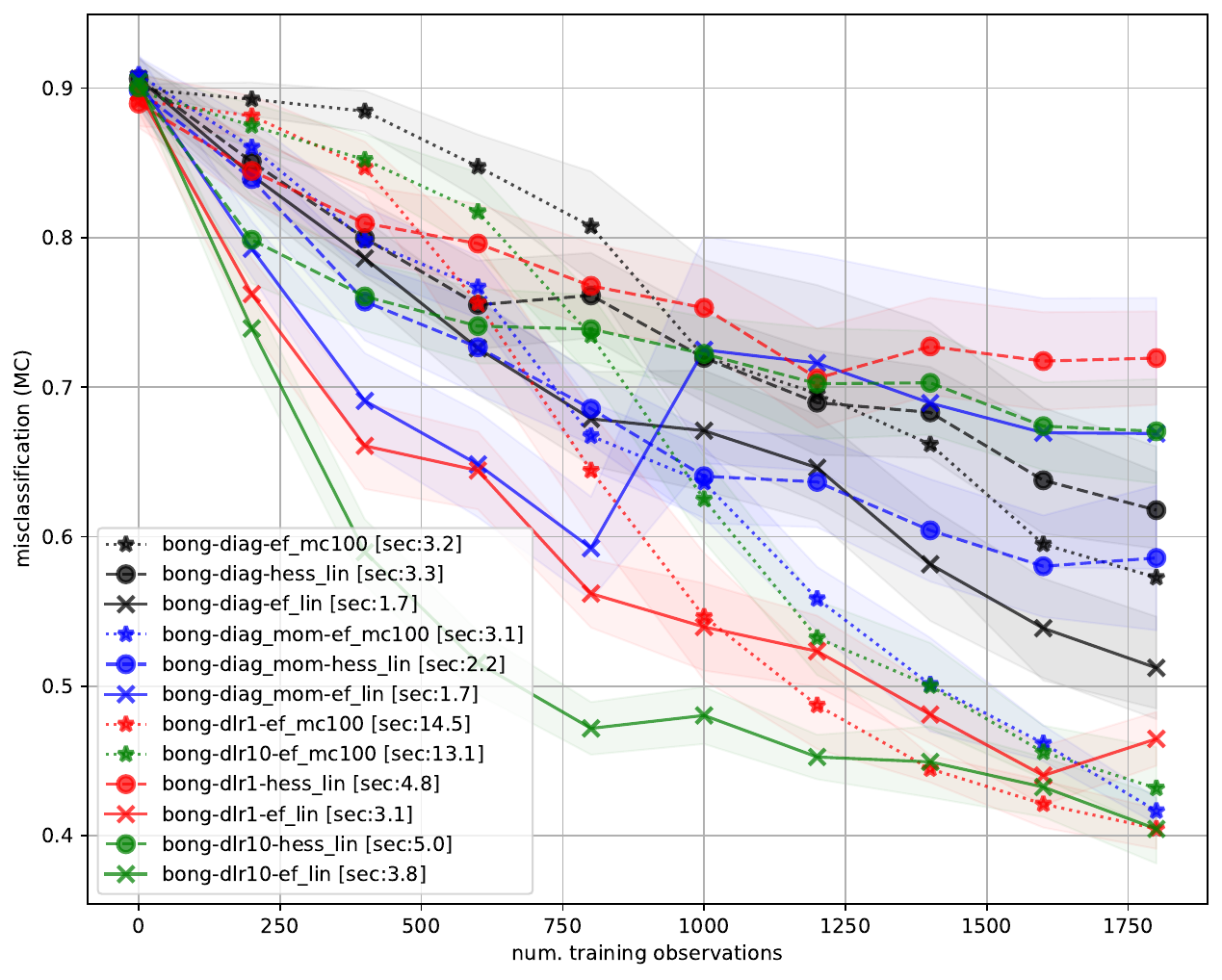}
    \includegraphics[width=0.4\textwidth]{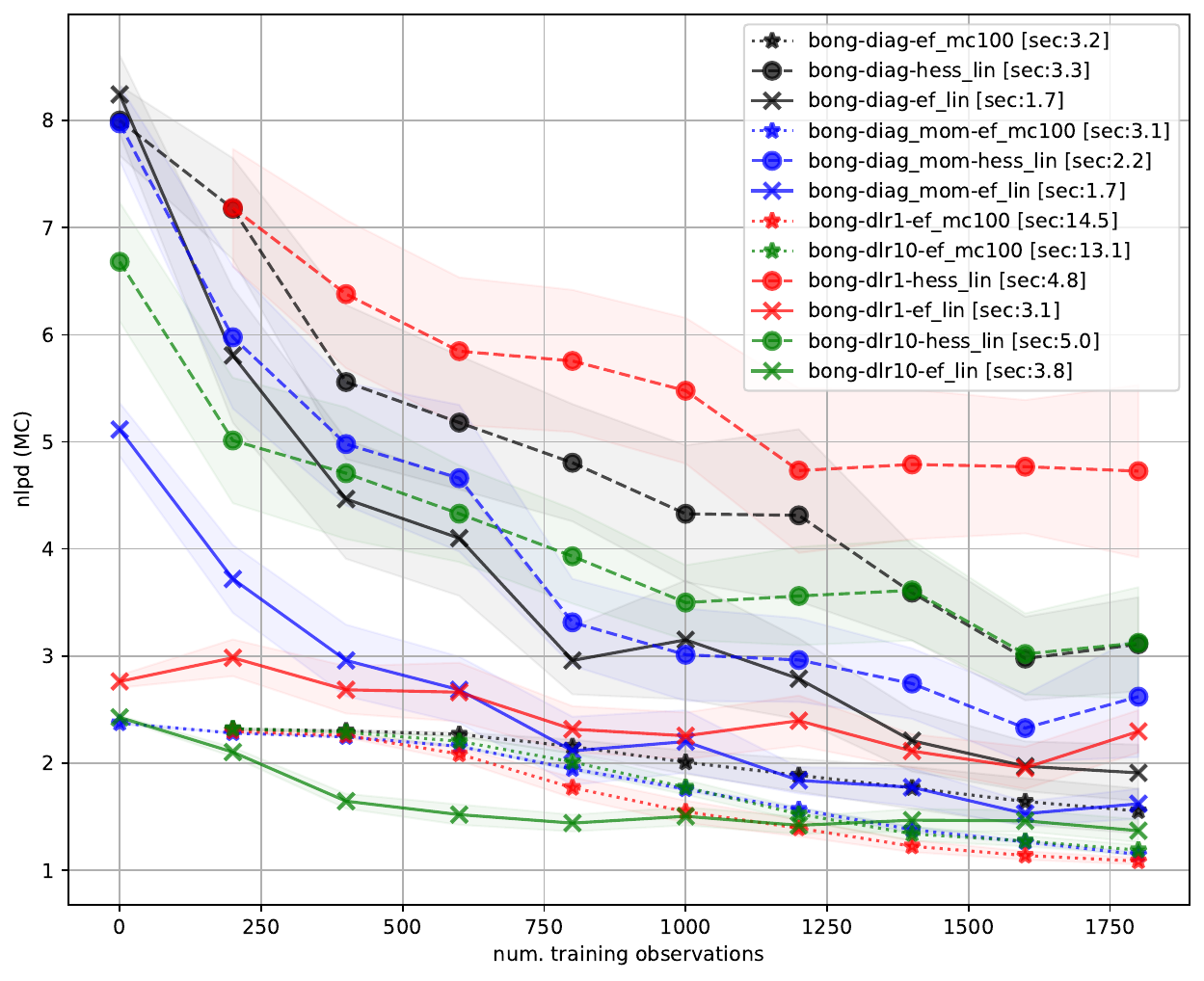}
    \\
    \includegraphics[width=0.4\textwidth]{figures/mnist-variants/runtime_bongs.pdf}
    \caption{MNIST results for \bong variants.
     \kpm{
    Left column shows misclassification rate, right column shows NLL.
    First row uses plugin approximation to the posterior predictive,
    second row uses linearized MC approximation,
    and third row uses standard MC approximation.
    }
    }
    \label{fig:mnist-bong}
\end{figure}

Here we report further metrics for the experiments in \cref{fig:mnist-main-dlr,fig:mnist-main-bong}.
We show 3 approximations to the NLPD:
plugin, MC, and Linearized MC.\footnote{
%
Lin-MC is  defined in  \cref{ft:MC-predictive}.
The motivation for this  approximation 
(from \cite{Immer21a})
is the following:
If we push posterior samples
through a nonlinear predictive model,
the results can be poor if the samples are far from the mean,
but if we linearize the predictive model,
extrapolations away from the mean are more sensible.
This is true even if the posterior was
not explicitly computed using 
a linear approximation.
}
For each of these  
approximations to the posterior predictive,
we also measure the corresponding
misclassification rate based on picking the most probable predicted class.
Results are shown in \cref{fig:mnist-cnn,fig:mnist-bong}.
    We see that the plugin and lin-MC approximations are similar, and both are generally much better than standard MC.
    
\begin{figure}[!th]
    \centering
    \includegraphics[width=\textwidth]{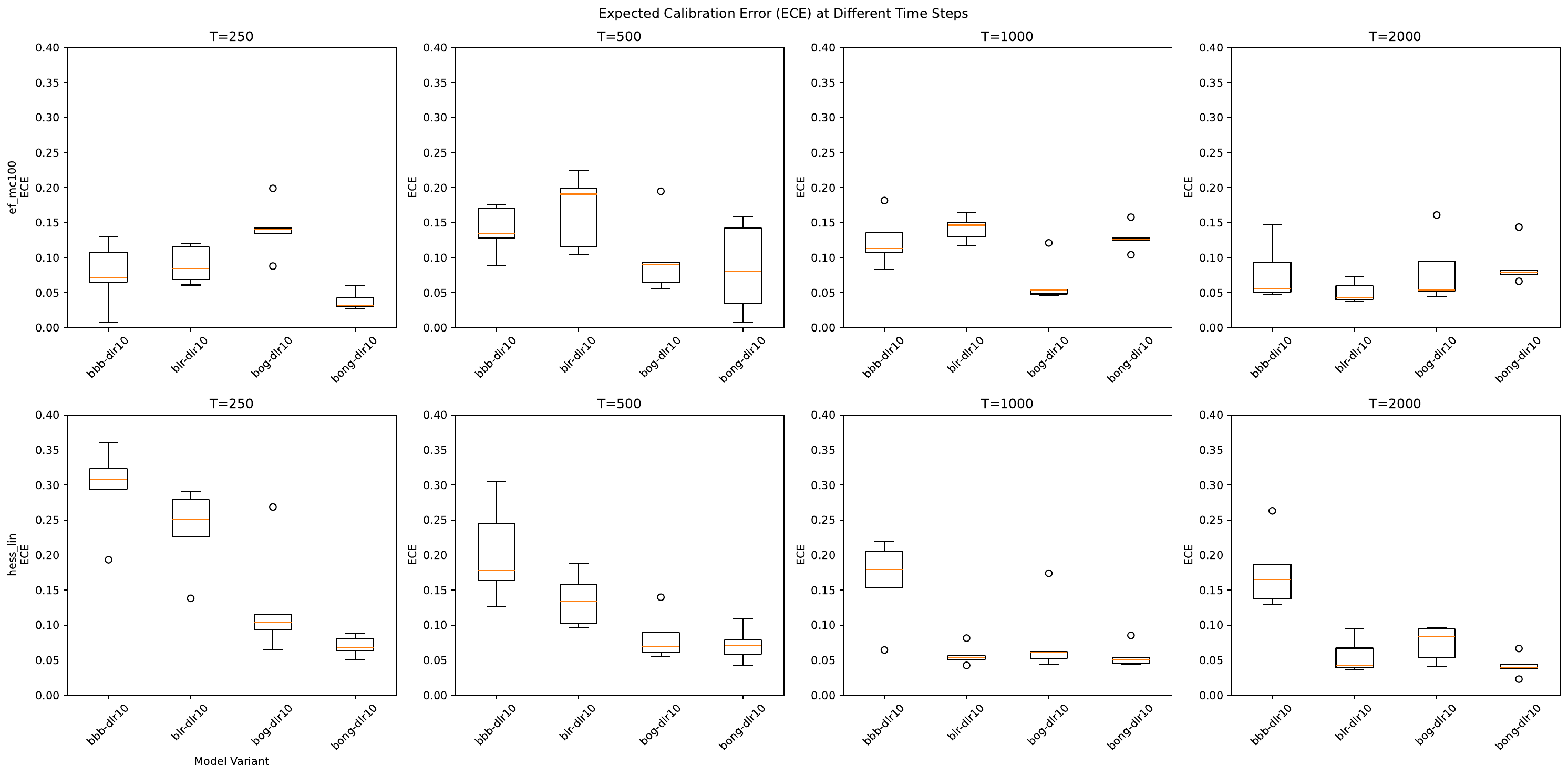}
    \caption{MNIST expected calibration error (ECE) results at selected timesteps for methods using DLR family.}
    \label{fig:mnist-ece-dlr}
\end{figure}

\begin{figure}[!th]
    \centering
    \includegraphics[width=\textwidth]{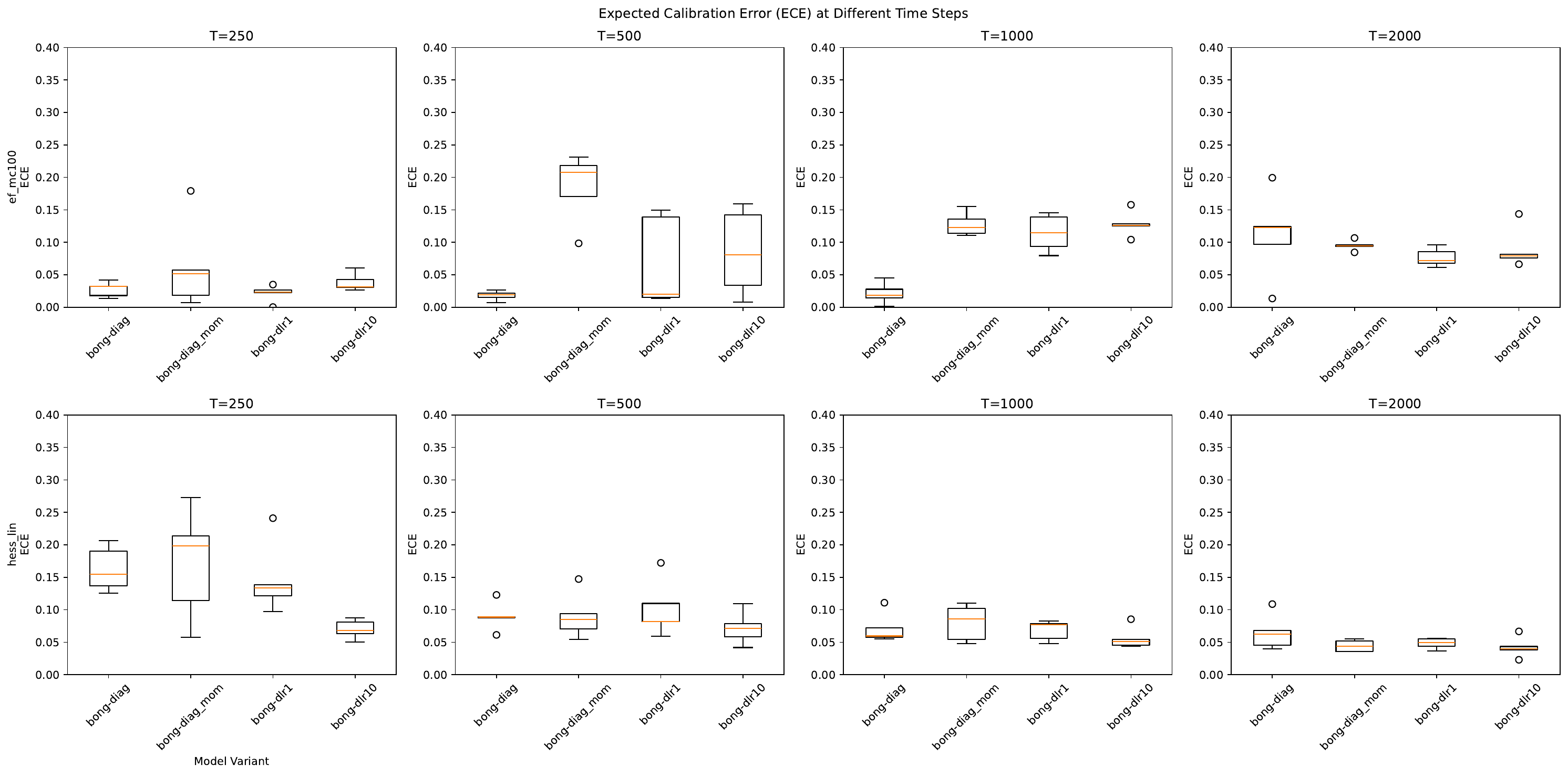}
    \caption{MNIST expected calibration error (ECE) results at selected timesteps for \bong variants.
    \kpm{
    We see that \bong and \bog are both well calibrated,
    as least when combined with \hesslin,
    with \bong have a slight edge, especially
    for small sample sizes, where there is more posterior uncertainty.
    }
    }
    
    \label{fig:mnist-ece-bong}
\end{figure}

Finally, in \cref{fig:mnist-ece-dlr} and \cref{fig:mnist-ece-bong}, we report the 
test-set expected calibration error (ECE) at time steps [250, 500, 1,000, 2,000], 
computed using $20$ bins. Note that among the \linhess variants, \bong-\dlr-10 
method is the most well-calibrated (in addition to exhibiting the strongest 
plugin and linearized-MC NLPD results), when compared to other \dlr-10 methods as well
as other \bong variants.


\subsection{\SARCOS dataset}

In addition to \MNIST, we report experiments on the \SARCOS regression dataset ($D=22$, $\Ntrain=44,\!484$,
$\Ntest=4449$, $C=1$).
This dataset derives 
from   an inverse dynamics problem for a seven degrees-of-freedom SARCOS anthropomorphic robot arm.
 The task is to map from a 21-dimensional input space (7 joint positions, 7 joint velocities, 7 joint accelerations) to the corresponding 7 joint torques.
 Following
 \cite{Rasmussen06}, 
 we pick a single target output dimension,
 so $C=1$.
The data is from
 \url{https://gaussianprocess.org/gpml/data/}.

We use a small MLP of size 21-20-20-1,
so there are $\nparams=881$ parameters.
For optimizing learning rates for \SARCOS, we use grid search on \NLPD-PI.
We fix the variance of the prior belief state to $\sigma_0^2=1.0$, which represents
a mild degree of regularization.\footnote{
This value was based on a small
amount of trial and error.
Using a smaller value of
$\sigma_0$
results in underfitting relative to a linear least squares baseline,
and using a much larger value
results in unstable posterior covariances,
causing the NLPD-MC samples to
result in NaNs after a few hundred steps.
}
We fix the observation variance to
$R_t = 0.1 \hat{R}$,
where $\hat{R}=\text{Var}(y_{1:T})$
is the maximum likelihood estimate
based on the training sequence;
we can think of this as a simple empirical Bayes approximation, and the factor of 0.1
accounts for the fact that the  variance of the residuals from the learned model will be smaller than from the unconditional baseline.
We focus on DLR approximation of rank 10.
This gives similar results to full covariance, but is much faster to compute.
We also focus on the plugin approximation to NLPD,
since the MC approximation gives much worse results (not shown).

\subsubsection{Comparison of \bong, \blr, \bbb and \bog}

\begin{figure}[h!]
    \begin{subfigure}[b]{0.49\textwidth}
    \centering
  \includegraphics[width=\textwidth]{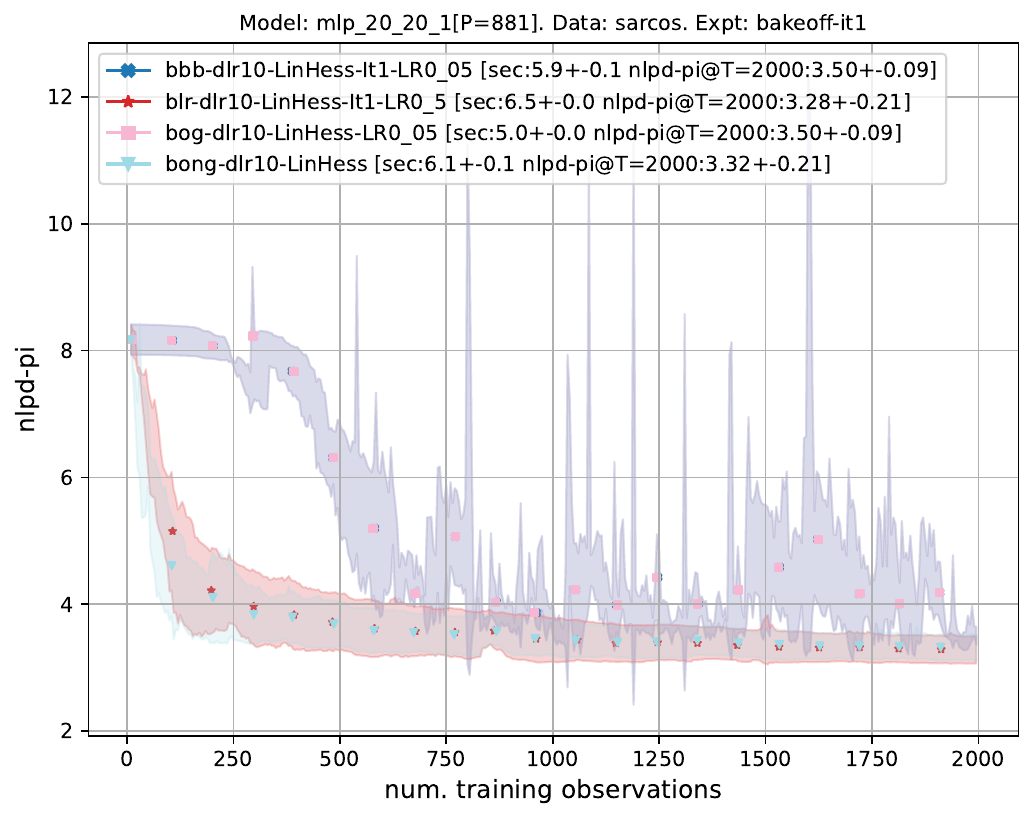}
    \caption{1 iteration.}
    \label{fig:lin-it1}
    \end{subfigure}
    \hfill
\begin{subfigure}[b]{0.49\textwidth}
    \centering
    \includegraphics[width=\textwidth]
    {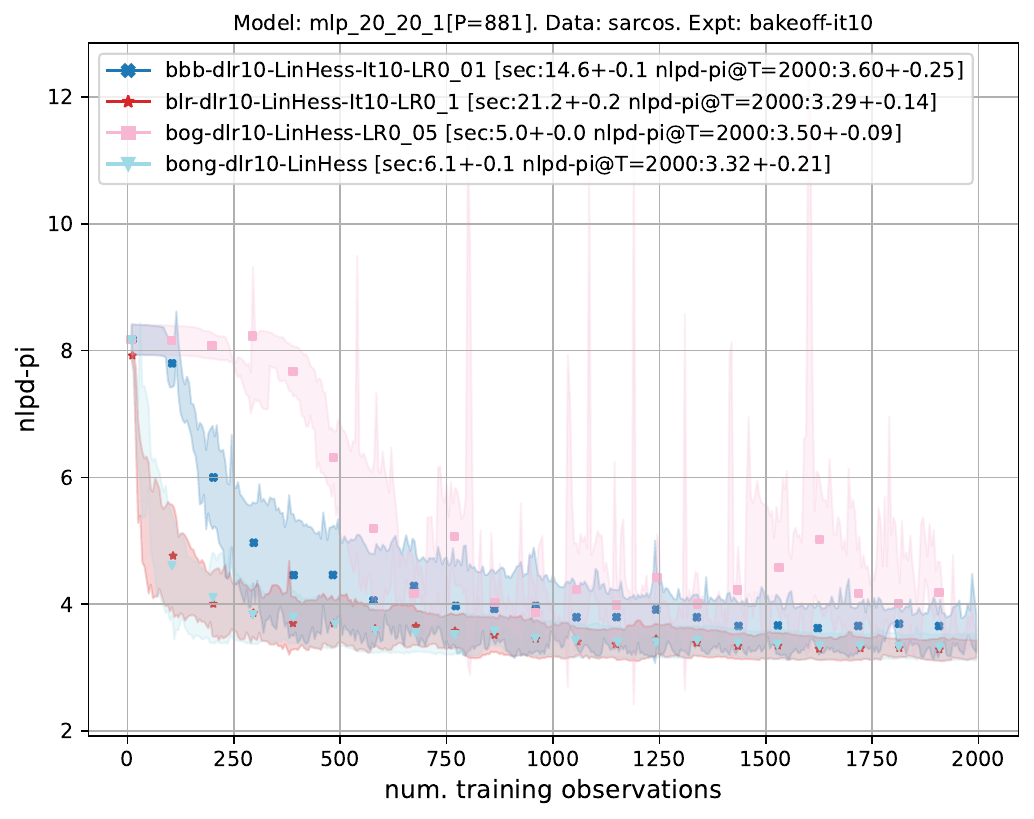}
    \caption{10 iterations.}
    \label{fig:lin-it10}
    \end{subfigure}
    \caption{
    Predictive performance
    on \SARCOS using
    MLP 21-20-20-1 with DLR rank 10.
    Error bars represent $\pm 1$ standard deviation
    computed from 3 random trials,
    randomizing over data order and
    initial state $\vmu_0$.
    (a) We show all  4 algorithms
    combined with \linhess approximation 
    and $\niter=1$.
    (b) Same as (a) but with $\niter=10$.
    }
    \label{fig:sarcos-lin}
\end{figure}

In \cref{fig:sarcos-lin} we show the results of using the \linhess approximation.
For 1 iteration per step,
we see that \bong and \blr are indistinguishable
in performance,
and \bbb and \bog are also  indistinguishable,
but much worse.
For 10 iterations per step,
we see that \bbb improves significantly,
and approaches \bong and \blr.
However, \blr and \bbb are now
about 10 times slower.
(In practice, the slowdown is less than 10,
due to constant factors of the implementation.)
(Note that \bong and \bog always use a single
iteration, so their performance does not change.)

\begin{figure}[h!]
    \begin{subfigure}[b]{0.49\textwidth}
    \centering
  \includegraphics[width=\textwidth]{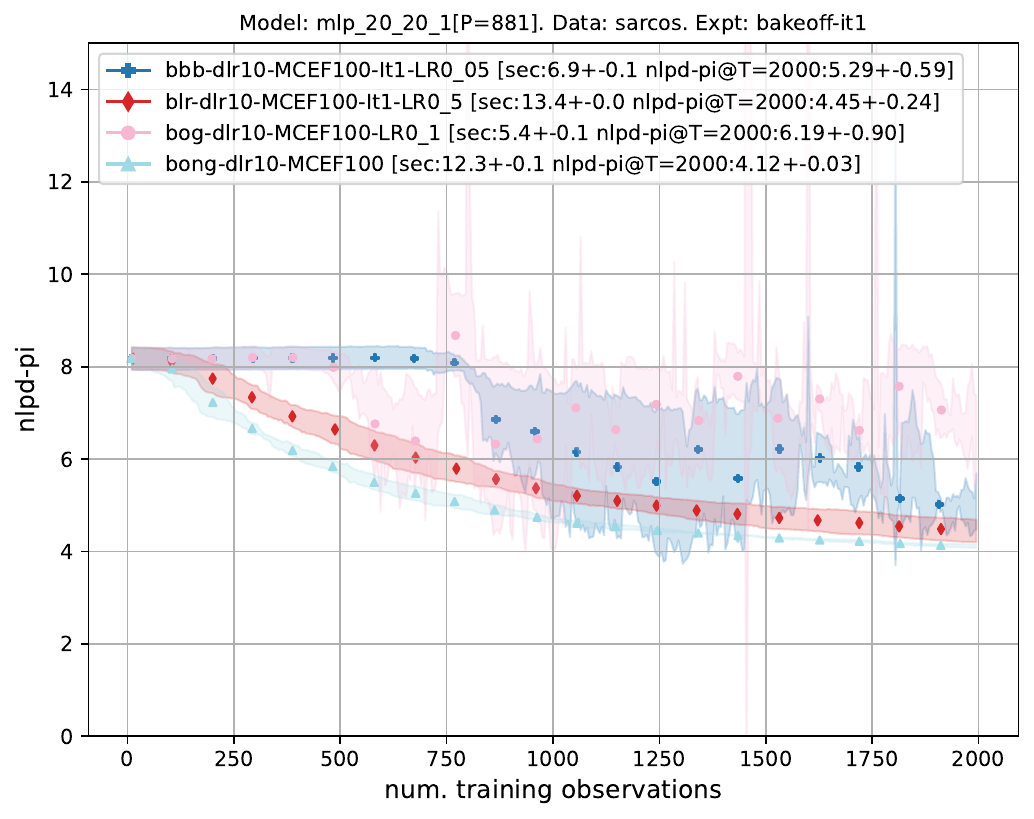}
    \caption{1 iteration.}
    \label{fig:ef-it1}
    \end{subfigure}
    \hfill
\begin{subfigure}[b]{0.49\textwidth}
    \centering
    \includegraphics[width=\textwidth]
    {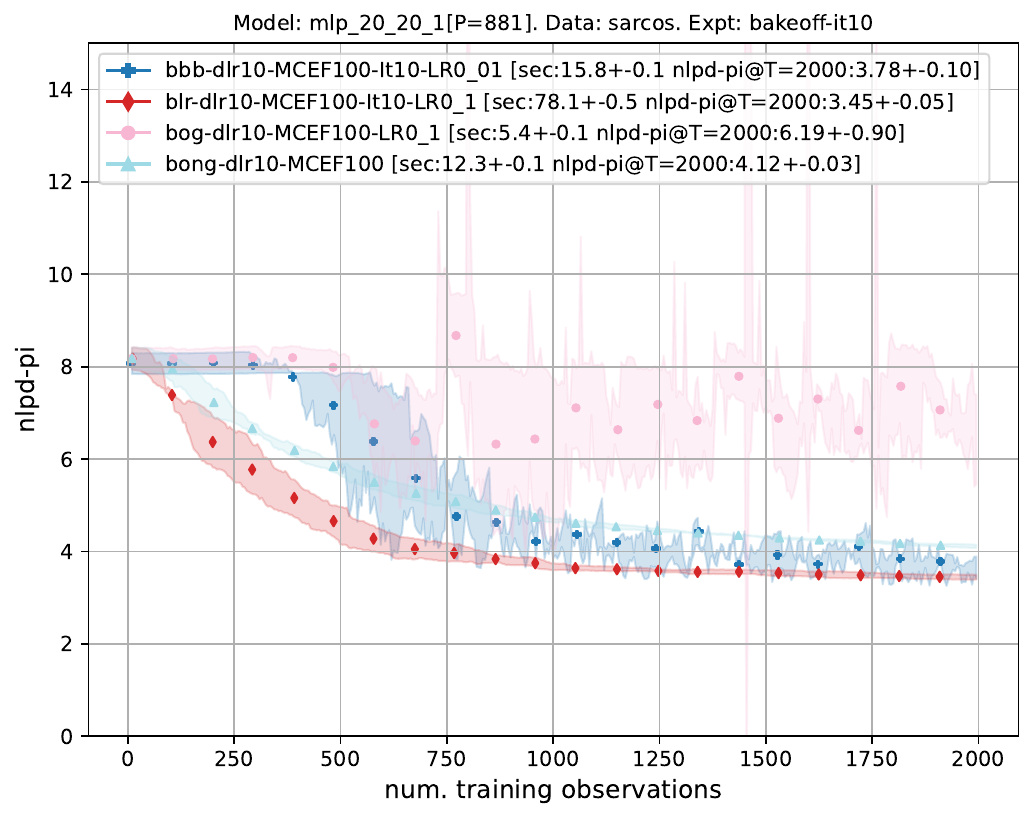}
    \caption{10 iterations.}
    \label{fig:ef-it2}
    \end{subfigure}
    \caption{
    Same as \cref{fig:sarcos-lin}
    except we use \mcef approximation
    with $\mc=100$.
    }
    \label{fig:sarcos-ef}
\end{figure}

In \cref{fig:sarcos-ef} we show the results of using the \mcef approximation with $\mc=100$ samples.
The trends are similar to the \linhess case.
In particular, for $\niter=1$,
\bong and \blr are similar,
with \bong having a slight edge;
and for $\niter=10$,
\bbb catches up with both \bong and \blr,
with \bog always in last place.
Finally, we see that the performance
of \mcef is slightly worse than \linhess
when $\niter=1$,
but catches up with $\niter=10$.
However, in larger scale experiments,
we usually find that \linhess is
significantly better than \mcef,
even with $\niter=10$.

\subsubsection{Learning rate sensitivity}
\label{sec:lr-sensitivity}

\begin{figure}[h!]
    \begin{subfigure}[b]{0.49\textwidth}
    \centering
  \includegraphics[width=\textwidth]{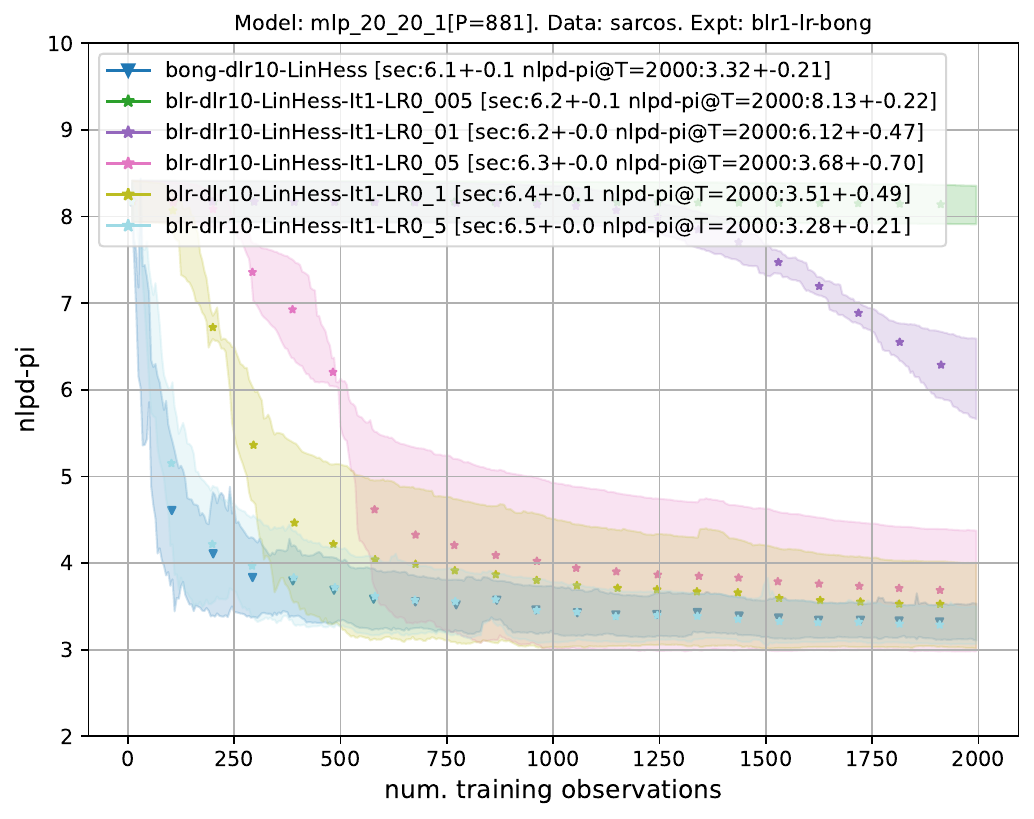}
    \caption{1 iteration.}
    \end{subfigure}
    \hfill
\begin{subfigure}[b]{0.49\textwidth}
    \centering
    \includegraphics[width=\textwidth]
   {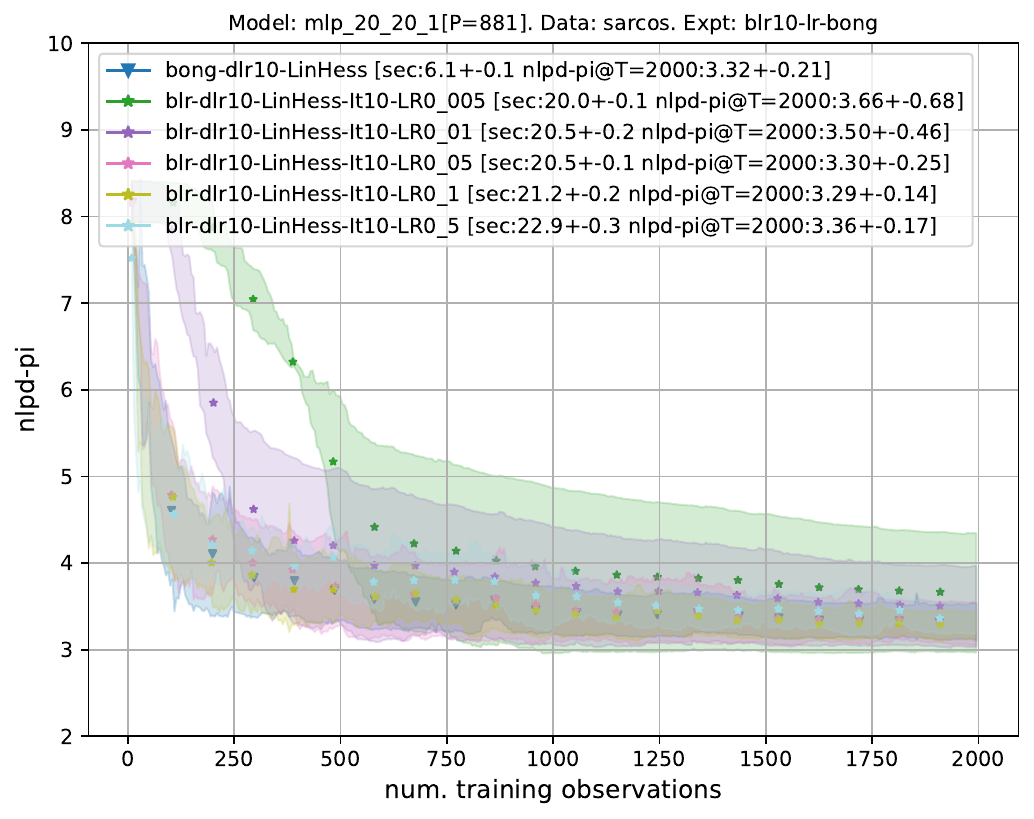}
    \caption{10 iterations.}
    \end{subfigure}
    \caption{
    Same setup as \cref{fig:sarcos-lin},
    except now we plot performance
    for \blr for 5 different learning rates.
    We also show \bong as a baseline,
    which uses a fixed learning rate step size of 1.0.
    }
    \label{fig:sarcos-lr-blr-lin}
\end{figure}

\eat{
\begin{figure}[h!]
    \begin{subfigure}[b]{0.49\textwidth}
    \centering
  \includegraphics[width=\textwidth]{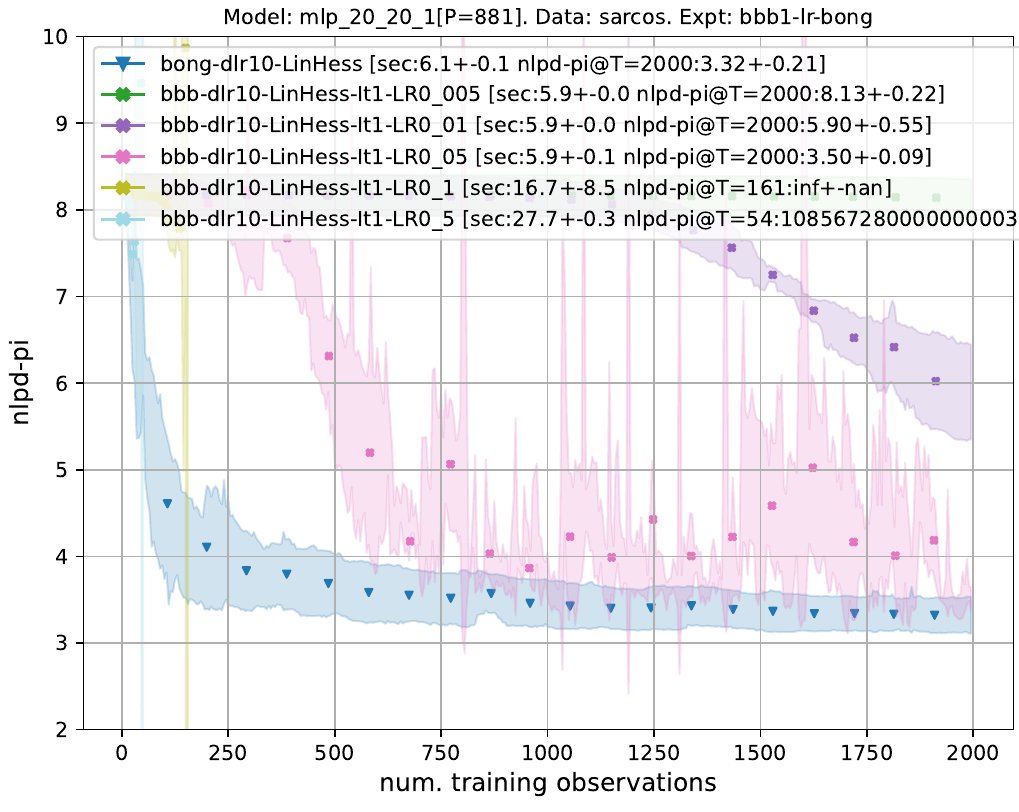}
    \caption{1 iteration.}
    \end{subfigure}
    \hfill
\begin{subfigure}[b]{0.49\textwidth}
    \centering
    \includegraphics[width=\textwidth]
   {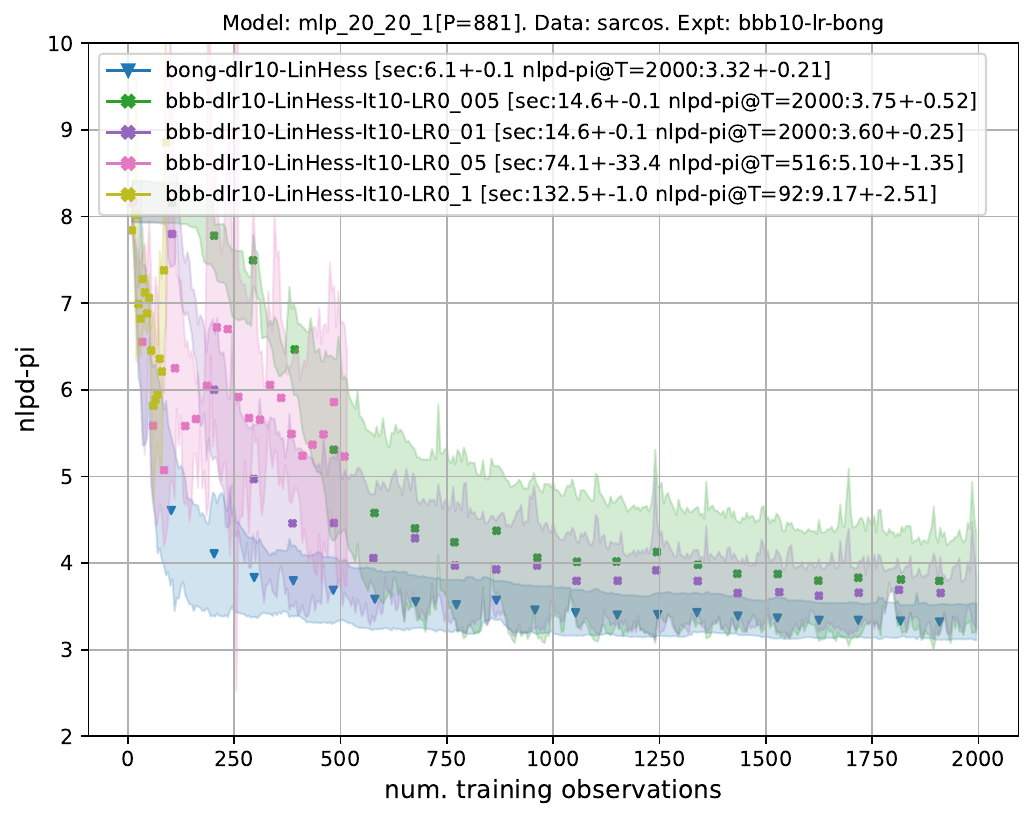}
    \caption{10 iterations.}
    \end{subfigure}
    \caption{
    Same setup as \cref{fig:sarcos-lin},
    except now we plot performance
    for \bbb for 5 different learning rates.
    We also show \bong as a baseline.
    }
    \label{fig:sarcos-lr-bbb-lin}
\end{figure}
}

\begin{figure}[h!]
    \begin{subfigure}[b]{0.49\textwidth}
    \centering
  \includegraphics[width=\textwidth]{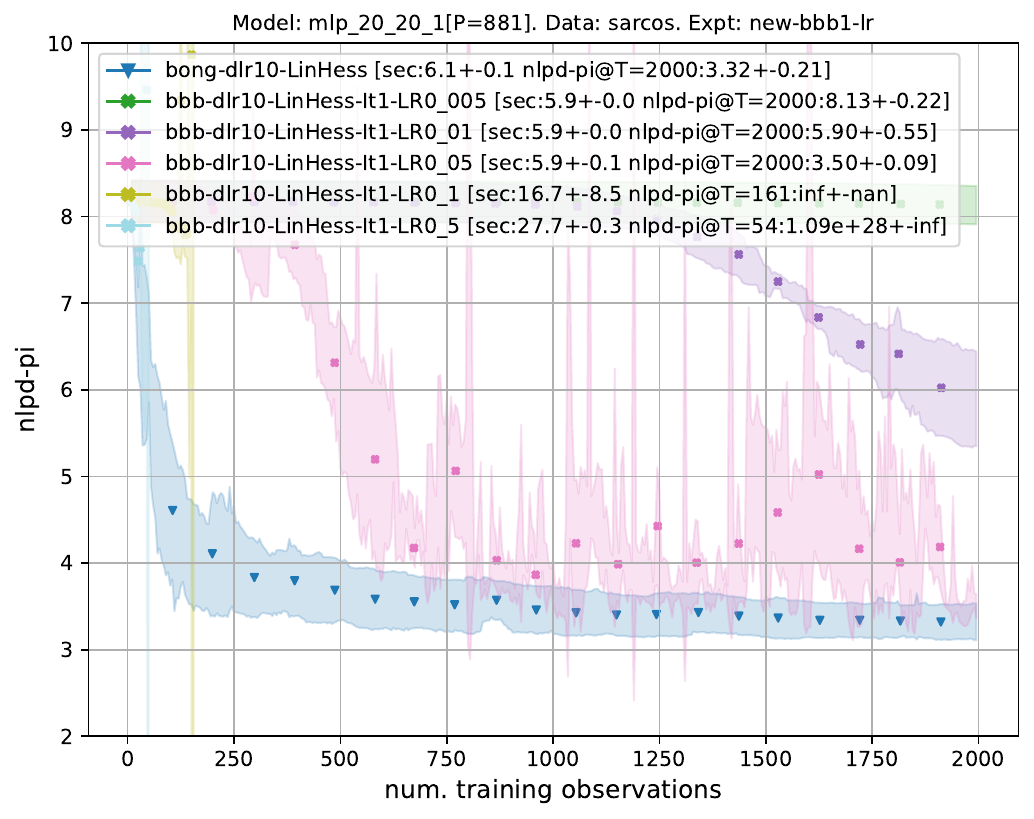}
    \caption{1 iteration.}
    \end{subfigure}
    \hfill
\begin{subfigure}[b]{0.49\textwidth}
    \centering
    \includegraphics[width=\textwidth]
   {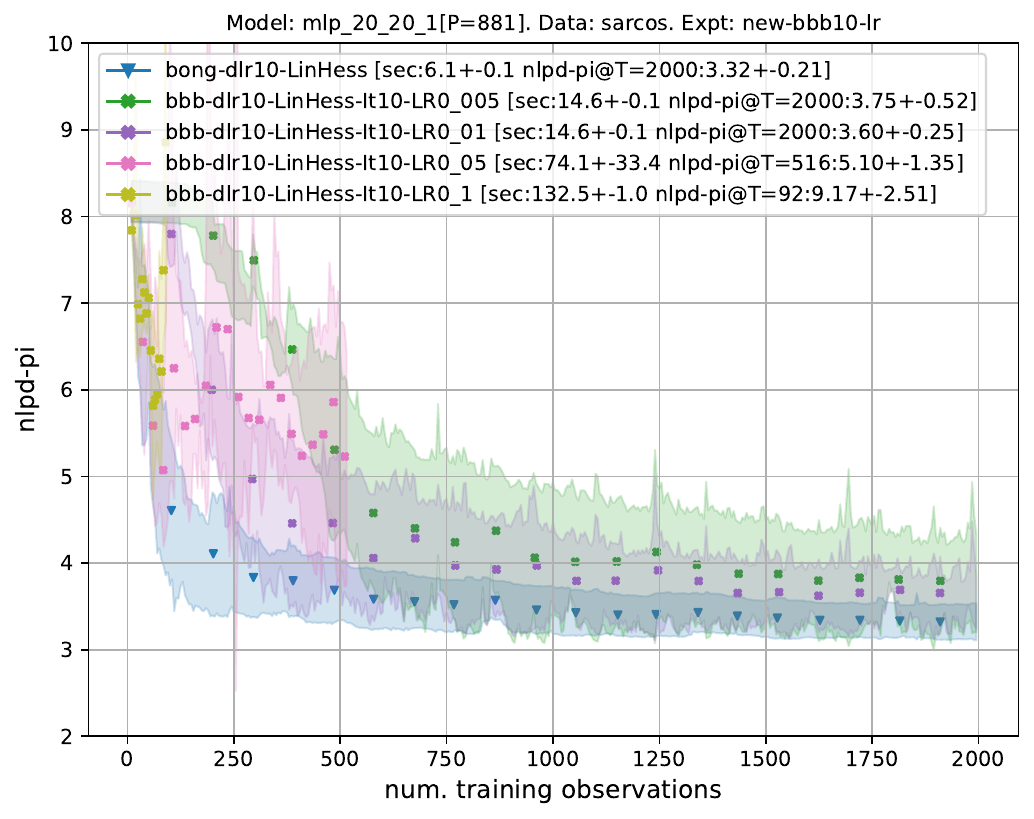}
    \caption{10 iterations.}
    \end{subfigure}
    \caption{
    Same setup as \cref{fig:sarcos-lin},
    except now we plot performance
    for \bbb for 5 different learning rates.
    We also show \bong as a baseline.
    }
    \label{fig:sarcos-lr-bbb-lin}
\end{figure}

\eat{
\begin{figure}[h!]
    \begin{subfigure}[b]{0.49\textwidth}
    \centering
  \includegraphics[width=\textwidth]{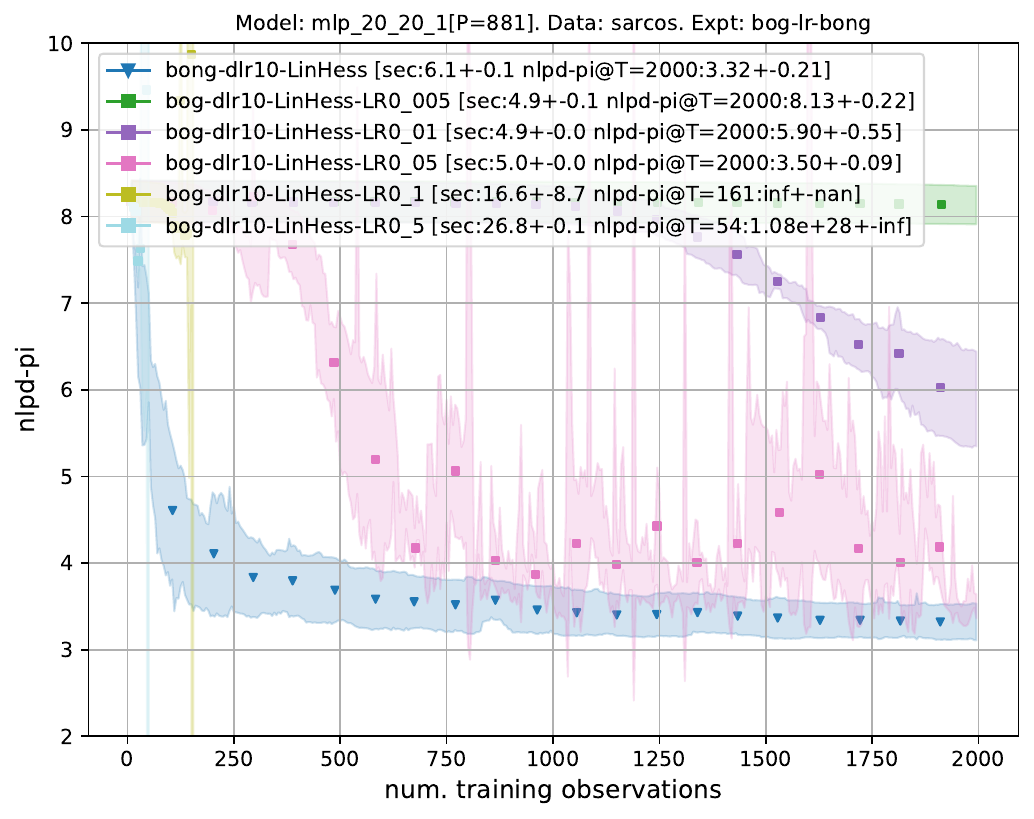}
    \caption{\bog with \linhess.}
     \label{fig:sarcos-lr-bog-linhess}
    \end{subfigure}
    \hfill
\begin{subfigure}[b]{0.49\textwidth}
    \centering
    \includegraphics[width=\textwidth]
   {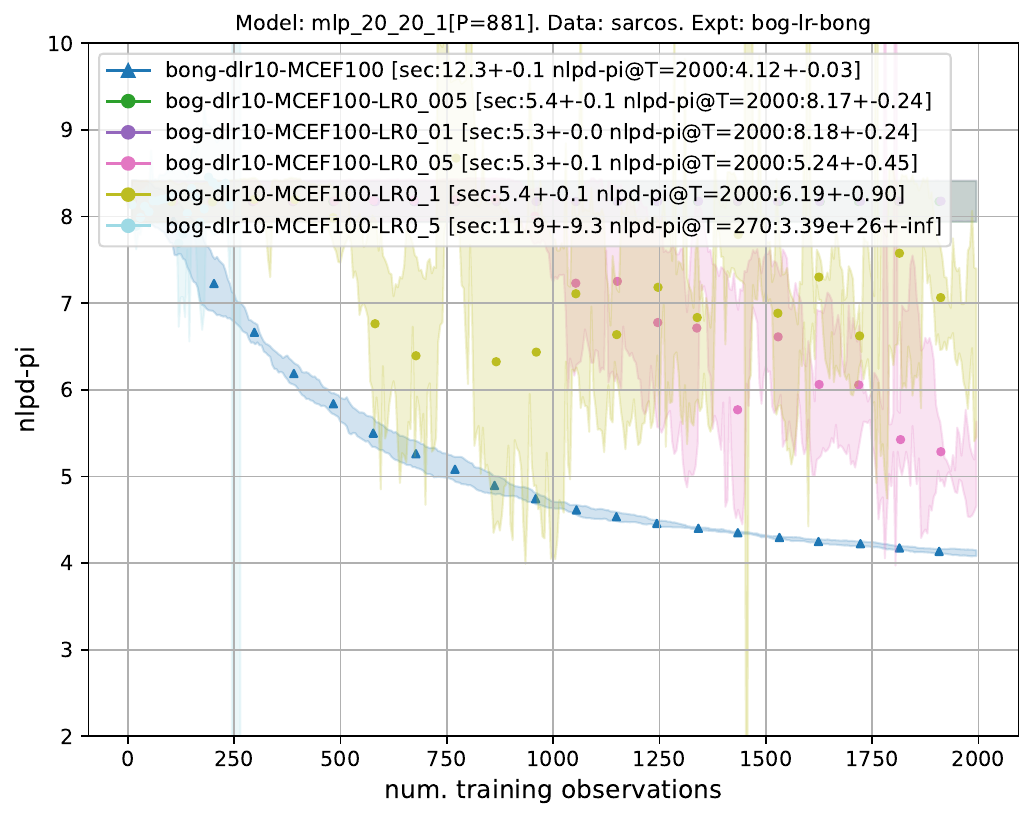}
    \caption{\bog with \mcef.}
     \label{fig:sarcos-lr-bog-mcef}
    \end{subfigure}
    \caption{
    We plot performance
    for \bog for 5 different learning rates.
    We also show \bong as a baseline.
    (a) \linhess approximation.
    (b) \mcef approximation.
    }
    \label{fig:sarcos-lr-bog}
\end{figure}
}

\begin{figure}[h!]
    \begin{subfigure}[b]{0.49\textwidth}
    \centering
  \includegraphics[width=\textwidth]{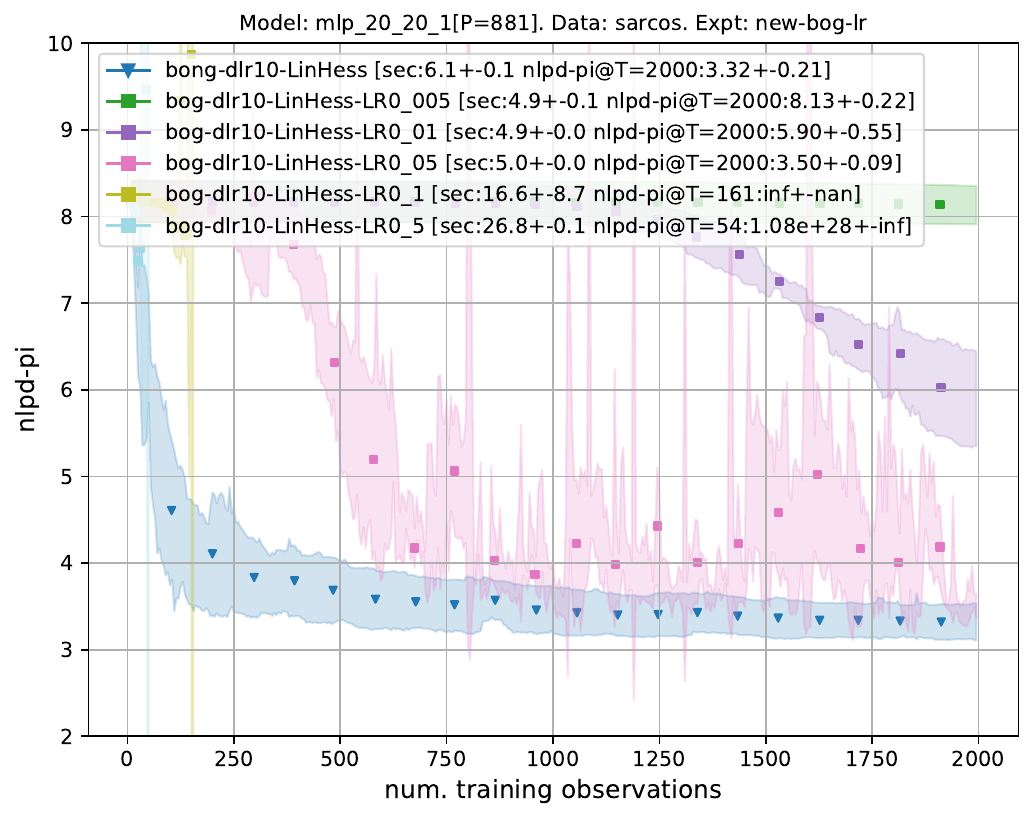}
    \caption{\bog with \linhess.}
     \label{fig:sarcos-lr-bog-linhess}
    \end{subfigure}
    \hfill
\begin{subfigure}[b]{0.49\textwidth}
    \centering
    \includegraphics[width=\textwidth]
   {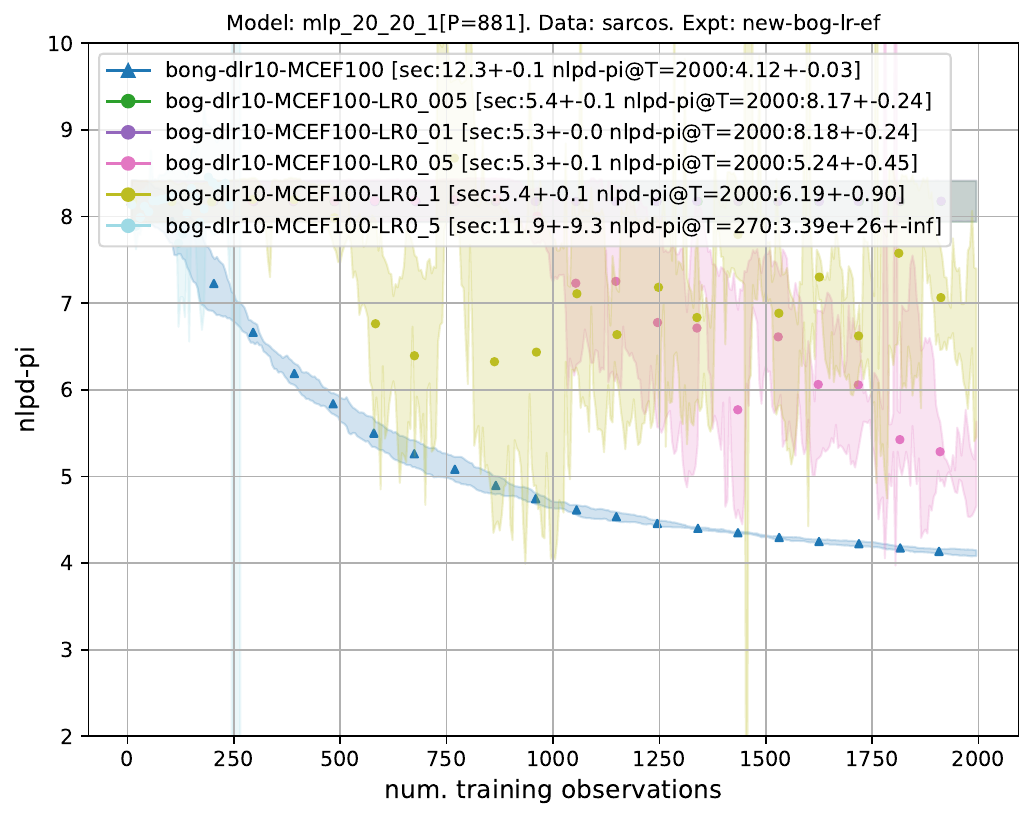}
    \caption{\bog with \mcef.}
     \label{fig:sarcos-lr-bog-mcef}
    \end{subfigure}
    \caption{
    We plot performance
    for \bog for 5 different learning rates.
    We also show \bong as a baseline.
    (a) \linhess approximation.
    (b) \mcef approximation.
    }
    \label{fig:sarcos-lr-bog}
\end{figure}

In \cref{fig:sarcos-lr-blr-lin}
we show the test set performance 
for \blr (with \linhess approximation)
for 5 different learning rates
(namely 
$5\times 10^{-3}$, 
$1\times 10^{-2}$,
$5\times 10^{-2}$,
$1\times 10^{-1}$, 
and $5\times 10^{-1}$).

When using 1 iteration per step,
the best learning rate is
$\alpha=0.5$,
which is also the value chosen based on 
validation set performance.
With this value, \blr matches 
\bong.
For other learning rates,
\blr performance is much worse.
When using 10 iterations per step,
there are several learning rates
all of which give performance as good
as \bong.

In \cref{fig:sarcos-lr-bbb-lin},
we show the analogous plot for \bbb. 
When using 1 iteration per step,
all learning rates result in poor
performance, with many resulting in 
NaNs.
When using 10 iterations per step,
there are some learning rates
that enable \bbb to get close to
(but still not match)
the performance of \bong. 

Finally,  in \cref{fig:sarcos-lr-bog-linhess},
we show the analogous plot for \bog
with \linhess, and 
in \cref{fig:sarcos-lr-bog-mcef}
with \mcef,
where results are much worse.

Overall we conclude that all the methods
(except \bong) are quite sensitive
to the learning rate.
In our experiments, we pick a value
based on performance on a validation set,
but in the truly online setting,
where there is just a single data stream,
picking an optimal learning rate is difficult,
which is an additional advantage of \bong. 

%% file: sections/proof.tex
\section{Proof of \Cref{thm:exact-when-conjugate,thm:lbong}}
\label{sec:proofs}

\begin{proof}[\Cref{thm:exact-when-conjugate}]
    To ease notation we write 
    the natural parameters of the prior as
    $\vpsi_{t\vert t-1} = [\vchi_{t\vert t-1};\nu_{t\vert t-1}]$,
    which can be interpreted as the prior sufficient statistics and prior sample size.
    Note that $\vx_{t}$ can be omitted as a constant. 
    Based on the prior 
    $q_{\vpsi_{t\vert t-1}}(\vtheta_{t})$ 
    the exact posterior is
    \begin{align}
        p(\vtheta_{t}\vert\data_{t}) 
        & \propto 
        q_{\vpsi_{t\vert t-1}} \! (\vtheta_{t}) 
           \, p_{t}(\vy_{t}\vert\vtheta_{t}) \\
       & \propto
        \exp \left(
            \vchi_{t|t-1}^{\trans} \vtheta_{t}
            - \nu_{t|t-1} A(\vtheta_{t})
        \right)
        \exp \left(
            \vtheta_{t}^{\trans} \vy_{t}
            - A(\vtheta_{t})
        \right)\\
        & \propto q_{\vpsi_{t}}(\vtheta_t)\\
        \vpsi_t
        &= \left[ \begin{array}{c}
            \vchi_{t|t-1} + \vy_{t}  \\
            \nu_{t|t-1}+1
        \end{array} \right]
        \label{eq:bayes-conjugate-update} 
    \end{align}
    For \bong, we first note the dual parameter is given by
    \begin{align}
        \vrho_{t|t-1}
        &= \expectQ
            {T(\vtheta_t)}
            {\vtheta_{t}\sim q_{\vpsi_{t\vert t-1}}}\\
        &= \expectQ
            {\begin{array}{c}
                \vtheta_{t}\\
                -A(\vtheta_{t})
            \end{array}}
            {\vtheta_{t}\sim q_{\vpsi_{t\vert t-1}}}
    \end{align}
    Therefore the natural gradient in \cref{eq:bong-MD} is
    \begin{align}
        \nabla_{\vrho_{t|t-1}} 
        \expectQ
            {\log p\left(\vy_{t}\vert\vtheta_{t}\right)}
            {\vtheta_{t}\sim q_{\vpsi_{t\vert t-1}}} 
        &= \nabla_{\vrho_{t|t-1}} 
        \expectQ
            {\vtheta_{t}^{\top}\vy_{t}-A\left(\vtheta_{t}\right)}
            {\vtheta_{t}\sim q_{\vpsi_{t\vert t-1}}} \\
        &= \nabla_{\vrho_{t|t-1}} \vrho_{t|t-1}^{\trans} \left[\begin{array}{c}
            \vy_{t} \\
            1
        \end{array}\right] \\
        &= \left[\begin{array}{c}
            \vy_{t}\\
            1
        \end{array}\right]
    \end{align}
    Therefore the \bong update yields 
    $\vpsi_{t} = [\vchi_{t|t-1} + \vy_{t}; \nu_{t|t-1} + 1]$
    in agreement with \cref{eq:bayes-conjugate-update}.
\end{proof}

\begin{proof}[\Cref{thm:lbong}]
The intuition behind this proof is as follows.
For the mean update in \cref{eq:RVGA-explicit-mean} $\nabla_{\vtheta_t}\ell_t$ is linear in $\vtheta_t$ so the expectation equals the value at the mean.
For the covariance update in \cref{eq:RVGA-explicit-cov} $\nabla^2_{\vtheta_t}\ell_t$ is independent of $\vtheta_t$ so we can drop the expectation operator. 
The tricky part is why we need different linearizations for the Hessians to agree. It has to do with making the Hessian of the NN disappear (as in GGN). In the Gaussian approximation this happens when the predicted mean ($\hat{\vy}_t = f_t(\vtheta_t)$) is linear in $\vtheta_t$. In the plugin approximation it happens when the outcome-dependent part of the loglikelihood ($h_t(\vtheta_t)^\trans \vy_t$) is linear in $\vtheta_t$. In the latter case the only nonlinear term remaining in the log-likelihood is the log-partition $A$, and the two methods end up agreeing because of the property that the Hessian of the log-partition equals the conditional variance $\vR_t$.

Formally, under the linear($h$)-Gaussian approximation in \cref{eq:hbar,eq:gaussian-likelihood}
the expected gradient and Hessian can be calculated directly:
\begin{align}
    \expectQ
        {\nabla_{\vtheta_{t}} \log \bar{p}^{\rev{\rm LG}}_t(\vy_{t}|\vtheta_{t}))}
        {\vtheta_{t} \sim q_{\vpsi_{t|t-1}}}
    &= \expectQ
        {\vH_t^\trans \vR_t^{-1} (\vy_t - \hat{\vy}_t - \vH_t (\vtheta_t - \vmu_{t|t-1})}
        {\vtheta_{t} \sim q_{\vpsi_{t|t-1}}} \\
    &= \vH_t^\trans \vR_t^{-1} (\vy_t - \hat{\vy}_t) \\
    \expectQ
        {\nabla_{\vtheta_{t}}^{2} \log \bar{p}^{\rev{\rm LG}}_t(\vy_{t} |  \vtheta_{t})}
        {\vtheta_{t} \sim q_{\vpsi_{t|t-1}}} 
    &= \expectQ
        {-\vH_t^\trans \vR_t^{-1} \vH_t}
        {\vtheta_{t} \sim q_{\vpsi_{t|t-1}}} \\
    &= - \vH_t^\trans \vR_t^{-1} \vH_t
\end{align}

For the linear($f$)-delta approximation we use the properties of exponential families that (1) the gradient of the log-partition $A$ with respect to the natural parameter $f_t(\vtheta_t)$ equals the expectation parameter $h_t(\vtheta_t)$, (2) the Hessian of the log-partition with respect to the natural parameter equals the conditional variance, and consequently (3) the Jacobian of the expectation parameter with respect to the natural parameter equals the conditional variance $\vR_t$:
\begin{align}
    \nabla_{\veta = f_t(\vmu_{t|t-1})} A(\veta)
    &= h_t(\vmu_{t|t-1})
    = \hat{\vy} \\
    \nabla^2_{\veta = f_t(\vmu_{t|t-1})} A(\veta)
    &= \var{\vy_t | \vtheta_t = \vmu_{t|t-1}} 
    = \vR_t \\
    \frac{\partial h_t(\vtheta_t)}{\partial f_t(\vtheta_t)}_{\vert\vtheta_t = \vmu_{t|t-1}}
    &= \vR_t
\end{align}
The last of these implies $\vF_t = \vR_t^{-1} \vH_t$.
Therefore the expected gradient and Hessian can be calculated as
\begin{align}
    \expectQ
        {\nabla_{\vtheta_{t}} 
        \log \bar{p}^{\rev{\rm LD}}_t(\vy_{t} | \vtheta_{t})}
        {\vtheta_{t} \sim \delta_{\vmu_{t|t-1}}}
    &= \nabla_{\vtheta_{t} = \vmu_{t|t-1}} 
    \log \bar{p}^{\rev{\rm LD}}_t(\vy_{t} | \vtheta_{t}) \\
    &= \vF_t^\trans \vy_t 
    - \vF_t^\trans \nabla_{\veta = f_t(\vmu_{t|t-1})} A(\veta) \\
    &= \vF_t^\trans (\vy_t - \hat{\vy}_t) \\
    &= \vH_t^\trans \vR_t^{-1} (\vy_t - \hat{\vy}_t) \\
    \expectQ
        {\nabla^2_{\vtheta_{t}} 
        \log \bar{p}^{\rev{\rm LD}}_t(\vy_{t} | \vtheta_{t})}
        {\vtheta_{t} \sim \delta_{\vmu_{t|t-1}}}
    &= \nabla^2_{\vtheta_{t} = \vmu_{t|t-1}} 
    \log \bar{p}^{\rev{\rm LD}}_t(\vy_{t} | \vtheta_{t}) \\
    &= - \vF_t^\trans
    \left(\nabla^2_{\veta = f_t(\vmu_{t|t-1})} A(\veta)\right) 
    \vF_t\\
    &= - \vF_t^\trans \vR_t \vF_t \\
    &= - \vH_t^\trans \vR_t^{-1} \vH_t
\end{align}
\end{proof}


%% file: sections/MD.tex
\section{Mirror descent formulation}
\label{sec:MD-formulation}

In this section we give a more detailed derivation of \bong as mirror descent
and use this to give two alternative interpretations of how \bong approximates exact VB:
(1) by approximating the expected NLL as linear in the expectation parameter $\vrho$, or 
(2) by replacing an implicit update with an explicit one.

Assume the variational family is an exponential one as introduced at the end of \cref{sec:background}, with natural and dual parameters $\vpsi$ and $\vrho$, sufficient statistics $T(\vtheta)$, and log-partition $\Phi(\vpsi)$:
\begin{align}
    q_{\vpsi}(\vtheta)
    &= \exp\left(\vpsi^{\trans} T(\vtheta) -\Phi(\vpsi\right)) \\
    \vrho &= 
    \expectQ {T(\vtheta)} {\vtheta\sim q_{\vpsi}}
\end{align}
We first review how NGD on an exponential family is a special case of mirror descent \citep{BLR,martens2020new}.
The mirror map is the gradient of the log-partition, which satisfies the thermodynamic identity
\begin{align}
    \vrho = \nabla\Phi(\vpsi)
\end{align} 
This is a bijection when $\Phi$ is convex (which includes the cases we study), so we can implicitly treat $\vpsi$ and $\vrho$ as functions of each other.
Given a loss function $L(\vpsi)$, MD iteratively solves the local optimization problem
\begin{align}
    \vpsi_{i+1} 
    &= \underset{\vpsi}{\arg\min}
    \left\langle \nabla_{\vrho_{i}} L(\vpsi_{i}),\vrho\right\rangle 
    + \frac{1}{\alpha} \mathbb{D}_{\Phi}(\vpsi_{i},\vpsi)
    \label{eq:MD-opt-problem}
\end{align}
The first term is a linear (in $\vrho$) approximation of $L$ about the previous iteration $\vpsi_{i}$ 
and the second term is the Bregman divergence
\begin{equation}
    \mathbb{D}_{\Phi}(\vpsi_{i},\vpsi_{i+1})
    = \Phi(\vpsi_{i})
    - \Phi(\vpsi_{i+1})
    - \left(\vpsi_{i}-\vpsi_{i+1}\right)^{\trans} \vrho_{i+1}
\end{equation}
The Bregman divergence acts as a regularizer toward $\vpsi_{i}$ and captures the intrinsic geometry of the parameter space because of its equivalence with the (reverse) KL divergence
\begin{align}
    \KLpq{q_{\vpsi_{i+1}}}{q_{\vpsi_{i}}}
    &= \expectQ
        {\left(\vpsi_{i+1}-\vpsi_{i}\right)^{\trans} T(\vtheta) + \Phi(\vpsi_{i}) - \Phi(\vpsi_{i+1})}
        {\vtheta\sim q_{\vpsi_{i+1}}}\\
    &= \mathbb{D}_{\Phi}(\vpsi_{i},\vpsi_{i+1})
    \label{eq:KL-bregman}
\end{align}
Importantly, this recursive regularizer is not part of the loss and serves only to define an iterated algorithm that converges to a local minimum of $L$.
Solving \cref{eq:MD-opt-problem} by differentiating by $\vrho$ yields the MD update
\begin{equation}
    \vpsi_{i+1}
    = \vpsi_{i} - \alpha \nabla_{\vrho_{i}} L(\vpsi_{i})
    \label{eq:MD-iterated}
\end{equation}
Because the Fisher matrix for an exponential family is $\vF_{\vpsi} = \partial\vrho / \partial\vpsi$, this is equivalent to NGD with respect to $\vpsi$.
\cite{BLR} offer this as a derivation of the BLR, when $L(\vpsi)$ is taken to be the variational loss from \cref{eq:nELBO}.

By applying this analysis to the online setting,
our approach can be seen to follow from two insights. 
First, the online variational loss in \cref{eq:online-variational-loss} already includes KL divergence from the previous step, 
so we do not need the artificial regularizer in \cref{eq:MD-opt-problem}. 
That is, if we start from the online variational problem in \cref{eq:online-VB} 
and define $L_t(\vpsi)$ as the expected NLL,
\begin{equation}
    L_t(\vpsi)
    = -\expectQ
        {\log p(\vy_{t}\vert f_t(\vtheta_{t}))}
        {\vtheta_{t}\sim q_{\vpsi}}
\end{equation}
then replacing $L_t(\vpsi)$ with a linear approximation based at $\vpsi_{t|t-1}$ and applying \cref{eq:KL-bregman} leads to  
\begin{align}
    \vpsi_{t} 
    &= \underset{\vpsi}{\arg\min}
    \left\langle \nabla_{\vrho_{t|t-1}}L_{t}(\vpsi_{t|t-1}), \vrho \right\rangle 
    + \mathbb{D}_{\Phi}(\vpsi_{t\vert t-1},\vpsi)
    \label{eq:bong-linearized-optimization}
\end{align}
By comparing to \cref{eq:MD-opt-problem} we see this defines an MD algorithm with unit learning rate that works in a single step rather than by iterating.
Paralleling the derivation of \cref{eq:MD-iterated} from \cref{eq:MD-opt-problem} we get
\begin{align}
    \vpsi_{t} 
    &= \vpsi_{t|t-1} - \nabla_{\vrho_{t|t-1}} L_{t}(\vpsi_{t|t-1})
    \label{eq:bong-as-linearized-optimization}
\end{align}
which matches the \bong update in \cref{eq:bong-MD}.
Thus \bong can be seen as an approximate solution of the online variational problem in \cref{eq:online-VB}
based on linearizing the expected NLL wrt $\vrho$.
(Note this is different from the assumption underlying \lbong that $f_t(\vtheta_t)$ or $h_t(\vtheta_t)$ is linear in $\vtheta_t$.)

Second, in the conjugate case, this linearity assumption is true: $L_{t}$ is linear in $\vrho$ (see proof of \cref{thm:exact-when-conjugate}).
Therefore \ref{eq:bong-linearized-optimization} is equivalent to solving \cref{eq:online-VB} exactly:
\begin{align}
    \vpsi_{t} 
    &= \underset{\vpsi}{\arg\min} \, L_{t}(\vpsi) + \mathbb{D}_{\Phi}(\vpsi_{t|t-1}, \vpsi)
    \label{eq:online-VB-as-MD}
\end{align}
This recapitulates \cref{thm:exact-when-conjugate} that \bong is Bayes optimal in the conjugate case. 
In general the exact solution to \cref{eq:online-VB-as-MD} is 
\begin{align}
    \vpsi_{t} 
    &= \vpsi_{t|t-1} - \nabla_{\vrho_{t}} L_{t}(\vpsi_{t})
    \label{eq:bong-implicit}
\end{align}
This is an implicit update because the gradient is evaluated at the (unknown) posterior, 
whereas \cref{eq:bong-as-linearized-optimization} is an explicit update because it evaluates the gradient at the prior.
(In the Gaussian case these can be shown to match the implicit and explicit RVGA updates of \citep{RVGA}.)
Therefore \bong can be also interpreted as an approximation of exact VB that replaces the implicit update, \cref{eq:bong-implicit}, with an explicit update, \cref{eq:bong-as-linearized-optimization}.

\eat{
\textcolor{red}{To do:}
\begin{itemize}
\item Alternatively we can work with the implicit update and solve it iteratively. 
\item Is \cref{eq:bong-implicit} the same as BLR or do we get something
new, or at least prettier? 
\item How does our iterated EKF fit here?
\item Derive implicit and explicit RVGA from \cref{eq:bong-implicit} and \cref{eq:bong-as-linearized-optimization} when $q_{\vpsi}$ is Gaussian
\end{itemize}
}

%% file: sections/derivations.tex
\section{Derivations}
\label{sec:derivations}

\input{sections/derivations-general}
\input{sections/derivations-FC}
\input{sections/derivations-FC-mom}
\input{sections/derivations-diag}
\input{sections/derivations-diag-mom}

\input{sections/derivations-DLR}
\input{sections/BLR-batch}

%% file: sections/derivations-general.tex
This section derives the update equations for all 80 algorithms we investigate (\cref{tab:methods} plus the \mchess and \linef variants). In \cref{sec:blr-batch} we also translate the \blr algorithms from our online setting back to the batch setting used in \cite{BLR}.

For an exponential variational family with natural parameters $\vpsi$
and dual parameters $\vrho$, we can derive all 16 methods (\bong, \blr, \bog, \bbb under all four Hessian approximations) from four quantities:
\begin{align}
&\nabla_{\vrho_{t,i-1}}\expectQ{\log p\left(\vy_{t}\vert f_{t}\left(\vtheta_{t}\right)\right)}{\vtheta_{t}\sim q_{\vpsi_{t,i-1}}}\\
&\nabla_{\vrho_{t,i-1}}\KLpq{q_{\vpsi_{t,i-1}}}{q_{\vpsi_{t|t-1}}}\\
&\nabla_{\vpsi_{t,i-1}}\expectQ{\log p\left(\vy_{t}\vert f_{t}\left(\vtheta_{t}\right)\right)}{\vtheta_{t}\sim q_{\vpsi_{t,i-1}}}\\
&\nabla_{\vpsi_{t,i-1}}\KLpq{q_{\vpsi_{t,i-1}}}{q_{\vpsi_{t|t-1}}}
\end{align}
The NGD methods (\bong and \blr) use gradients with respect to $\vrho_{t,t-i}$
while the GD methods (\bog and \bbb) use gradients with respect to $\vpsi_{t,i-1}$.
For \bong and \bog the $D_{\mathbb{KL}}$ term is not relevant, and there is
no inner loop so $\vpsi_{t,i-1}=\vpsi_{t|t-1}$ and $\vg_{t,i}=\vg_{t}$,
$\vG_{t,i}=\vG_{t}$.

When $\vpsi$ is not the natural parameter of an exponential family
we must explicitly compute the inverse-Fisher preconditioner for the
NGD methods. Therefore the updates can be derived from these three
quantities:
\begin{align}
&\vF_{\vpsi_{t,i-1}}\\
&\nabla_{\vpsi_{t,i-1}}\expectQ{\log p\left(\vy_{t}\vert f_{t}\left(\vtheta_{t}\right)\right)}{\vtheta_{t}\sim q_{\vpsi_{t,i-1}}}\\
&\nabla_{\vpsi_{t,i-1}}\KLpq{q_{\vpsi_{t,i-1}}}{q_{\vpsi_{t|t-1}}}
\end{align}

We will frequently use Bonnet's and Price's theorems \citep{bonnet1964,price1958useful}
\begin{align}
\nabla_{\vmu_{t,i-1}}\expectQ{\log p\left(\vy_{t}|f_{t}\left(\vtheta_{t}\right)\right)}{\gauss\left(\vmu_{t,i-1},\vSigma_{t,i-1}\right)} & =\expectQ{\nabla_{\vtheta_{t}}\log p\left(\vy_{t}|f_{t}\left(\vtheta_{t}\right)\right)}{\gauss\left(\vmu_{t,i-1},\vSigma_{t,i-1}\right)}\\
&=\vg_{t,i}\\
\nabla_{\vSigma_{t,i-1}}\expectQ{\log p\left(\vy_{t}|f_{t}\left(\vtheta_{t}\right)\right)}{\gauss\left(\vmu_{t,i-1},\vSigma_{t,i-1}\right)} & =\tfrac{1}{2}\expectQ{\nabla_{\vtheta_{t}}^{2}\log p\left(\vy_{t}|f_{t}\left(\vtheta_{t}\right)\right)}{\gauss\left(\vmu_{t,i-1},\vSigma_{t,i-1}\right)} \\
&=\tfrac{1}{2}\vG_{t,i}
\end{align}
For diagonal Gaussians with covariance $\Diag\left(\vsigma^{2}\right)$,
Price's theorem also implies%
\footnote{We use $\diag(\vA)$ to denote the vector of diagonal elements of matrix $\vA$ and $\Diag(\vv)$ to denote the matrix whose diagonal entries are $\vv$ and off-diagonal entries are $0$.}
\begin{equation}
\nabla_{\vsigma_{t,i-1}^{2}}\expectQ{\log p\left(\vy_{t}|f_{t}\left(\vtheta_{t}\right)\right)}{\vtheta_{t}\sim\gauss\left(\vmu_{t,i-1},\vSigma_{t,i-1}\right)}=\tfrac{1}{2}\diag\left(\vG_{t,i}\right)
\end{equation}

Update equations for \hessmc, \efmc and \eflin methods are displayed together in the subsections that follow,
because for the most part they differ only in the choice of 
$\GMCH_t$, $\GMCEF_t$ or $\GLEF_t$ to approximate $\vG_t$
and $\gMC_t$ or $\gL_t$ to approximate $\vg_t$.
We note cases where decomposing 
$\GMCEF_t = -\frac{1}{M}\gradmat_t \gradmatT_t$
or $\GLEF_t = -\gL_t \left(\gL_t\right)^\trans$
allows a more efficient update.

We derive updates for \bong-\hesslin and \bog-\hesslin from the corresponding \bong-\hessmc
and \bog-\hessmc updates using \cref{thm:lbong} which entails substituting
\begin{align}
\vg_{t} & \to\vH_{t}^{\trans}\vR_{t}^{-1}(\vy_{t}-\hat{\vy}_{t})\\
\vG_{t} & \to-\vH_{t}^{\trans}\vR_{t}^{-1}\vH_{t}
\end{align}
For the algorithms with inner loops (\blr and \bbb) we adapt the
notation of \cref{sec:linearized} as follows:
\begin{align}
\vy_{t,i} & =h_{t}\left(\vmu_{t,i-1}\right)\\
\vH_{t,i} &=
\rev{\frac{\partial h_t}{\partial\vtheta_t}_{\vert\vtheta=\vmu_{t,i-1}}}\\
\vR_{t,i} & =\var{\vy_{t}|\vtheta_{t}=\vmu_{t,i-1}}\\
\vg_{t,i} & =\vH_{t,i}^{\trans}\vR_{t,i}^{-1}\left(\vy_{t}-\hat{\vy}_{t,i}\right)\\
\vG_{t,i} & =-\vH_{t,i}^{\trans}\vR_{t,i}^{-1}\vH_{t,i}
\end{align}
This corresponds to basing the linear($f$)-Gaussian and linear($h$)-delta
approximations at $\vmu_{t,i-1}$ instead of $\vmu_{t|t-1}$. Thus
the updates for \blr-\hesslin and \bbb-\hesslin are obtained by substituting 
\begin{align}
\vg_{t,i} & \to\vH_{t,i}^{\trans}\vR_{t,i}^{-1}(\vy_{t}-\hat{\vy}_{t,i})\\
\vG_{t,i} & \to-\vH_{t,i}^{\trans}\vR_{t,i}^{-1}\vH_{t,i}
\end{align}

%% file: sections/derivations-FC.tex
\subsection{Full covariance Gaussian, natural parameters\label{sec:FC-Nat}}

The natural and dual parameters for a general Gaussian are given by
\begin{alignat}{3}
    \vpsi_{t,i-1}^{(1)}
    &=\vSigma_{t,i-1}^{-1}\vmu_{t,i-1} 
    &\qquad \vrho_{t,i-1}^{(1)}
    &=\vmu_{t,i-1}\\
    \vpsi_{t,i-1}^{(2)}
    &=-\tfrac{1}{2}\vSigma_{t,i-1}^{-1} 
    & \vrho_{t,i-1}^{(2)}
    &=\vmu_{t,i-1}\vmu_{t,i-1}^{\trans}+\vSigma_{t,i-1}
\end{alignat}
Inverting these relationships gives
\begin{alignat}{3}
    \vmu_{t,i-1} 
    &=-\tfrac{1}{2}\vpsi_{t,i-1}^{\left(2\right)-1}\vpsi_{t,i-1}^{\left(1\right)}
    &&=\vrho_{t,i-1}^{\left(1\right)}\\
    \vSigma_{t,i-1} 
    & =-\tfrac{1}{2}\vpsi_{t,i-1}^{\left(2\right)-1}
    &&=\vrho_{t,i-1}^{\left(2\right)}-\vrho_{t,i-1}^{\left(1\right)}\vrho_{t,i-1}^{\left(1\right)\trans}
\end{alignat}

The KL divergence in the VI loss is
\begin{align}
    \KLpq{q_{\vpsi_{t,i-1}}}{q_{\vpsi_{t|t-1}}}
    &=\tfrac{1}{2}\left(\vmu_{t,i-1}-\vmu_{t|t-1}\right)^{\trans}\vSigma_{t|t-1}^{-1}\left(\vmu_{t,i-1}-\vmu_{t|t-1}\right) \nonumber\\
    &\quad+\tfrac{1}{2}Tr\left(\vSigma_{t|t-1}^{-1}\vSigma_{t,i-1}\right)
    -\tfrac{1}{2}\log\left|\vSigma_{t,i-1}\right|)+\const
\end{align}
with gradients
\begin{align}
\nabla_{\vmu_{t,i-1}}\KLpq{q_{\vpsi_{t,i-1}}}{q_{\vpsi_{t|t-1}}} & =\vSigma_{t|t-1}^{-1}\left(\vmu_{t,i-1}-\vmu_{t|t-1}\right)\\
\nabla_{\vSigma_{t,i-1}}\KLpq{q_{\vpsi_{t,i-1}}}{q_{\vpsi_{t|t-1}}} & =\tfrac{1}{2}\left(\vSigma_{t|t-1}^{-1}-\vSigma_{t,i-1}^{-1}\right)
\end{align}

Following Appendix C of \citep{khan2018fast}, for any scalar function $\vell$
the chain rule gives
\begin{align}
\nabla_{\vrho_{t,i-1}^{\left(1\right)}}\ell & =\frac{\partial\vmu_{t,i-1}}{\partial\vrho_{t,i-1}^{\left(1\right)}}\nabla_{\vmu_{t,i-1}}\ell+\frac{\partial\vSigma_{t,i-1}}{\partial\vrho_{t,i-1}^{\left(1\right)}}\nabla_{\vSigma_{t,i-1}}\ell\\
 & =\nabla_{\vmu_{t,i-1}}\ell-2\left(\nabla_{\vSigma_{t,i-1}}\ell\right)\vmu_{t,i-1}\\
\nabla_{\vrho_{t,i-1}^{\left(2\right)}}\ell & =\frac{\partial\vmu_{t,i-1}}{\partial\vrho_{t,i-1}^{\left(2\right)}}\nabla_{\vmu_{t,i-1}}\ell+\frac{\partial\vSigma_{t,i-1}}{\partial\vrho_{t,i-1}^{\left(2\right)}}\nabla_{\vSigma_{t,i-1}}\ell\\
 & =\nabla_{\vSigma_{t,i-1}}\ell
\end{align}
Therefore
\begin{align}
\nabla_{\vrho_{t,i-1}^{\left(1\right)}}\expectQ{\log p\left(\vy_{t}\vert f_{t}\left(\vtheta_{t}\right)\right)}{\vtheta_{t}\sim q_{\vpsi_{t,i-1}}} & =\vg_{t,i}-\vG_{t,i}\vmu_{t,i-1}\label{eq:FC-NLL-rho1}\\
\nabla_{\vrho_{t,i-1}^{\left(2\right)}}\expectQ{\log p\left(\vy_{t}\vert f_{t}\left(\vtheta_{t}\right)\right)}{\vtheta_{t}\sim q_{\vpsi_{t,i-1}}} & =\tfrac{1}{2}\vG_{t,i}\label{eq:FC-NLL-rho2}
\end{align}
and
\begin{align}
\nabla_{\vrho_{t,i-1}^{\left(1\right)}}\KLpq{q_{\vpsi_{t,i-1}}}{q_{\vpsi_{t|t-1}}} & =\vSigma_{t,i-1}^{-1}\vmu_{t,i-1}-\vSigma_{t|t-1}^{-1}\vmu_{t|t-1}\label{eq:FC-KL-rho1}\\
\nabla_{\vrho_{t,i-1}^{\left(2\right)}}\KLpq{q_{\vpsi_{t,i-1}}}{q_{\vpsi_{t|t-1}}} & =\tfrac{1}{2}\left(\vSigma_{t|t-1}^{-1}-\vSigma_{t,i-1}^{-1}\right)\label{eq:FC-KL-rho2}
\end{align}

Following the same approach for $\vpsi$ gives 
\begin{align}
\nabla_{\vpsi_{t,i-1}^{\left(1\right)}}\ell & =\frac{\partial\vmu_{t,i-1}}{\partial\vpsi_{t,i-1}^{\left(1\right)}}\nabla_{\vmu_{t,i-1}}\ell+\frac{\partial\vSigma_{t,i-1}}{\partial\vpsi_{t,i-1}^{\left(1\right)}}\nabla_{\vSigma_{t,i-1}}\ell\\
 & =-\tfrac{1}{2}\vpsi_{t,i-1}^{\left(2\right)-1}\nabla_{\vmu_{t,i-1}}\ell\\
 & =\vSigma_{t,i-1}\nabla_{\vmu_{t,i-1}}\ell\\
\nabla_{\vpsi_{t,i-1}^{\left(2\right)}}\ell & =\frac{\partial\vmu_{t,i-1}}{\partial\vpsi_{t,i-1}^{\left(2\right)}}\nabla_{\vmu_{t,i-1}}\ell+\frac{\partial\vSigma_{t,i-1}}{\partial\vpsi_{t,i-1}^{\left(2\right)}}\nabla_{\vSigma_{t,i-1}}\ell\\
 & =\tfrac{1}{2}\vpsi_{t,i-1}^{\left(2\right)-1}\left(\nabla_{\vmu_{t,i-1}}\ell\right)\vpsi_{t,i-1}^{\left(1\right)\trans}\vpsi_{t,i-1}^{\left(2\right)-1}+\tfrac{1}{2}\vpsi_{t,i-1}^{\left(2\right)-1}\left(\nabla_{\vSigma_{t,i-1}}\ell\right)\vpsi_{t,i-1}^{\left(2\right)-1}\\
 & =2\vSigma_{t,i-1}\left(\nabla_{\vmu_{t,i-1}}\ell\right)\vmu_{t,i-1}^{\trans}+2\vSigma_{t,i-1}\left(\nabla_{\vSigma_{t,i-1}}\ell\right)\vSigma_{t,i-1}
\end{align}
Therefore
\begin{align}
\nabla_{\vpsi_{t,i-1}^{\left(1\right)}}\expectQ{\log p\left(\vy_{t}\vert f_{t}\left(\vtheta_{t}\right)\right)}{\vtheta_{t}\sim q_{\vpsi_{t,i-1}}} & =\vSigma_{t,i-1}\vg_{t,i}\label{eq:FC-NLL-psi1}\\
\nabla_{\vpsi_{t,i-1}^{\left(2\right)}}\expectQ{\log p\left(\vy_{t}\vert f_{t}\left(\vtheta_{t}\right)\right)}{\vtheta_{t}\sim q_{\vpsi_{t,i-1}}} & =2\vSigma_{t,i-1}\vg_{t,i}\vmu_{t,i-1}^{\trans}+\vSigma_{t,i-1}\vG_{t,i}\vSigma_{t,i-1}\label{eq:FC-NLL-psi2}
\end{align}
and
\begin{align}
    \nabla_{\vpsi_{t,i-1}^{\left(1\right)}}\KLpq{q_{\vpsi_{t,i-1}}}{q_{\vpsi_{t|t-1}}} 
    & =\vSigma_{t,i-1}\vSigma_{t|t-1}^{-1}\left(\vmu_{t,i-1}-\vmu_{t|t-1}\right)\label{eq:FC-KL-psi1}\\
    \nabla_{\vpsi_{t,i-1}^{\left(2\right)}}\KLpq{q_{\vpsi_{t,i-1}}}{q_{\vpsi_{t|t-1}}} 
    & =2\vSigma_{t,i-1}\vSigma_{t|t-1}^{-1}\left(\vmu_{t,i-1}-\vmu_{t|t-1}\right)\vmu_{t,i-1}^{\trans} \nonumber\\
    &\quad +\vSigma_{t,i-1}\left(\vSigma_{t|t-1}^{-1}-\vSigma_{t,i-1}^{-1}\right)\vSigma_{t,i-1}\label{eq:FC-KL-psi2}
\end{align}

\subsubsection{\Bong FC (explicit \RVGA)}
\label{sec:FC-Nat-bong}

Substituting \cref{eq:FC-NLL-rho1,eq:FC-NLL-rho2} into \cref{eq:bong-MD}
gives
\begin{align}
\vpsi_{t}^{\left(1\right)} & =\vpsi_{t\vert t-1}^{\left(1\right)}+\vg_{t}-\vG_{t}\vmu_{t|t-1}\\
\vpsi_{t}^{\left(2\right)} & =\vpsi_{t\vert t-1}^{\left(2\right)}+\tfrac{1}{2}\vG_{t}
\end{align}
Translating to $\left(\vmu_{t},\vSigma_{t}\right)$ gives the \bong-\fc update
\begin{align}
\vmu_{t} & =\vmu_{t|t-1}+\vSigma_{t}\vg_{t}\label{eq:bong-FC-mean}\\
\vSigma_{t}^{-1} & =\vSigma_{t\vert t-1}^{-1}-\vG_{t}\label{eq:bong-FC-prec}
\end{align}
This is equivalent to the explicit update form of RVGA \citep{RVGA}.
Using $\GMCH_{t}$
this update takes $O(\nparam^{3})$ because of the matrix inversion. 
Using $\GMCEF_{t}$ and the Woodbury matrix identity
we can write the update in a form that takes $O(\nsample\nparam^{2}+\nsample^{3})$:
\begin{align}
\vmu_{t} & =\vmu_{t|t-1}+\vK_{t}\bm{1}_{\nsample}\\
\vSigma_{t} & =\vSigma_{t|t-1}-\vK_{t}\gradmatT_{t} \vSigma_{t|t-1}\\
\vK_{t} & =\vSigma_{t|t-1}\gradmat_{t}\left(\nsample\vI_{\nsample}+ \gradmatT_{t} \vSigma_{t|t-1}\gradmat_{t}\right)^{-1}
\label{eq:bong-FC-K}
\end{align}
Likewise using $\GLEF_t$ takes $O(P^2)$:
\begin{align}
\vmu_{t} 
    &= \vmu_{t|t-1} + \vK_{t}\\
\vSigma_{t} 
    &= \vSigma_{t|t-1} - \vK_{t} \left(\gL_{t}\right)^\trans \vSigma_{t|t-1}\\
\vK_{t} 
    &= \frac
        {\vSigma_{t|t-1}\gL_{t}}
        {1 + \left(\gL_{t}\right)^\trans \vSigma_{t|t-1} \gL_t}
    \label{eq:bong-lin-EF-FC-K}
\end{align}

\subsubsection{\Bong-\hesslin FC (\CMEKF)}
\label{sec:lbong-fc}

Applying \cref{thm:lbong} to \cref{eq:bong-FC-mean,eq:bong-FC-prec} gives the \bong-\hesslin-\fc update
\begin{align}
\vmu_{t} & =\vmu_{t|t-1}+\vSigma_{t}\vH_{t}^{\trans}\vR_{t}^{-1}\left(\vy_{t}-\hat{\vy}_{t}\right)\\
\vSigma_{t}^{-1} & =\vSigma_{t\vert t-1}^{-1}+\vH_{t}^{\trans}\vR_{t}^{-1}\vH_{t}
\end{align}
This is equivalent to \CMEKF \citep{Tronarp2018,Ollivier2018}
and can be rewritten using the Woodbury identity in a form that takes $O(\nout\nparam^{2}+\nout^{3})$: 
\begin{align}
\vmu_{t} & =\vmu_{t|t-1}+\vK_{t}(\vy_{t}-\hat{\vy}_{t})\\
\vSigma_{t} & =\vSigma_{t|t-1}-\vK_{t}\vH_{t}\vSigma_{t|t-1}\\
\vK_{t} & =\vSigma_{t|t-1}\vH_{t}^{\trans}\left(\vR_{t}+\vH_{t}\vSigma_{t|t-1}\vH_{t}^{\trans}\right)^{-1} 
\end{align}

\subsubsection{\Blr FC}

Substituting \cref{eq:FC-NLL-rho1,eq:FC-NLL-rho2,eq:FC-KL-rho1,eq:FC-KL-rho2}
into \cref{eq:blr-MD} gives
\begin{align}
\vpsi_{t,i}^{\left(1\right)} & =\vpsi_{t,i-1}^{\left(1\right)}+\alpha\left(\vg_{t,i}-\vG_{t,i}\vmu_{t,i-1}-\vSigma_{t,i-1}^{-1}\vmu_{t,i-1}+\vSigma_{t|t-1}^{-1}\vmu_{t|t-1}\right)\\
\vpsi_{t,i}^{\left(2\right)} & =\vpsi_{t,i-1}^{\left(2\right)}+\frac{\alpha}{2}\left(\vG_{t,i}+\vSigma_{t,i-1}^{-1}-\vSigma_{t|t-1}^{-1}\right)
\end{align}
Translating to $\left(\vmu_{t,i},\vSigma_{t,i}\right)$ gives the
\blr-\fc update
\begin{align}
\vmu_{t,i} & =\vmu_{t,i-1}+\alpha\vSigma_{t,i}\vSigma_{t|t-1}^{-1}\left(\vmu_{t|t-1}-\vmu_{t,i-1}\right)+\alpha\vSigma_{t,i}\vg_{t,i}\label{eq:blr-FC-nat-mu}\\
\vSigma_{t,i}^{-1} & =\left(1-\alpha\right)\vSigma_{t,i-1}^{-1}+\alpha\vSigma_{t|t-1}^{-1}-\alpha\vG_{t,i}\label{eq:BLR-FC-Nat-prec}
\end{align}
This update takes $O(\nparam^3)$ per iteration because of the matrix inversion.
In \cref{sec:FC-Nat-bong} we were able to use the Woodbury identity to exploit the low rank of
$\GMCEF_t$ and $\GLEF_t$
and obtain \bong updates with complexity quadratic in $\nparam$.
This does not appear possible with \cref{eq:BLR-FC-Nat-prec} because of the extra precision term on the RHS
(applying Woodbury would require inverting $(1-\alpha) \vSigma_{t,i-1}^{-1} + \alpha \vSigma_{t|t-1}^{-1}$).
Therefore unlike \bong-\fc, \blr-\fc requires time cubic in the model size,
for reasons that can be traced back to the KL term in \cref{eq:blr-MD}.

\subsubsection{\Blr-\hesslin FC}

Applying \cref{thm:lbong} to \cref{eq:blr-FC-nat-mu,eq:BLR-FC-Nat-prec} gives the \blr-\hesslin-\fc update
\begin{align}
\vmu_{t,i} & =\vmu_{t,i-1}+\alpha\vSigma_{t,i}\vSigma_{t|t-1}^{-1}\left(\vmu_{t|t-1}-\vmu_{t,i-1}\right)+\alpha\vSigma_{t,i}\vH_{t,i}^{\trans}\vR_{t,i}^{-1}\left(\vy_{t}-\hat{\vy}_{t,i}\right)\\
\vSigma_{t,i}^{-1} & =\left(1-\alpha\right)\vSigma_{t,i-1}^{-1}+\alpha\vSigma_{t|t-1}^{-1}+\alpha\vH_{t,i}^{\trans}\vR_{t,i}^{-1}\vH_{t,i}
\end{align}
This update takes $O(\nparam^3)$ per iteration because of the matrix inversion.

\subsubsection{\Bog FC}

Substituting \cref{eq:FC-NLL-psi1,eq:FC-NLL-psi2} into \cref{eq:bog}
gives
\begin{align}
\vpsi_{t}^{\left(1\right)} & =\vpsi_{t\vert t-1}^{\left(1\right)}+\alpha\vSigma_{t|t-1}\vg_{t}\\
\vpsi_{t}^{\left(2\right)} & =\vpsi_{t\vert t-1}^{\left(2\right)}+2\alpha\vSigma_{t|t-1}\vg_{t}\vmu_{t|t-1}^{\trans}+\alpha\vSigma_{t|t-1}\vG_{t}\vSigma_{t|t-1}
\end{align}
Translating to $\left(\vmu_{t},\vSigma_{t}\right)$ gives the \bog-\fc
update
\begin{align}
\vmu_{t} & =\vSigma_{t}\vSigma_{t|t-1}^{-1}\vmu_{t|t-1}+\alpha\vSigma_{t}\vSigma_{t|t-1}\vg_{t}
\label{eq:bog-FC-nat-mu}\\
\vSigma_{t}^{-1} & =\vSigma_{t|t-1}^{-1}-4\alpha\vSigma_{t|t-1}\vg_{t}\vmu_{t|t-1}^{\trans}-2\alpha\vSigma_{t|t-1}\vG_{t}\vSigma_{t|t-1}
\label{eq:bog-FC-nat-prec}
\end{align}
This update takes $O(\nparam^3)$ because of the matrix inversion.
The greater cost of the \bog-\fc update relative to \bong-\fc can be traced to the difference between GD and NGD: the NLL gradients wrt $\vpsi_{t|t-1}$ in \cref{eq:FC-NLL-psi1,eq:FC-NLL-psi2} are more complicated than the gradients wrt $\vrho_{t|t-1}$ in \cref{eq:FC-NLL-rho1,eq:FC-NLL-rho2}.

\subsubsection{\Bog-\hesslin FC}

Applying \cref{thm:lbong} to \cref{eq:bog-FC-nat-mu,eq:bog-FC-nat-prec} gives the \bog-\hesslin-\fc update
\begin{align}
\vmu_{t} & =\vSigma_{t}\vSigma_{t|t-1}^{-1}\vmu_{t|t-1}+\alpha\vSigma_{t}\vSigma_{t|t-1}\vH_{t}^{\trans}\vR_{t}^{-1}\left(\vy_{t}-\hat{\vy}_{t}\right)\\
\vSigma_{t}^{-1} & =\vSigma_{t|t-1}^{-1}-4\alpha\vSigma_{t|t-1}\vH_{t}^{\trans}\vR_{t}^{-1}\left(\vy_{t}-\hat{\vy}_{t}\right)\vmu_{t|t-1}^{\trans}+2\alpha\vSigma_{t|t-1}\vH_{t}^{\trans}\vR_{t}^{-1}\vH_{t}\vSigma_{t|t-1}
\end{align}
This update takes $O(\nparam^3)$ because of the matrix inversion.

\subsubsection{\BBB FC}

Substituting \cref{eq:FC-NLL-psi1,eq:FC-NLL-psi2,eq:FC-KL-psi1,eq:FC-KL-psi2}
into \cref{eq:bbb} gives
\begin{align}
\vpsi_{t,i}^{\left(1\right)} & =\vpsi_{t,i-1}^{\left(1\right)}+\alpha\vSigma_{t,i-1}\vg_{t,i}-\alpha\vSigma_{t,i-1}\vSigma_{t|t-1}^{-1}\left(\vmu_{t,i-1}-\vmu_{t|t-1}\right)\\
\vpsi_{t}^{\left(2\right)} 
    & =\vpsi_{t,i-1}^{\left(2\right)}
    +2\alpha\vSigma_{t,i-1}\vg_{t,i}\vmu_{t,i-1}^{\trans}+\alpha\vSigma_{t|t-1}\vG_{t}\vSigma_{t|t-1}
    \nonumber\\
    &\quad -2\alpha\vSigma_{t,i-1}\vSigma_{t|t-1}^{-1}\left(\vmu_{t,i-1}-\vmu_{t|t-1}\right)\vmu_{t,i-1}^{\trans}
    -\alpha\vSigma_{t,i-1}\left(\vSigma_{t|t-1}^{-1}-\vSigma_{t,i-1}^{-1}\right)\vSigma_{t,i-1}
\end{align}
Translating to $\left(\vmu_{t,i},\vSigma_{t,i}\right)$ gives the \bbb-\fc update
\begin{align}
\vmu_{t,i} & =\vSigma_{t,i}\vSigma_{t,i-1}^{-1}\vmu_{t,i-1}+\alpha\vSigma_{t,i}\vSigma_{t,i-1}\left(\vg_{t,i}+\vSigma_{t|t-1}^{-1}\left(\vmu_{t|t-1}-\vmu_{t,i-1}\right)\right)
\label{eq:bbb-FC-nat-mu}\\
\vSigma_{t,i}^{-1} 
    & =\vSigma_{t,i-1}^{-1}
    -2\alpha\vSigma_{t,i-1}
    \left( \begin{array}{c}
         2\vSigma_{t|t-1}^{-1}\left(\vmu_{t|t-1}-\vmu_{t,i-1}\right)\vmu_{t,i-1}^{\trans} 
         +\vI_{\nparam}\\
         +2\vg_{t,i}\vmu_{t,i-1}^{\trans}
         +\left(\vG_{t,i}-\vSigma_{t|t-1}^{-1}\right)\vSigma_{t,i-1}
    \end{array} \right)
    \label{eq:bbb-FC-nat-prec}
\end{align}
This update takes $O(\nparam^3)$ per iteration because of the matrix inversion.

\subsubsection{\BBB-\hesslin FC}

Applying \cref{thm:lbong} to \cref{eq:bbb-FC-nat-mu,eq:bbb-FC-nat-prec} gives the \bbb-\hesslin-\fc update
\begin{align}
\vmu_{t,i} 
    & =\vSigma_{t,i}\vSigma_{t,i-1}^{-1}\vmu_{t,i-1}
    \nonumber\\
    &\quad +\alpha\vSigma_{t,i}\vSigma_{t,i-1}\left(\vH_{t,i}^{\trans}\vR_{t,i}^{-1}\left(\vy_{t}-\hat{\vy}_{t,i}\right)+\vSigma_{t|t-1}^{-1}\left(\vmu_{t|t-1}-\vmu_{t,i-1}\right)\right)\\
\vSigma_{t,i}^{-1} 
    & =\vSigma_{t,i-1}^{-1} 
    -2\alpha\vSigma_{t,i-1}
    \left( \begin{array}{c}
        2\vSigma_{t|t-1}^{-1}\left(\vmu_{t|t-1}-\vmu_{t,i-1}\right)\vmu_{t,i-1}^{\trans} 
        +\vI_{\nparam} \\
        +2\vH_{t,i}^{\trans}\vR_{t,i}^{-1}\left(\vy_{t}-\hat{\vy}_{t,i}\right)\vmu_{t,i-1}^{\trans} \\
        -\left(\vH_{t,i}^{\trans}\vR_{t,i}^{-1}\vH_{t,i}+\vSigma_{t|t-1}^{-1}\right)\vSigma_{t,i-1}
    \end{array} \right)
\end{align}
This update takes $O(\nparam^3)$ per iteration because of the matrix inversion.

%% file: sections/derivations-FC-mom.tex
\subsection{Full covariance Gaussian, Moment parameters\label{sec:FC-Mom}}

The Bonnet and Price theorems give
\begin{align}
\nabla_{\vmu_{t,i-1}}\expectQ{\log p\left(\vy_{t}|f_{t}\left(\vtheta_{t}\right)\right)}{\vtheta_{t}\sim q_{\vpsi_{t,i-1}}} & =\vg_{t,i}\label{eq:FC-NLL-mu}\\
\nabla_{\vSigma_{t,i-1}}\expectQ{\log p\left(\vy_{t}|f_{t}\left(\vtheta_{t}\right)\right)}{\vtheta_{t}\sim q_{\vpsi_{t,i-1}}} & =\tfrac{1}{2}\vG_{t,i}\label{eq:FC-NLL-Sigma}
\end{align}
From \cref{sec:FC-Nat} we have
\begin{align}
\nabla_{\vmu_{t,i-1}}\KLpq{q_{\vpsi_{t,i-1}}}{q_{\vpsi_{t|t-1}}} & =\vSigma_{t|t-1}^{-1}\left(\vmu_{t,i-1}-\vmu_{t|t-1}\right)\label{eq:FC-KL-mu}\\
\nabla_{\vSigma_{t,i-1}}\KLpq{q_{\vpsi_{t,i-1}}}{q_{\vpsi_{t|t-1}}} & =\tfrac{1}{2}\left(\vSigma_{t|t-1}^{-1}-\vSigma_{t,i-1}^{-1}\right)\label{eq:FC-KL-Sigma}
\end{align}
We write the Fisher with respect to the moment parameters $\vpsi=(\vmu,{\rm vec}(\vSigma))$
as a block matrix: 
\begin{equation}
\vF=\left[\begin{array}{cc}
\vF_{\vmu,\vmu} & \vF_{\vmu,\vSigma}\\
\vF_{\vSigma,\vmu} & \vF_{\vSigma,\vSigma}
\end{array}\right]
\end{equation}
The blocks can be calculated by the second-order Fisher formula 
\begin{align}
\vF_{\vmu,\vmu} & =-\expectQ{\nabla_{\vmu,\vmu}\log q_{\vpsi}(\vtheta)}{q_{\vpsi}}\\
 & =\vSigma^{-1}\\
\vF_{\vmu,\vSigma} & =-\expectQ{\nabla_{\vmu,\vSigma}\log q_{\vpsi}(\vtheta)}{q_{\vpsi}}\\
 & =-\expectQ{\left(\nabla_{\vSigma}\vSigma^{-1}\right)(\vtheta-\vmu)}{q_{\vpsi}}\\
 & =\bm{0}\\
\vF_{\vSigma,\vSigma} & =-\expectQ{\nabla_{\vSigma,\vSigma}\log q_{\vpsi}(\vtheta)}{q_{\vpsi}}\\
 & =-\expectQ{\nabla_{\vSigma}\left(\tfrac{1}{2}\vSigma^{-1}(\vtheta-\vmu)(\vtheta-\vmu)^{\trans}\vSigma^{-1}-\tfrac{1}{2}\vSigma^{-1}\right)}{q_{\vpsi}}\\
    & =-\tfrac{1}{2} 
    \expectQ
        { \begin{array}{c}
             \left(\nabla_{\vSigma}\vSigma^{-1}\right)(\vtheta-\vmu)(\vtheta-\vmu)^{\trans}\vSigma^{-1} \\
             +\vSigma^{-1}(\vtheta-\vmu)(\vtheta-\vmu)^{\trans}\left(\nabla_{\vSigma}\vSigma^{-1}\right)
             -\left(\nabla_{\vSigma}\vSigma^{-1}\right)
        \end{array}}
        {q_{\vpsi}}\\
 & =-\tfrac{1}{2}\nabla_{\vSigma}\vSigma^{-1}\\
 & =\tfrac{1}{2}\vSigma^{-1}\otimes\vSigma^{-1}
\end{align}
 In the final line we used 
\begin{align}
\nabla_{\Sigma_{k\ell}}\left(\vSigma^{-1}\right)_{ij} & =-\left(\vSigma^{-1}\right)_{ik}\left(\vSigma^{-1}\right)_{j\ell}\\
 & =-\left(\vSigma^{-1}\otimes\vSigma^{-1}\right)_{ij,k\ell}
\end{align}
 with $ij$ and $k\ell$ treated as composite indices in $[\nparam^{2}]$.
Therefore the preconditioner in the NGD methods is 
\begin{align}
\vF_{\vpsi_{t,i-1}}^{-1} & =\left[\begin{array}{cc}
\vSigma_{t,i-1} & \bm{0}\\
\bm{0} & 2\vSigma_{t,i-1}\otimes\vSigma_{t,i-1}
\end{array}\right]\label{eq:FC-mom-Fisher}
\end{align}

\subsubsection{\Bong FC, Moment}

Substituting \cref{eq:FC-NLL-mu,eq:FC-NLL-Sigma,eq:FC-mom-Fisher}
into \cref{eq:bong} gives the \bong-\fcmom update
\begin{align}
\vmu_{t} & =\vmu_{t\vert t-1}+\vSigma_{t|t-1}\vg_{t}
\label{eq:bong-FC-mom-mu}\\
\vSigma_{t} & =\vSigma_{t\vert t-1}+\vSigma_{t\vert t-1}\vG_{t}\vSigma_{t\vert t-1}
\label{eq:bong-FC-mom-cov}
\end{align}
This update takes $O(\nparam^3)$ using $\GMCH_t$,
$O(\nsample\nparam^2)$ using $\GMCEF_t$,
and $O(\nparam^2)$ using $\GLEF_t$.
The update is similar to the \bong-\fc update except that \cref{eq:bong-FC-mom-cov} ignores the 
$\gradmatT_t \vSigma_{t|t-1} \gradmat_t$
term in \cref{eq:bong-FC-K}
or the $\left(\gL_{t}\right)^\trans \vSigma_{t|t-1} \gL_t$
term in \cref{eq:bong-lin-EF-FC-K}
which estimate the epistemic part of predictive uncertainty.

\subsubsection{\Bong-\hesslin FC, Moment}

Applying \cref{thm:lbong} to \cref{eq:bong-FC-mom-mu,eq:bong-FC-mom-cov} gives the \bong-\hesslin-\fcmom update
\begin{align}
\vmu_{t} & =\vmu_{t\vert t-1}+\vSigma_{t|t-1}\vH_{t}^{\trans}\vR_{t}^{-1}(\vy_{t}-\hat{\vy}_{t})\\
\vSigma_{t} & =\vSigma_{t\vert t-1}-\vSigma_{t\vert t-1}\vH_{t}^{\trans}\vR_{t}^{-1}\vH_{t}\vSigma_{t\vert t-1}
\end{align}
This update takes $O(\nout\nparam^2)$.

\subsubsection{\Blr FC, Moment}

Substituting \cref{eq:FC-NLL-mu,eq:FC-NLL-Sigma,eq:FC-KL-mu,eq:FC-KL-Sigma,eq:FC-mom-Fisher}
into \cref{eq:blr} gives the \blr-\fcmom update
\begin{align}
\vmu_{t,i} & =\vmu_{t,i-1}+\alpha\vSigma_{t,i-1}\vSigma_{t|t-1}^{-1}\left(\vmu_{t|t-1}-\vmu_{t,i-1}\right)+\alpha\vSigma_{t,i-1}\vg_{t,i}
\label{eq:blr-FC-mom-mu}\\
\vSigma_{t,i} & =\left(1+\alpha\right)\vSigma_{t,i-1}+\alpha\vSigma_{t,i-1}\left(\vG_{t,i}-\vSigma_{t|t-1}^{-1}\right)\vSigma_{t,i-1}
\label{eq:blr-FC-mom-cov}
\end{align}
This update takes $O(\nparam^3)$ per iteration because of the matrix inversion.
Using 
$\GMCEF_t = -\frac{1}{\nsample} \gradmat_t \gradmatT_t$
or $\GLEF = -\gL_t\left(\gL_{t}\right)^\trans$
allows the \bong-\fcmom covariance update in \cref{eq:bong-FC-mom-cov} to scale quadratically,
but this does not help here.
Instead the \blr-\fcmom update scales cubically because of the additional 
$\vSigma^{-1}_{t|t-1}$
term that comes from the KL divergence in the VI objective.

\subsubsection{\Blr-\hesslin FC, Moment}

Applying \cref{thm:lbong} to \cref{eq:blr-FC-mom-mu,eq:blr-FC-mom-cov} gives the \blr-\hesslin-\fcmom update
\begin{align}
\vmu_{t,i} & =\vmu_{t,i-1}+\alpha\vSigma_{t,i-1}\vSigma_{t|t-1}^{-1}\left(\vmu_{t|t-1}-\vmu_{t,i-1}\right)+\alpha\vSigma_{t,i-1}\vH_{t,i}^{\trans}\vR_{t,i}^{-1}(\vy_{t}-\hat{\vy}_{t,i})\\
\vSigma_{t,i} & =\left(1+\alpha\right)\vSigma_{t,i-1}-\alpha\vSigma_{t,i-1}\left(\vSigma_{t|t-1}^{-1}+\vH_{t,i}^{\trans}\vR_{t,i}^{-1}\vH_{t,i}\right)\vSigma_{t,i-1}
\end{align}
This update takes $O(\nparam^3)$ per iteration because of the matrix inversion in the
$\vSigma^{-1}_{t|t-1}$
term that comes from the KL divergence in the VI objective.

\subsubsection{\Bog FC, Moment}

Substituting \cref{eq:FC-NLL-mu,eq:FC-NLL-Sigma} into \cref{eq:bog}
gives the \bog-\fcmom update
\begin{align}
\vmu_{t} & =\vmu_{t\vert t-1}+\alpha\vg_{t}
\label{eq:bog-fc-mom-mu}\\
\vSigma_{t} & =\vSigma_{t\vert t-1}+\frac{\alpha}{2}\vG_{t}
\label{eq:bog-fc-mom-cov}
\end{align}
Note the mean update is vanilla 
online gradient descent (OGD)
and does not depend on the covariance.
This update takes $O(\nsample\nparam^2)$ using $\GMCH_t$ or $\GMCEF_t$
and $O(\nparam^2)$ using $\GLEF_t$.

\subsubsection{\Bog-\hesslin FC, Moment}

Applying \cref{thm:lbong} to \cref{eq:bog-fc-mom-mu,eq:bog-fc-mom-cov} gives the \bog-\hesslin-\fcmom update
\begin{align}
\vmu_{t} & =\vmu_{t\vert t-1}+\alpha\vH_{t}^{\trans}\vR_{t}^{-1}(\vy_{t}-\hat{\vy}_{t})
\label{eq:lbog-fc-mom-mu}\\
\vSigma_{t} & =\vSigma_{t\vert t-1}-\frac{\alpha}{2}\vH_{t}^{\trans}\vR_{t}^{-1}\vH_{t}
\label{eq:lbog-fc-mom-cov}
\end{align}
This update takes $O(\nout\nparam^2)$.

\subsubsection{\BBB FC, Moment}

Substituting \cref{eq:FC-NLL-mu,eq:FC-NLL-Sigma,eq:FC-KL-mu,eq:FC-KL-Sigma}
into \cref{eq:bbb} gives the \bbb-\fcmom
\begin{align}
\vmu_{t,i} & =\vmu_{t,i-1}+\alpha\vSigma_{t|t-1}^{-1}\left(\vmu_{t|t-1}-\vmu_{t,i-1}\right)+\alpha\vg_{t,i}
\label{bbb-FC-mom-mu}\\
\vSigma_{t,i} & =\vSigma_{t,i-1}+\frac{\alpha}{2}\left(\vSigma_{t,i-1}^{-1}-\vSigma_{t|t-1}^{-1}+\vG_{t,i}\right)
\label{bbb-FC-mom-cov}
\end{align}
This update takes $O(\nparam^3)$ per iteration because of the matrix inversion, which traces back to the VI objective.
Comparing to the \bog-\fcmom update in \cref{eq:bog-fc-mom-mu,eq:bog-fc-mom-cov} (which has quadratic complexity in $\nparam$),
the extra terms here come from the KL part of \cref{eq:bbb}.

\subsubsection{\BBB-\hesslin FC, Moment}

Applying \cref{thm:lbong} to \cref{bbb-FC-mom-mu,bbb-FC-mom-cov} gives the \bbb-\hesslin-\fcmom update
\begin{align}
\vmu_{t,i} & =\vmu_{t,i-1}+\alpha\vSigma_{t|t-1}^{-1}\left(\vmu_{t|t-1}-\vmu_{t,i-1}\right)+\alpha\vH_{t,i}^{\trans}\vR_{t,i}^{-1}(\vy_{t}-\hat{\vy}_{t,i})\\
\vSigma_{t} & =\vSigma_{t,i-1}+\frac{\alpha}{2}\left(\vSigma_{t,i-1}^{-1}-\vSigma_{t|t-1}^{-1}-\vH_{t,i}^{\trans}\vR_{t,i}^{-1}\vH_{t,i}\right)
\end{align}
This update takes $O(\nparam^3)$ per iteration because of the matrix inversion.
Comparing to the \bog-\hesslin-\fcmom update in \cref{eq:lbog-fc-mom-mu,eq:lbog-fc-mom-cov} (which has quadratic complexity in $\nparam$),
the extra terms here come from the KL part of \cref{eq:bbb}.

%% file: sections/derivations-diag.tex
\subsection{Diagonal Gaussian, Natural parameters\label{sec:diag-Nat}}

Throughout this subsection, vector multiplication and exponents are elementwise.

The natural and dual parameters for a diagonal Gaussian are given by
\begin{alignat}{3}
\vpsi_{t,i-1}^{\left(1\right)}
&=\vsigma_{t,i-1}^{-2}\vmu_{t,i-1} 
&\qquad \vrho_{t,i-1}^{\left(1\right)}
&=\vmu_{t,i-1}\\
\vpsi_{t,i-1}^{\left(2\right)}
&=-\tfrac{1}{2}\vsigma_{t,i-1}^{-2} 
& \vrho_{t,i-1}^{\left(2\right)}
&=\vmu_{t,i-1}\vmu_{t,i-1}^{\trans}+\vsigma_{t,i-1}^{2}
\end{alignat}
Inverting these relationships gives
\begin{alignat}{3}
\vmu_{t,i-1} 
& =-\tfrac{1}{2}\left(\vpsi_{t,i-1}^{\left(2\right)}\right)^{-1}\vpsi_{t,i-1}^{\left(1\right)}
&&=\vrho_{t,i-1}^{\left(1\right)}\\
\vsigma_{t,i-1}^{2} 
& =-\tfrac{1}{2}\left(\vpsi_{t,i-1}^{\left(2\right)}\right)^{-1}
&&=\vrho_{t,i-1}^{\left(2\right)}-\left(\vrho_{t,i-1}^{\left(1\right)}\right)^{2}
\end{alignat}

The KL divergence in the VI loss is
\begin{equation}
\KLpq{q_{\vpsi_{t,i-1}}}{q_{\vpsi_{t|t-1}}}=\tfrac{1}{2}\left(\vmu_{t,i-1}-\vmu_{t|t-1}\right)^{2}\vsigma_{t|t-1}^{-2}+\tfrac{1}{2}\sum\left(\vsigma_{t|t-1}^{-2}\vsigma_{t,i-1}^{2}-\log\vsigma_{t,i-1}^{2}\right)+\const
\end{equation}
with gradients
\begin{align}
\nabla_{\vmu_{t,i-1}}\KLpq{q_{\vpsi_{t,i-1}}}{q_{\vpsi_{t|t-1}}} & =\vsigma_{t|t-1}^{-2}\left(\vmu_{t,i-1}-\vmu_{t|t-1}\right)\\
\nabla_{\vsigma_{t,i-1}^{2}}\KLpq{q_{\vpsi_{t,i-1}}}{q_{\vpsi_{t|t-1}}} & =\tfrac{1}{2}\left(\vsigma_{t|t-1}^{-2}-\vsigma_{t,i-1}^{-2}\right)
\end{align}

For any scalar function $\vell$ the chain rule gives
\begin{align}
\nabla_{\vrho_{t,i-1}^{\left(1\right)}}\ell & =\frac{\partial\vmu_{t,i-1}}{\partial\vrho_{t,i-1}^{\left(1\right)}}\nabla_{\vmu_{t,i-1}}\ell+\frac{\partial\vsigma_{t,i-1}^{2}}{\partial\vrho_{t,i-1}^{\left(1\right)}}\nabla_{\vsigma_{t,i-1}^{2}}\ell\\
 & =\nabla_{\vmu_{t,i-1}}\ell-2\vmu_{t,i-1}\nabla_{\vsigma_{t,i-1}^{2}}\ell\\
\nabla_{\vrho_{t,i-1}^{\left(2\right)}}\ell & =\frac{\partial\vmu_{t,i-1}}{\partial\vrho_{t,i-1}^{\left(2\right)}}\nabla_{\vmu_{t,i-1}}\ell+\frac{\partial\vsigma_{t,i-1}^{2}}{\partial\vrho_{t,i-1}^{\left(2\right)}}\nabla_{\vsigma_{t,i-1}^{2}}\ell\\
 & =\nabla_{\vsigma_{t,i-1}^{2}}\ell
\end{align}
Therefore
\begin{align}
\nabla_{\vrho_{t,i-1}^{\left(1\right)}}\expectQ{\log p\left(\vy_{t}\vert f_{t}\left(\vtheta_{t}\right)\right)}{\vtheta_{t}\sim q_{\vpsi_{t,i-1}}} & =\vg_{t,i}-\diag\left(\vG_{t,i}\right)\vmu_{t,i-1}\label{eq:diag-NLL-rho1}\\
\nabla_{\vrho_{t,i-1}^{\left(2\right)}}\expectQ{\log p\left(\vy_{t}\vert f_{t}\left(\vtheta_{t}\right)\right)}{\vtheta_{t}\sim q_{\vpsi_{t,i-1}}} & =\tfrac{1}{2}\diag\left(\vG_{t,i}\right)\label{eq:diag-NLL-rho2}
\end{align}
and
\begin{align}
\nabla_{\vrho_{t,i-1}^{\left(1\right)}}\KLpq{q_{\vpsi_{t,i-1}}}{q_{\vpsi_{t|t-1}}} & =\vsigma_{t,i-1}^{-2}\vmu_{t,i-1}-\vsigma_{t|t-1}^{-2}\vmu_{t|t-1}\label{eq:diag-KL-rho1}\\
\nabla_{\vrho_{t,i-1}^{\left(2\right)}}\KLpq{q_{\vpsi_{t,i-1}}}{q_{\vpsi_{t|t-1}}} & =\tfrac{1}{2}\left(\vsigma_{t|t-1}^{-2}-\vsigma_{t,i-1}^{-2}\right)\label{eq:diag-KL-rho2}
\end{align}

Following the same approach for $\vpsi$ gives 
\begin{align}
\nabla_{\vpsi_{t,i-1}^{\left(1\right)}}\ell & =\frac{\partial\vmu_{t,i-1}}{\partial\vpsi_{t,i-1}^{\left(1\right)}}\nabla_{\vmu_{t,i-1}}\ell+\frac{\partial\vsigma_{t,i-1}^{2}}{\partial\vpsi_{t,i-1}^{\left(1\right)}}\nabla_{\vSigma_{t,i-1}}\ell\\
 & =-\tfrac{1}{2}\left(\vpsi_{t,i-1}^{\left(2\right)}\right)^{-1}\nabla_{\vmu_{t,i-1}}\ell\\
 & =\vsigma_{t,i-1}^{2}\nabla_{\vmu_{t,i-1}}\ell\\
\nabla_{\vpsi_{t,i-1}^{\left(2\right)}}\ell & =\frac{\partial\vmu_{t,i-1}}{\partial\vpsi_{t,i-1}^{\left(2\right)}}\nabla_{\vmu_{t,i-1}}\ell+\frac{\partial\vsigma_{t,i-1}^{2}}{\partial\vpsi_{t,i-1}^{\left(2\right)}}\nabla_{\vsigma_{t,i-1}^{2}}\ell\\
 & =\tfrac{1}{2}\left(\vpsi_{t,i-1}^{\left(2\right)}\right)^{-2}\vpsi_{t,i-1}^{\left(1\right)}\nabla_{\vmu_{t,i-1}}\ell+\tfrac{1}{2}\left(\vpsi_{t,i-1}^{\left(2\right)}\right)^{-2}\nabla_{\vsigma_{t,i-1}^{2}}\ell\\
 & =2\vsigma_{t,i-1}^{2}\vmu_{t,i-1}\nabla_{\vmu_{t,i-1}}\ell+2\vsigma_{t,i-1}^{4}\nabla_{\vsigma_{t,i-1}^{2}}\ell
\end{align}
Therefore
\begin{align}
\nabla_{\vpsi_{t,i-1}^{\left(1\right)}}\expectQ{\log p\left(\vy_{t}\vert f_{t}\left(\vtheta_{t}\right)\right)}{\vtheta_{t}\sim q_{\vpsi_{t,i-1}}} & =\vsigma_{t,i-1}^{2}\vg_{t,i}\label{eq:diag-NLL-psi1}\\
\nabla_{\vpsi_{t,i-1}^{\left(2\right)}}\expectQ{\log p\left(\vy_{t}\vert f_{t}\left(\vtheta_{t}\right)\right)}{\vtheta_{t}\sim q_{\vpsi_{t,i-1}}} & =2\vsigma_{t,i-1}^{2}\vmu_{t,i-1}\vg_{t,i}+\vsigma_{t,i-1}^{4}\diag\left(\vG_{t,i}\right)\label{eq:diag-NLL-psi2}
\end{align}
and
\begin{align}
\nabla_{\vpsi_{t,i-1}^{\left(1\right)}}\KLpq{q_{\vpsi_{t,i-1}}}{q_{\vpsi_{t|t-1}}} & =\vsigma_{t,i-1}^{2}\vsigma_{t|t-1}^{-2}\left(\vmu_{t,i-1}-\vmu_{t|t-1}\right)\label{eq:diag-KL-psi1}\\
\nabla_{\vpsi_{t,i-1}^{\left(2\right)}}\KLpq{q_{\vpsi_{t,i-1}}}{q_{\vpsi_{t|t-1}}} & =2\vsigma_{t,i-1}^{2}\vsigma_{t|t-1}^{-2}\vmu_{t,i-1}\left(\vmu_{t,i-1}-\vmu_{t|t-1}\right)
\nonumber\\
&\qquad+\vsigma_{t,i-1}^{4}\left(\vsigma_{t|t-1}^{-2}-\vsigma_{t,i-1}^{-2}\right)\label{eq:diag-KL-psi2}
\end{align}

Our implementations often make use of the following trick:
Suppose $\vA \in \real^{n \times m}$
and $\vB \in \real^{m \times n}$.
Then we can efficiently compute
$\diag(\vA \vB)$ in $O(m n)$ time
using
$(\vA \vB)_{ii} = \sum_{j=1}^M
A_{ij} B_{ji}$.

For \hessmc methods,
we approximate the diagonal of the Hessian for each MC sample $\hat{\vtheta}_t^{(m)}$
using Hutchinson's trace estimation method \citep{hutchinson1989stochastic} 
which has been used in other DNN optimization papers such as adahessian \cite{adahessian}.
This involves an extra inner loop with size denoted $\nhutchinson$.

\subsubsection{\Bong Diag}

Substituting \cref{eq:diag-NLL-rho1,eq:diag-NLL-rho2} into \cref{eq:bong-MD}
gives
\begin{align}
\vpsi_{t}^{\left(1\right)} & =\vpsi_{t\vert t-1}^{\left(1\right)}+\vg_{t}-\diag\left(\vG_{t}\right)\vmu_{t|t-1}
\label{eq:bong-diag-mu}\\
\vpsi_{t}^{\left(2\right)} & =\vpsi_{t\vert t-1}^{\left(2\right)}+\tfrac{1}{2}\diag\left(\vG_{t}\right)
\label{eq:bong-diag-prec}
\end{align}
Translating to $\left(\vmu_{t},\vsigma_{t}^{2}\right)$ gives the
\bong-\dg update
\begin{align}
\vmu_{t} & =\vmu_{t|t-1}+\vsigma_{t}^{2}\vg_{t}\\
\vsigma_{t}^{-2} & =\vsigma_{t\vert t-1}^{-2}-\diag\left(\vG_{t}\right)
\label{eq:bong-diag-prec2}
\end{align}
This update takes $O(\nsample \nparam)$ to estimate $\vG_t$ using $\GMCEF_t$, $O(\nhutchinson \nsample \nparam)$ using $\GMCH_t$ and Hutchinson's method, and $O(\nparam)$ using $\GLEF_t$.

\subsubsection{\Bong-\hesslin Diag (\VDEKF)}

Applying \cref{thm:lbong} to \cref{eq:bong-diag-mu,eq:bong-diag-prec} gives the \bong-\hesslin-\dg update
\begin{align}
\vmu_{t} & =\vmu_{t|t-1}+\vsigma_{t}^{2}\left(\vH_{t}^{\trans}\vR_{t}^{-1}\left(\vy_{t}-\hat{\vy}_{t}\right)\right)\\
\vsigma_{t}^{-2} & =\vsigma_{t\vert t-1}^{-2}+\diag\left(\vH_{t}^{\trans}\vR_{t}^{-1}\vH_{t}\right)
\end{align}
This update is equivalent to \VDEKF \citep{diaglofi} and takes $O(\nout^2 \nparam)$.

\subsubsection{\Blr Diag (\VON)}

Substituting \cref{eq:diag-NLL-rho1,eq:diag-NLL-rho2,eq:diag-KL-rho1,eq:diag-KL-rho2}
into \cref{eq:blr-MD} gives
\begin{align}
\vpsi_{t,i}^{\left(1\right)} & =\vpsi_{t,i-1}^{\left(1\right)}+\alpha\left(\vg_{t,i}-\diag\left(\vG_{t,i}\right)\vmu_{t,i-1}-\vsigma_{t,i-1}^{-2}\vmu_{t,i-1}+\vsigma_{t|t-1}^{-2}\vmu_{t|t-1}\right)\\
\vpsi_{t,i}^{\left(2\right)} & =\vpsi_{t,i-1}^{\left(2\right)}+\frac{\alpha}{2}\left(\diag\left(\vG_{t,i}\right)+\vsigma_{t,i-1}^{-2}-\vsigma_{t|t-1}^{-2}\right)
\end{align}
Translating to $\left(\vmu_{t,i},\vsigma_{t,i}^{2}\right)$ gives
the \blr-\dg update
\begin{align}
\vmu_{t,i} & =\vmu_{t,i-1}+\alpha\vsigma_{t,i}^{2}\vsigma_{t|t-1}^{-2}\left(\vmu_{t|t-1}-\vmu_{t,i-1}\right)+\alpha\vsigma_{t,i}^{2}\vg_{t,i}
\label{eq:blr-diag-mu}\\
\vsigma_{t,i}^{-2} & =\left(1-\alpha\right)\vsigma_{t,i-1}^{-2}+\alpha\vsigma_{t|t-1}^{-2}-\alpha\,\diag\left(\vG_{t,i}\right)
\label{eq:blr-diag-prec}
\end{align}
This update takes $O(\nsample \nparam)$ per iteration to estimate $\vG_t$ using $\GMCEF_t$, $O(\nhutchinson \nsample \nparam)$ per iteration using $\GMCH_t$ and Hutchinson's method, and $O(\nparam)$ per iteration using $\GLEF_t$.

The \hessmc and \efmc versions of this update are respectively equivalent to \VON and \VOGN \citep{VON} in the batch setting where we replace $q_{\vpsi_{t|t-1}}$ with a spherical prior $\gauss (\bm{0},\lambda^{-1}\vI_{\nparam})$ (see \cref{sec:blr-batch}).

\subsubsection{\Blr-\hesslin Diag}

Applying \cref{thm:lbong} to \cref{eq:blr-diag-mu,eq:blr-diag-prec} gives the \blr-\hesslin-\dg update
\begin{align}
\vmu_{t,i} & =\vmu_{t,i-1}+\alpha\vsigma_{t,i}^{2}\vsigma_{t|t-1}^{-2}\left(\vmu_{t|t-1}-\vmu_{t,i-1}\right)+\alpha\vsigma_{t,i}^{2}\left(\vH_{t,i}^{\trans}\vR_{t,i}^{-1}\left(\vy_{t}-\hat{\vy}_{t,i}\right)\right)\\
\vsigma_{t,i}^{-2} & =\left(1-\alpha\right)\vsigma_{t,i-1}^{-2}+\alpha\vsigma_{t|t-1}^{-2}+\alpha\,\diag\left(\vH_{t,i}^{\trans}\vR_{t,i}^{-1}\vH_{t,i}\right)
\end{align}
This update takes $O(\nout^2 \nparam)$ per iteration.

\subsubsection{\Bog Diag}

Substituting \cref{eq:diag-NLL-psi1,eq:diag-NLL-psi2} into \cref{eq:bog}
gives
\begin{align}
\vpsi_{t}^{\left(1\right)} & =\vpsi_{t\vert t-1}^{\left(1\right)}+\alpha\vsigma_{t|t-1}^{2}\vg_{t}\\
\vpsi_{t}^{\left(2\right)} & =\vpsi_{t\vert t-1}^{\left(2\right)}+2\alpha\vsigma_{t|t-1}^{2}\vmu_{t|t-1}\vg_{t}+\alpha\vsigma_{t|t-1}^{4}\diag\left(\vG_{t}\right)
\end{align}
Translating to $\left(\vmu_{t},\vsigma_{t}^{2}\right)$ gives the
\bog-\dg update
\begin{align}
\vmu_{t} & =\vsigma_{t}^{2}\vsigma_{t|t-1}^{-2}\vmu_{t|t-1}+\alpha\vsigma_{t}^{2}\vsigma_{t|t-1}^{2}\vg_{t}
\label{eq:bog-diag-mu}\\
\vsigma_{t}^{-2} & =\vsigma_{t|t-1}^{-2}-4\alpha\vsigma_{t|t-1}^{2}\vmu_{t|t-1}\vg_{t}-2\alpha\vsigma_{t|t-1}^{4}\diag\left(\vG_{t}\right)
\label{eq:bog-diag-prec}
\end{align}
This update takes $O(\nsample \nparam)$ to estimate $\vG_t$ using $\GMCEF_t$, $O(\nhutchinson \nsample \nparam)$ using $\GMCH_t$ and Hutchinson's method, and $O(\nparam)$ using $\GLEF_t$.

\subsubsection{\Bog-\hesslin Diag}

Applying \cref{thm:lbong} to \cref{eq:bog-diag-mu,eq:bog-diag-prec} gives the \bog-\hesslin-\dg update
\begin{align}
\vmu_{t} & =\vsigma_{t}^{2}\vsigma_{t|t-1}^{-2}\vmu_{t|t-1}+\alpha\vsigma_{t}^{2}\vsigma_{t|t-1}^{2}\left(\vH_{t}^{\trans}\vR_{t}^{-1}\left(\vy_{t}-\hat{\vy}_{t}\right)\right)\\
\vsigma_{t}^{-2} & =\vsigma_{t|t-1}^{-2}-4\alpha\vsigma_{t|t-1}^{2}\vmu_{t|t-1}\left(\vH_{t}^{\trans}\vR_{t}^{-1}\left(\vy_{t}-\hat{\vy}_{t}\right)\right)+2\alpha\vsigma_{t|t-1}^{4}\diag\left(\vH_{t}^{\trans}\vR_{t}^{-1}\vH_{t}\right)
\end{align}
This update takes $O(\nout^2 \nparam)$.

\subsubsection{\BBB Diag}

Substituting \cref{eq:diag-NLL-psi1,eq:diag-NLL-psi2,eq:diag-KL-psi1,eq:diag-KL-psi2}
into \cref{eq:bbb} gives
\begin{align}
\vpsi_{t,i}^{\left(1\right)} & =\vpsi_{t,i-1}^{\left(1\right)}+\alpha\vsigma_{t,i-1}^{2}\vg_{t,i}-\alpha\vsigma_{t,i-1}^{2}\vsigma_{t|t-1}^{-2}\left(\vmu_{t,i-1}-\vmu_{t|t-1}\right)\\
\vpsi_{t}^{\left(2\right)} 
    & =\vpsi_{t,i-1}^{\left(2\right)}
    +2\alpha\vsigma_{t,i-1}^{2}\vmu_{t,i-1}\vg_{t,i}+\alpha\vsigma_{t,i-1}^{4}\diag\left(\vG_{t,i}\right)
    \nonumber \\
    &\quad -2\alpha\vsigma_{t,i-1}^{2}\vsigma_{t|t-1}^{-2}\vmu_{t,i-1}\left(\vmu_{t,i-1}-\vmu_{t|t-1}\right)
    -\alpha\vsigma_{t,i-1}^{4}\left(\vsigma_{t|t-1}^{-2}-\vsigma_{t,i-1}^{-2}\right)
\end{align}
Translating to $\left(\vmu_{t,i},\vSigma_{t,i}\right)$ gives the \bbb-\dg update
\begin{align}
\vmu_{t,i} & =\vsigma_{t,i}^{2}\vsigma_{t,i-1}^{-2}\vmu_{t,i-1}+\alpha\vsigma_{t,i}^{2}\vsigma_{t,i-1}^{2}\vg_{t,i}+\alpha\vsigma_{t,i}^{2}\vsigma_{t,i-1}^{2}\vsigma_{t|t-1}^{-2}\left(\vmu_{t|t-1}-\vmu_{t,i-1}\right)
\label{eq:bbb-diag-mu}\\
\vsigma_{t,i}^{-2} 
    & =\vsigma_{t,i-1}^{-2}
    -4\alpha\vsigma_{t,i-1}^{2}\vmu_{t,i-1}\vg_{t,i}
    -2\alpha\vsigma_{t,i-1}^{4}\diag\left(\vG_{t,i}\right) 
    \nonumber \\
    &\quad +4\alpha\vsigma_{t,i-1}^{2}\vsigma_{t|t-1}^{-2}\vmu_{t,i-1}\left(\vmu_{t,i-1}-\vmu_{t|t-1}\right)
    +2\alpha\vsigma_{t,i-1}^{4}\left(\vsigma_{t|t-1}^{-2}-\vsigma_{t,i-1}^{-2}\right)
    \label{eq:bbb-diag-prec}
\end{align}
This update takes $O(\nsample \nparam)$ per iteration to estimate $\vG_t$ using $\GMCEF_t$, $O(\nhutchinson \nsample \nparam)$ per iteration using $\GMCH_t$ and Hutchinson's method, and $O(\nparam)$ per iteration using $\GLEF_t$.

\subsubsection{\BBB-\hesslin Diag}

Applying \cref{thm:lbong} to \cref{eq:bbb-diag-mu,eq:bbb-diag-prec} gives the \bbb-\hesslin-\dg update
\begin{align}
\vmu_{t,i} & =\vsigma_{t,i}^{2}\vsigma_{t,i-1}^{-2}\vmu_{t,i-1}+\alpha\vsigma_{t,i}^{2}\vsigma_{t,i-1}^{2}\left(\vH_{t,i}^{\trans}\vR_{t,i}^{-1}\left(\vy_{t}-\hat{\vy}_{t,i}\right)\right)\nonumber \\
 & \quad+\alpha\vsigma_{t,i}^{2}\vsigma_{t,i-1}^{2}\vsigma_{t|t-1}^{-2}\left(\vmu_{t|t-1}-\vmu_{t,i-1}\right)\\
\vsigma_{t,i}^{-2} & =\vsigma_{t,i-1}^{-2}-4\alpha\vsigma_{t,i-1}^{2}\vmu_{t,i-1}\left(\vH_{t,i}^{\trans}\vR_{t,i}^{-1}\left(\vy_{t}-\hat{\vy}_{t,i}\right)\right)+2\alpha\vsigma_{t,i-1}^{4}\diag\left(\vH_{t,i}^{\trans}\vR_{t,i}^{-1}\vH_{t,i}\right)\nonumber \\
 & \quad+4\alpha\vsigma_{t,i-1}^{2}\vsigma_{t|t-1}^{-2}\vmu_{t,i-1}\left(\vmu_{t,i-1}-\vmu_{t|t-1}\right)+2\alpha\vsigma_{t,i-1}^{4}\vsigma_{t|t-1}^{-2}-2\alpha\vsigma_{t,i-1}^{2}
\end{align}
This update takes $O(\nout^2 \nparam)$ per iteration.

%% file: sections/derivations-diag-mom.tex
\subsection{Diagonal Gaussian, Moment parameters}

Throughout this subsection, vector multiplication and exponents are elementwise.

The Bonnet and Price theorems give
\begin{align}
\nabla_{\vmu_{t,i-1}}\expectQ{\log p\left(\vy_{t}|f_{t}\left(\vtheta_{t}\right)\right)}{\vtheta_{t}\sim q_{\vpsi_{t,i-1}}} & =\vg_{t,i}\label{eq:diag-NLL-mu}\\
\nabla_{\vsigma_{t,i-1}^{2}}\expectQ{\log p\left(\vy_{t}|f_{t}\left(\vtheta_{t}\right)\right)}{\vtheta_{t}\sim q_{\vpsi_{t,i-1}}} & =\tfrac{1}{2}\diag\left(\vG_{t,i}\right)\label{eq:diag-NLL-Sigma}
\end{align}
From \cref{sec:diag-Nat} we have
\begin{align}
\nabla_{\vmu_{t,i-1}}\KLpq{q_{\vpsi_{t,i-1}}}{q_{\vpsi_{t|t-1}}} & =\vsigma_{t|t-1}^{-2}\left(\vmu_{t,i-1}-\vmu_{t|t-1}\right)\label{eq:diag-KL-mu}\\
\nabla_{\vsigma_{t,i-1}^{2}}\KLpq{q_{\vpsi_{t,i-1}}}{q_{\vpsi_{t|t-1}}} & =\tfrac{1}{2}\left(\vsigma_{t|t-1}^{-2}-\vsigma_{t,i-1}^{-2}\right)\label{eq:diag-KL-Sigma}
\end{align}
We write the Fisher with respect to the moment parameters $\vpsi=(\vmu,\vsigma^{2})$
as a block matrix: 
\begin{equation}
\vF_{\vpsi}=\left[\begin{array}{cc}
\vF_{\vmu,\vmu} & \vF_{\vmu,\vsigma^{2}}\\
\vF_{\vsigma^{2},\vmu} & \vF_{\vsigma^{2},\vsigma^{2}}
\end{array}\right]
\end{equation}
The blocks can be calculated by the second-order Fisher formula 
\begin{align}
\vF_{\vmu,\vmu} & =-\expectQ{\nabla_{\vmu,\vmu}\log q_{\vpsi}(\vtheta)}{q_{\vpsi}}\\
 & =\Diag\left(\vsigma^{-2}\right)\\
\vF_{\vmu,\vsigma^{2}} & =-\expectQ{\nabla_{\vmu,\vsigma^{2}}\log q_{\vpsi}(\vtheta)}{q_{\vpsi}}\\
 & =\expectQ{\Diag\left((\vtheta-\vmu)\vsigma^{-4}\right)}{q_{\vpsi}}\\
 & =\bm{0}\\
\vF_{\vsigma^{2},\vsigma^{2}} & =-\expectQ{\nabla_{\vsigma^{2},\vsigma^{2}}\log q_{\vpsi}(\vtheta)}{q_{\vpsi}}\\
 & =-\expectQ{\Diag\left(-\left(\vmu-\vtheta\right)^{2}\vsigma^{-6}+\tfrac{1}{2}\vsigma^{-4}\right)}{q_{\vpsi}}\\
 & =\tfrac{1}{2}\Diag\left(\vsigma^{-4}\right)
\end{align}
Therefore the preconditioner for the NGD methods is 
\begin{align}
\vF_{\vpsi_{t,i-1}}^{-1} & =\left[\begin{array}{cc}
\Diag\left(\sigma_{t,i-1}^{2}\right) & \bm{0}\\
\bm{0} & 2\Diag\left(\sigma_{t,i-1}^{4}\right)
\end{array}\right]\label{eq:diag-mom-Fisher}
\end{align}

\subsubsection{\Bong Diag, Moment}

Substituting \cref{eq:diag-NLL-mu,eq:diag-NLL-Sigma,eq:diag-mom-Fisher}
into \cref{eq:bong} gives the \bong-\diagmom update
\begin{align}
\vmu_{t} & =\vmu_{t\vert t-1}+\vsigma_{t|t-1}^{2}\vg_{t}
\label{eq:bong-diag-mom-mu}\\
\vsigma_{t}^{2} & =\vsigma_{t\vert t-1}^{2}+\vsigma_{t\vert t-1}^{4}\diag\left(\vG_{t}\right)
\label{eq:bong-diag-mom-var}
\end{align}
This update takes $O(\nsample \nparam)$ to estimate $\vG_t$ using $\GMCEF_t$, $O(\nhutchinson \nsample \nparam)$ using $\GMCH_t$ and Hutchinson's method, and $O(\nparam)$ using $\GLEF_t$.

\subsubsection{\Bong-\hesslin Diag, Momemt}

Applying \cref{thm:lbong} to \cref{eq:bong-diag-mom-mu,eq:bong-diag-mom-var} gives the \bong-\hesslin-\diagmom update
\begin{align}
\vmu_{t} & =\vmu_{t\vert t-1}+\vsigma_{t|t-1}^{2}\left(\vH_{t}^{\trans}\vR_{t}^{-1}(\vy_{t}-\hat{\vy}_{t})\right)\\
\vsigma_{t}^{2} & =\vsigma_{t\vert t-1}^{2}-\vsigma_{t\vert t-1}^{4}\diag\left(\vH_{t}^{\trans}\vR_{t}^{-1}\vH_{t}\right)
\end{align}
This update takes $O(\nout^2 \nparam)$.

\subsubsection{\Blr Diag, Moment}

Substituting \cref{eq:diag-NLL-mu,eq:diag-NLL-Sigma,eq:diag-KL-mu,eq:diag-KL-Sigma,eq:diag-mom-Fisher}
into \cref{eq:blr} gives the \blr-\diagmom update
\begin{align}
\vmu_{t,i} & =\vmu_{t,i-1}+\alpha\vsigma_{t,i-1}^{2}\vsigma_{t|t-1}^{-2}\left(\vmu_{t|t-1}-\vmu_{t,i-1}\right)+\alpha\vsigma_{t,i-1}^{2}\vg_{t,i}
\label{eq:blr-diag-mom-mu}\\
\vsigma_{t,i}^{2} & =\vsigma_{t,i-1}^{2}+\alpha\vsigma_{t,i-1}^{4}\left(\vsigma_{t,i-1}^{-2}-\vsigma_{t|t-1}^{-2}\right)+\alpha\vsigma_{t,i-1}^{4}\diag\left(\vG_{t,i}\right)
\label{eq:blr-diag-mom-var}
\end{align}
This update takes $O(\nsample \nparam)$ per iteration to estimate $\vG_t$ using $\GMCEF_t$, $O(\nhutchinson \nsample \nparam)$ per iteration using $\GMCH_t$ and Hutchinson's method, and $O(\nparam)$ per iteration using $\GLEF_t$.

\subsubsection{\Blr-\hesslin Diag, Moment}

Applying \cref{thm:lbong} to \cref{eq:blr-diag-mom-mu,eq:blr-diag-mom-var} gives the \blr-\hesslin-\diagmom update
\begin{align}
\vmu_{t,i} & =\vmu_{t,i-1}+\alpha\vsigma_{t,i-1}^{2}\vsigma_{t|t-1}^{-2}\left(\vmu_{t|t-1}-\vmu_{t,i-1}\right)+\alpha\vsigma_{t,i-1}^{2}\left(\vH_{t,i}^{\trans}\vR_{t,i}^{-1}(\vy_{t}-\hat{\vy}_{t,i})\right)\\
\vsigma_{t,i}^{2} & =\vsigma_{t,i-1}^{2}+\alpha\vsigma_{t,i-1}^{4}\left(\vsigma_{t,i-1}^{-2}-\vsigma_{t|t-1}^{-2}\right)-\alpha\vsigma_{t,i-1}^{4}\diag\left(\vH_{t,i}^{\trans}\vR_{t,i}^{-1}\vH_{t,i}\right)
\end{align}
This update takes $O(\nout^2 \nparam)$ per iteration.

\subsubsection{\Bog Diag, Moment}

Substituting \cref{eq:diag-NLL-mu,eq:diag-NLL-Sigma} into \cref{eq:bog}
gives the \bog-\diagmom update
\begin{align}
\vmu_{t} & =\vmu_{t\vert t-1}+\alpha\vg_{t}
\label{eq:bog-diag-mom-mu}\\
\vsigma_{t}^2 &= \vsigma_{t|t-1}^2 + \frac{\alpha}{2} \diag\left(\vG_{t}\right)
\label{eq:bog-diag-mom-var}
\end{align}
This update takes $O(\nsample \nparam)$ to estimate $\vG_t$ using $\GMCEF_t$, $O(\nhutchinson \nsample \nparam)$ using $\GMCH_t$ and Hutchinson's method, and $O(\nparam)$ using $\GLEF_t$.

\subsubsection{\Bog-\hesslin Diag, Moment}

Applying \cref{thm:lbong} to \cref{eq:bog-diag-mom-mu,eq:bog-diag-mom-var} gives the \bog-\hesslin-\diagmom update
\begin{align}
\vmu_{t} & =\vmu_{t\vert t-1}+\alpha\vH_{t}^{\trans}\vR_{t}^{-1}(\vy_{t}-\hat{\vy}_{t})\\
\vsigma_{t}^{2} & =\vsigma_{t\vert t-1}^{2}-\frac{\alpha}{2}\diag\left(\vH_{t}^{\trans}\vR_{t}^{-1}\vH_{t}\right)
\end{align}
This update takes $O(\nout^2 \nparam)$.

\subsubsection{\BBB Diag, Moment}

Substituting \cref{eq:diag-NLL-mu,eq:diag-NLL-Sigma,eq:diag-KL-mu,eq:diag-KL-Sigma}
into \cref{eq:bbb} gives the \bbb-\diagmom
\begin{align}
\vmu_{t,i} & =\vmu_{t,i-1}+\alpha\vsigma_{t|t-1}^{-2}\left(\vmu_{t|t-1}-\vmu_{t,i-1}\right)+\alpha\vg_{t,i}
\label{eq:bbb-diag-mom-mu}\\
\vsigma_{t}^{2} & =\vsigma_{t,i-1}^{2}+\frac{\alpha}{2}\left(\vsigma_{t,i-1}^{-2}-\vsigma_{t|t-1}^{-2}\right)+\frac{\alpha}{2}\diag\left(\vG_{t,i}\right)
\label{eq:bbb-diag-mom-var}
\end{align}
This update takes $O(\nsample \nparam)$ per iteration to estimate $\vG_t$ using $\GMCEF_t$, $O(\nhutchinson \nsample \nparam)$ per iteration using $\GMCH_t$ and Hutchinson's method, and $O(\nparam)$ per iteration using $\GLEF_t$.

This is similar to the original diagonal Gaussian method in \citep{BBB} except (1) they reparameterize $\vsigma = \log(1+\exp(\vrho))$ (elementwise) and do GD on $(\vmu,\vrho)$ instead of $(\vmu, \vsigma^2)$, and (2) they use the reparameterization trick instead of Price's theorem for calculating the gradient with respect to $\vrho$ (via $\vsigma$).

\subsubsection{\BBB-\hesslin Diag, Moment}

Applying \cref{thm:lbong} to \cref{eq:bbb-diag-mom-mu,eq:bbb-diag-mom-var} gives the \bbb-\hesslin-\diagmom update
\begin{align}
\vmu_{t,i} & =\vmu_{t,i-1}+\alpha\vsigma_{t|t-1}^{-2}\left(\vmu_{t|t-1}-\vmu_{t,i-1}\right)+\alpha\vH_{t,i}^{\trans}\vR_{t,i}^{-1}(\vy_{t}-\hat{\vy}_{t,i})\\
\vsigma_{t}^{2} & =\vsigma_{t,i-1}^{2}+\frac{\alpha}{2}\left(\vsigma_{t,i-1}^{-2}-\vsigma_{t|t-1}^{-2}\right)-\frac{\alpha}{2}\diag\left(\vH_{t,i}^{\trans}\vR_{t,i}^{-1}\vH_{t,i}\right)
\end{align}
This update takes $O(\nout^2 \nparam)$ per iteration.

%% file: sections/derivations-DLR.tex
\subsection{Diagonal plus low rank}
\label{sec:DLR-deriv}

Assume the prior is given by
\begin{equation}
q_{\vpsi_{t|t-1}}\left(\vtheta_{t}\right)=\gauss\left(\vtheta_{t}|\vmu_{t|t-1},\left(\vUpsilon_{t|t-1}+\vW_{t|t-1}\vW_{t|t-1}^{\trans}\right)^{-1}\right)\label{eq:DLR-prior}
\end{equation}
with $\vW\in\real^{\nparam\times\rank}$ and diagonal $\vUpsilon_{t|t-1}\in\real^{\nparam\times \nparam}$.
We sometimes abuse notation by writing $\vUpsilon_{t|t-1}$ for the
vector $\diag\left(\vUpsilon_{t|t-1}\right)$ when the meaning is
clear from context.

Substituting the DLR form in the gradients for the KL divergence derived
in \cref{sec:FC-Mom} gives
\begin{align}
\nabla_{\vmu_{t,i-1}}\KLpq{q_{\vpsi_{t,i-1}}}{q_{\vpsi_{t|t-1}}} & =\left(\vUpsilon_{t|t-1}+\vW_{t|t-1}\vW_{t|t-1}^{\trans}\right)\left(\vmu_{t,i-1}-\vmu_{t|t-1}\right)\label{eq:DLR-KL-mu}\\
\nabla_{\vSigma_{t,i-1}}\KLpq{q_{\vpsi_{t,i-1}}}{q_{\vpsi_{t|t-1}}} & =\tfrac{1}{2}\left(\vUpsilon_{t|t-1}-\vUpsilon_{t,i-1}+\vW_{t|t-1}\vW_{t|t-1}^{\trans}-\vW_{t,i-1}\vW_{t,i-1}^{\trans}\right)\label{eq:DLR-KL-Sigma}
\end{align}

For any function $\ell$ the chain rule gives
\begin{align}
\nabla_{\vUpsilon_{t|t-1}}\ell & =-\diag\left(\left(\vUpsilon_{t|t-1}+\vW_{t|t-1}\vW_{t|t-1}^{\trans}\right)^{-1}\left(\nabla_{\vSigma_{t|t-1}}\ell\right)\left(\vUpsilon_{t|t-1}+\vW_{t|t-1}\vW_{t|t-1}^{\trans}\right)^{-1}\right)\\
\nabla_{\vW_{t|t-1}}\ell & =-2\left(\vUpsilon_{t|t-1}+\vW_{t|t-1}\vW_{t|t-1}^{\trans}\right)^{-1}\left(\nabla_{\vSigma_{t|t-1}}\ell\right)\left(\vUpsilon_{t|t-1}+\vW_{t|t-1}\vW_{t|t-1}^{\trans}\right)^{-1}\vW_{t|t-1}
\end{align}
Therefore the gradients we need are
\begin{align}
\nabla_{\vmu_{t,i-1}}
\expectQ
    {\log p\left(\vy_{t}\vert h_{t}\left(\vtheta_{t}\right)\right)}
    {\vtheta_{t}\sim q_{\vpsi_{t,i-1}}} 
    &= \vg_{\vpsi_{t,i}}
    \label{eq:DLR-NLL-mu}\\
\nabla_{\vUpsilon_{t,i-1}}
\expectQ
    {\log p\left(\vy_{t}\vert h_{t}\left(\vtheta_{t}\right)\right)}
    {\vtheta_{t}\sim q_{\vpsi_{t,i-1}}}
    &= -\tfrac{1}{2} \diag
    \left( \begin{array}{l}
         \left(\vUpsilon_{t,i-1}+\vW_{t,i-1}\vW_{t,i-1}^{\trans}\right)^{-1}
         \vG_{t,i} \\
         \times \left(\vUpsilon_{t,i-1}+\vW_{t,i-1}\vW_{t,i-1}^{\trans}\right)^{-1}
    \end{array}
    \right)
    \label{eq:DLR-NLL-Ups}\\
\nabla_{\vW_{t,i-1}}
\expectQ
    {\log p\left(\vy_{t}\vert h_{t}\left(\vtheta_{t}\right)\right)}
    {\vtheta_{t}\sim q_{\vpsi_{t,i-1}}} 
    &=
    -\left(\vUpsilon_{t,i-1}+\vW_{t,i-1}\vW_{t,i-1}^{\trans}\right)^{-1}
    \vG_{t,i}
    \nonumber \\
    & \quad\enskip
    \times \left(\vUpsilon_{t,i-1}+\vW_{t,i-1}\vW_{t,i-1}^{\trans}\right)^{-1}\vW_{t,i-1}
    \label{eq:DLR-NLL-W}
\end{align}
and
\begin{align}
&\nabla_{\vUpsilon_{t,i-1}} \KLpq{q_{\vpsi_{t,i-1}}}
{q_{\vpsi_{t|t-1}}} =
\nonumber \\
&\qquad\qquad
-\tfrac{1}{2} \diag
\left( \begin{array}{l}
    \left(\vUpsilon_{t,i-1}+\vW_{t,i-1}\vW_{t,i-1}^{\trans}\right)^{-1} \\
    \times \left(\vUpsilon_{t|t-1}-\vUpsilon_{t,i-1}+\vW_{t|t-1}\vW_{t|t-1}^{\trans}-\vW_{t,i-1}\vW_{t,i-1}^{\trans}\right) \\
    \times \left(\vUpsilon_{t,i-1}+\vW_{t,i-1}\vW_{t,i-1}^{\trans}\right)^{-1}
\end{array} \right)
\label{eq:DLR-KL-Ups}
\end{align}
\begin{align}
\nabla_{\vW_{t,i-1}}\KLpq{q_{\vpsi_{t,i-1}}}{q_{\vpsi_{t|t-1}}} 
&=-\left(\vUpsilon_{t,i-1}+\vW_{t,i-1}\vW_{t,i-1}^{\trans}\right)^{-1}
\nonumber \\
&\quad \times \left(\vUpsilon_{t|t-1}-\vUpsilon_{t,i-1}+\vW_{t|t-1}\vW_{t|t-1}^{\trans}-\vW_{t,i-1}\vW_{t,i-1}^{\trans}\right)
\nonumber \\
&\quad \times \left(\vUpsilon_{t,i-1}+\vW_{t,i-1}\vW_{t,i-1}^{\trans}\right)^{-1}
\vW_{t,i-1}
\label{eq:DLR-KL-W}
\end{align}

The Fisher matrix can be decomposed as a block-diagonal with blocks for $\vmu_{t,i}$ and for $(\vUpsilon_{t,i},\vW_{t,i})$, but (in contrast to the FC Gaussian case in \cref{eq:FC-mom-Fisher})
we have not found an efficient way to analytically invert the latter block, which has size $\nparam + \rank\nparam$.
To avoid the $O\left(\rank^{3} \nparam^{3}\right)$ cost of brute force inversion,
we use a different strategy for \bong and \blr of performing the
update on the natural parameters of the FC Gaussian as in \cref{sec:FC-Nat}
and projecting the result back to rank $\rank$ using SVD. Specifically,
assume the updated precision from applying \cref{eq:bong-FC-prec}
for \bong or \cref{eq:BLR-FC-Nat-prec} for \blr can be written
as
\begin{equation}
\tilde{\vSigma}_{t,i}^{-1}=\tilde{\vUpsilon}_{t,i}+\tilde{\vW}_{t,i}\tilde{\vW}_{t,i}^{\trans}
\end{equation}
 and let the SVD of $\tilde{\vW}_{t,i}$ be
\begin{equation}
\tilde{\vW}_{t,i}=\vU_{t,i}\vLambda_{t,i}\vV_{t,i}^{\trans}
\end{equation}
with $\vU_{t,i},\vV_{t,i}$ orthogonal and $\vLambda_{t,i}$ rectangular-diagonal.
Following \citep{SLANG,lofi} we define the update
\begin{align}
\vW_{t,i} & =\vU_{t,i}\left[:,{:\rank}\right]\vLambda_{t,i}\left[{:\rank},{:\rank}\right]\\
\vUpsilon_{t,i} & =\tilde{\vUpsilon}_{t,i}+\diag\left(\tilde{\vW}_{t,i}\tilde{\vW}_{t,i}^{\trans}-\vW_{t,i}\vW_{t,i}^{\trans}\right)
\end{align}
so that $\vW_{t,i}$ contains the top $\rank$ singular vectors and values
of the FC posterior and the diagonal is preserved: $\diag\left(\vUpsilon_{t,i}+\vW_{t,i}\vW_{t,i}^{\trans}\right)=\diag\left(\tilde{\vSigma}_{t,i}^{-1}\right)$.
This approach works for \efmc and \eflin methods but not \mchess, which we omit.

Finally, in the \mcef methods we sample from the DLR prior using the routine in \citep{SLANG,LRVGA} which takes $O(\rank(\rank+\nsample)\nparam)$.

\subsubsection{\Bong DLR}

Substituting the DLR prior from \cref{eq:DLR-prior} into the FC precision
update from \cref{eq:bong-FC-prec} and using the \mcef approximation
yields the posterior precision
\begin{align}
\tilde{\vSigma}_{t}^{-1} & =\vUpsilon_{t|t-1}+\vW_{t|t-1}\vW_{t|t-1}^{\trans}-\GMCEF_t\\
 & =\tilde{\vUpsilon}_{t}+\tilde{\vW}_{t}\tilde{\vW}_{t}^{\trans}\\
\tilde{\vUpsilon}_{t} & =\vUpsilon_{t|t-1}\\
\tilde{\vW}_{t} & =\left[\vW_{t|t-1},\frac{1}{\sqrt{\nsample}}\gradmat_{t}\right]
\label{eq:bong-dlr-Wtilde}
\end{align}
Note $\tilde{\vW}_{t}\in\real^{\nparam\times\left(\rank+\nsample\right)}$. Using this
posterior precision in the mean update from \cref{eq:bong-FC-mean}
yields
\begin{align}
\vmu_{t} & =\vmu_{t|t-1}+\left(\vUpsilon_{t|t-1}+\tilde{\vW}_{t}\tilde{\vW}_{t}^{\trans}\right)^{-1}\vg_{t}\\
 & =\vmu_{t|t-1}+\left(\vUpsilon_{t|t-1}^{-1}-\vUpsilon_{t|t-1}^{-1}\tilde{\vW}_{t}\left(\vI_{\rank+\nsample}+\tilde{\vW}_{t}^{\trans}\vUpsilon_{t|t-1}^{-1}\tilde{\vW}_{t}\right)^{-1}\tilde{\vW}_{t}^{\trans}\vUpsilon_{t|t-1}^{-1}\right)\vg_{t}
 \label{eq:bong-dlr-mu}
\end{align}
where the second line comes from the Woodbury identity and can be
computed in $O((\rank+\nsample)^{2}\nparam+(\rank+\nsample)^{3})$.
Applying the SVD projection gives
\begin{align}
\vW_{t} & =\vU_{t}\left[:,{:\rank}\right]\vLambda_{t}\left[{:\rank},{:\rank}\right]
\label{eq:bong-dlr-W}\\
\vUpsilon_{t} & =\vUpsilon_{t|t-1}+\diag\left(\tilde{\vW}_{t}\tilde{\vW}_{t}^{\trans}-\vW_{t}\vW_{t}^{\trans}\right)
\label{eq:bong-dlr-Ups}\\
\left(\vU_{t},\vLambda_{t},\_\right) & ={\rm SVD}\left(\tilde{\vW}_{t}\right)
\label{eq:bong-dlr-svd}
\end{align}
which takes $O((\rank+\nsample)^{2}\nparam)$ for the SVD. 
Therefore the \bong-\efmc-\dlr update is defined by 
\cref{eq:bong-dlr-W,eq:bong-dlr-mu,eq:bong-dlr-Ups,eq:bong-dlr-Wtilde,eq:bong-dlr-svd}
and takes $O((\rank+\nsample)^{2}\nparam+(\rank+\nsample)^{3})$.

The \bong-\eflin-\dlr update comes from replacing $\frac{1}{\sqrt{M}}\gradmat_t$ with $\gL_t$ in \cref{eq:bong-dlr-Wtilde} 
and replacing $\vI_{\rank+\nsample}$ with $\vI_{\rank+1}$ in \cref{eq:bong-dlr-mu}. 
This update takes $O((\rank+1)^{2}\nparam + (\rank+1)^{3})$.

\subsubsection{\Bong-\hesslin DLR (\LOFI)}

Applying \cref{thm:lbong} to \cref{eq:bong-dlr-mu,eq:bong-dlr-Wtilde} gives the \bong-\hesslin-\dlr update:
\begin{align}
\vmu_{t} 
    &= \vmu_{t|t-1}
    + \left(\vUpsilon_{t|t-1}^{-1}-\vUpsilon_{t|t-1}^{-1}\tilde{\vW}_{t}\left(\vI_{\rank+\nout}+\tilde{\vW}_{t}^{\trans}\vUpsilon_{t|t-1}^{-1}\tilde{\vW}_{t}\right)^{-1}\tilde{\vW}_{t}^{\trans}\vUpsilon_{t|t-1}^{-1}\right) \nonumber \\
    & \qquad \times \vH_{t}^{\trans}\vR_{t}^{-1}\left(\vy_{t}-\hat{\vy}_{t}\right)\\
\vW_{t} & =\vU_{t}\left[:,{:\rank}\right]\vLambda_{t}\left[{:\rank},{:\rank}\right]\\
\vUpsilon_{t} & =\vUpsilon_{t|t-1}+\diag\left(\tilde{\vW}_{t}\tilde{\vW}_{t}^{\trans}-\vW_{t}\vW_{t}^{\trans}\right)\\
\left(\vU_{t},\vLambda_{t},\_\right) & ={\rm SVD}\left(\tilde{\vW}_{t}\right)\\
\tilde{\vW}_{t} & =\left[\vW_{t|t-1},\vH_{t}^{\trans}\vA_{t}^{\trans}\right]\\
\vA_{t} & ={\rm chol}\left(\vR_{t}^{-1}\right)
\end{align}
This is equivalent to \LOFI \citep{lofi} and takes $O((\rank+\nout)^{2}\nparam+(\rank+\nout)^{3})$.

\subsubsection{\Blr DLR (\SLANG)}
\label{sec:BLR-DLR}

Substituting DLR forms for $q_{\vpsi_{t|t-1}}$ and $q_{\vpsi_{t,i-1}}$
into the FC precision update from \cref{eq:BLR-FC-Nat-prec} and
using the \mcef approximation yields the posterior precision
\begin{align}
\tilde{\vSigma}_{t,i}^{-1} & =\left(1-\alpha\right)\left(\vUpsilon_{t,i-1}+\vW_{t,i-1}\vW_{t,i-1}^{\trans}\right)+\alpha\left(\vUpsilon_{t|t-1}+\vW_{t|t-1}\vW_{t|t-1}^{\trans}\right)-\alpha\GMCEF_{t,i}\\
 & =\tilde{\vUpsilon}_{t,i}+\tilde{\vW}_{t,i}\tilde{\vW}_{t,i}^{\trans}\\
\tilde{\vUpsilon}_{t,i} & =\left(1-\alpha\right)\vUpsilon_{t,i-1}+\alpha\vUpsilon_{t|t-1}
\label{eq:blr-dlr-UpsTilde}\\
\tilde{\vW}_{t,i} & =\left[\sqrt{1-\alpha}\vW_{t,i-1},\sqrt{\alpha}\vW_{t|t-1},\sqrt{\frac{\alpha}{\nsample}}\gradmat_{t,i}\right]
\label{eq:blr-dlr-Wtilde}
\end{align}
Note $\tilde{\vW}_{t}\in\real^{\nparam\times(2\rank+\nsample)}$. Using
this posterior precision in the mean update from \cref{eq:blr-FC-nat-mu}
yields
\begin{align}
\vmu_{t,i} 
&= \vmu_{t,i-1}
    + \alpha \left(\tilde{\vUpsilon}_{t,i}+\tilde{\vW}_{t,i}\tilde{\vW}_{t,i}^{\trans}\right)^{-1}
    \nonumber \\
    &\qquad \times 
    \left(\left(\vUpsilon_{t|t-1}+\vW_{t|t-1}\vW_{t|t-1}^{\trans}\right)\left(\vmu_{t|t-1}-\vmu_{t,i-1}\right)+\vg_{t,i}\right)\\
 &= \vmu_{t,i-1}
    + \alpha \left(\tilde{\vUpsilon}_{t,i}^{-1}-\tilde{\vUpsilon}_{t,i}^{-1}\tilde{\vW}_{t,i}\left(\vI_{2\rank+\nsample}+\tilde{\vW}_{t,i}^{\trans}\tilde{\vUpsilon}_{t,i}^{-1}\tilde{\vW}_{t,i}\right)^{-1}\tilde{\vW}_{t,i}^{\trans}\tilde{\vUpsilon}_{t,i}^{-1}\right)
    \nonumber \\
    &\qquad \times 
    \left(\left(\vUpsilon_{t|t-1}+\vW_{t|t-1}\vW_{t|t-1}^{\trans}\right)\left(\vmu_{t|t-1}-\vmu_{t,i-1}\right)+\vg_{t,i}\right)
    \label{eq:blr-dlr-mu}
\end{align}
where the second line comes from the Woodbury identity and can be
computed in $O((2\rank+\nsample)^{2}\nparam+(2\rank+\nsample)^{3})$.
Applying the SVD projection gives
\begin{align}
\vW_{t,i} & =\vU_{t,i}\left[:,{:\rank}\right]\vLambda_{t,i}\left[{:\rank},{:\rank}\right]
\label{eq:blr-dlr-W}\\
\vUpsilon_{t,i} & =\tilde{\vUpsilon}_{t,i}+\diag\left(\tilde{\vW}_{t,i}\tilde{\vW}_{t,i}^{\trans}-\vW_{t,i}\vW_{t,i}^{\trans}\right)
\label{eq:blr-dlr-Ups}\\
\left(\vU_{t,i},\vLambda_{t,i},\_\right) & ={\rm SVD}\left(\tilde{\vW}_{t,i}\right)
\label{eq:blr-dlr-svd}
\end{align}
which takes $O((2\rank+\nsample)^{2}\nparam)$ for the SVD. 
Therefore the \blr-\efmc-\dlr update is defined by
\cref{eq:blr-dlr-W,eq:blr-dlr-mu,eq:blr-dlr-svd,eq:blr-dlr-Ups,eq:blr-dlr-Wtilde,eq:blr-dlr-UpsTilde}
and takes $O((2\rank+\nsample)^{2}\nparam+(2\rank+\nsample)^{3})$ per iteration. 
Notice $\tilde{\vW}_t$ has larger rank for \blr than for \bong ($2\rank+\nsample$ vs.\ $\rank+\nsample$) because of the extra $\sqrt{\alpha} \vW_{t|t-1}$ term that originates in the KL part of the VI loss. This difference will slow \blr-\efmc-\dlr relative to \bong-\efmc-\dlr especially when $\rank$ is not small relative to $\nsample$.

This method closely resembles \SLANG \cite{SLANG} in the batch setting where we replace $q_{\vpsi_{t|t-1}}$ with a spherical prior $\gauss (\bm{0},\lambda^{-1}\vI_{\nparam})$ (see \cref{sec:blr-batch}).

We can also define a \blr-\linef-\dlr method by replacing $\sqrt{\frac{\alpha}{M}}\gradmat_t$ with $\sqrt{\alpha}\gL_t$ in \cref{eq:blr-dlr-Wtilde}
and $\vI_{2\rank+\nsample}$ with $\vI_{2\rank+1}$ in \cref{eq:blr-dlr-mu}.
This update takes $O((2\rank+1)^{2}\nparam + (2\rank+1)^{3})$ per iteration.

\subsubsection{\Blr-\hesslin DLR}
\label{sec:LBLR-DLR}

Applying \cref{thm:lbong} to \cref{eq:blr-dlr-mu,eq:blr-dlr-Wtilde} gives the \blr-\hesslin-\dlr update:
\begin{align}
\vmu_{t,i} & =\vmu_{t,i-1}+\alpha\left(\tilde{\vUpsilon}_{t,i}^{-1}-\tilde{\vUpsilon}_{t,i}^{-1}\tilde{\vW}_{t,i}\left(\vI_{2\rank+\nout}+\tilde{\vW}_{t,i}^{\trans}\tilde{\vUpsilon}_{t,i}^{-1}\tilde{\vW}_{t,i}\right)^{-1}\tilde{\vW}_{t,i}^{\trans}\tilde{\vUpsilon}_{t,i}^{-1}\right)\nonumber \\
 & \qquad\times \left(\left(\vUpsilon_{t|t-1}+\vW_{t|t-1}\vW_{t|t-1}^{\trans}\right)\left(\vmu_{t|t-1}-\vmu_{t,i-1}\right)+\vH_{t,i}^{\trans}\vR_{t,i}^{-1}\left(\vy_{t}-\hat{\vy}_{t,i}\right)\right)\\
\vW_{t,i} & =\vU_{t,i}\left[:,{:\rank}\right]\vLambda_{t,i}\left[{:\rank},{:\rank}\right]\\
\vUpsilon_{t,i} & =\tilde{\vUpsilon}_{t,i}+\diag\left(\tilde{\vW}_{t,i}\tilde{\vW}_{t,i}^{\trans}-\vW_{t,i}\vW_{t,i}^{\trans}\right)\\
\left(\vU_{t,i},\vLambda_{t,i},\_\right) & ={\rm SVD}\left(\tilde{\vW}_{t,i}\right)\\
\tilde{\vW}_{t,i} & =\left[\sqrt{1-\alpha}\vW_{t,i-1},\sqrt{\alpha}\vW_{t|t-1},\sqrt{\alpha}\vH_{t,i}^{\trans}\vA_{t,i}^{\trans}\right]\\
\vA_{t,i} & ={\rm chol}\left(\vR_{t,i}^{-1}\right) \\
\tilde{\vUpsilon}_{t,i} & =\left(1-\alpha\right)\vUpsilon_{t,i-1}+\alpha\vUpsilon_{t|t-1}
\end{align}
This update takes $O((2\rank+\nout)^{2}\nparam+(2\rank+\nout)^{3})$ per iteration.
As with the \ef versions of \blr-\dlr, $\tilde{\vW}_t$ has larger rank for \blr-\hesslin-\dlr than for \bong-\hesslin-\dlr ($2\rank+\nout$ vs.\ $\rank+\nout$). This difference will slow \blr-\hesslin-\dlr relative to \bong-\hesslin-\dlr especially when $\rank$ is not small relative to $\nout$.

\subsubsection{\Bog DLR}

Substituting \cref{eq:DLR-NLL-mu,eq:DLR-NLL-Ups,eq:DLR-NLL-W} into
\cref{eq:bog} gives the \bog-\dlr update
\begin{align}
\vmu_{t} & =\vmu_{t\vert t-1}+\alpha\vg_{t}
\label{eq:bog-dlr-mu}\\
\vUpsilon_{t} & =\vUpsilon_{t\vert t-1}-\frac{\alpha}{2}\diag\left(\left(\vUpsilon_{t|t-1}+\vW_{t|t-1}\vW_{t|t-1}^{\trans}\right)^{-1}\vG_{t}\left(\vUpsilon_{t|t-1}+\vW_{t|t-1}\vW_{t|t-1}^{\trans}\right)^{-1}\right)\\
\vW_{t} & =\vW_{t|t-1}-\alpha\left(\vUpsilon_{t|t-1}+\vW_{t|t-1}\vW_{t|t-1}^{\trans}\right)^{-1}\vG_{t}\left(\vUpsilon_{t|t-1}+\vW_{t|t-1}\vW_{t|t-1}^{\trans}\right)^{-1}\vW_{t|t-1}
\end{align}
Using the EF approximation and Woodbury, the \bog-\efmc-\dlr update can be rewritten as
\begin{align}
\vUpsilon_{t} & =\vUpsilon_{t\vert t-1}+\frac{\alpha}{2\nsample}\diag\left(\vB_{t}\vB_{t}^{\trans}\right)
\label{eq:bog-dlr-Ups}\\
\vW_{t} & =\vW_{t|t-1}+\frac{\alpha}{\nsample}\vB_{t}\vB_{t}^{\trans}\vW_{t|t-1}
\label{eq:bog-dlr-W}\\
\vB_{t} & =\left(\vUpsilon_{t|t-1}^{-1}-\vUpsilon_{t|t-1}^{-1}\vW_{t|t-1}\left(\vI_{\rank}+\vW_{t|t-1}^{\trans}\vUpsilon_{t|t-1}^{-1}\vW_{t|t-1}\right)^{-1}\vW_{t|t-1}^{\trans}\vUpsilon_{t|t-1}^{-1}\right)\gradmat_{t}
\label{eq:bog-dlr-B}
\end{align}
which takes $O(\rank\nsample\nparam+\nsample\rank^{2}+\rank^{3})$.

The \bog-\linef-\dlr update comes from replacing $\gradmat_t$ with $\gL_t$ 
in \cref{eq:bog-dlr-B} and dropping the $M^{-1}$ factors in \cref{eq:bog-dlr-Ups,eq:bog-dlr-W}.
This update takes $O(\rank\nparam + \rank^{3})$.

\subsubsection{\Bog-\hesslin DLR}

Applying \cref{thm:lbong} to \cref{eq:bog-dlr-B,eq:bog-dlr-W,eq:bog-dlr-mu,eq:bog-dlr-Ups} gives the \bog-\hesslin-\dlr update
\begin{align}
\vmu_{t} & =\vmu_{t\vert t-1}+\alpha\vH_{t}^{\trans}\vR_{t}^{-1}(\vy_{t}-\hat{\vy}_{t})\\
\vUpsilon_{t} & =\vUpsilon_{t\vert t-1}+\frac{\alpha}{2}\diag\left(\vB_{t}\vB_{t}^{\trans}\right)\\
\vW_{t} & =\vW_{t|t-1}+\alpha\vB_{t}\vB_{t}^{\trans}\vW_{t|t-1}\\
\vB_{t} & =\left(\vUpsilon_{t|t-1}^{-1}-\vUpsilon_{t|t-1}^{-1}\vW_{t|t-1}\left(\vI_{\rank}+\vW_{t|t-1}^{\trans}\vUpsilon_{t|t-1}^{-1}\vW_{t|t-1}\right)^{-1}\vW_{t|t-1}^{\trans}\vUpsilon_{t|t-1}^{-1}\right)\vH_{t}^{\trans}\vA_{t}^{\trans}
\end{align}
This update takes $O(\nout(\nout+\rank)\nparam+\nout\rank^{2}+\rank^{3})$.

\subsubsection{\BBB DLR}
\label{sec:bbb-dlr}

Substituting \cref{eq:DLR-NLL-mu,eq:DLR-NLL-Ups,eq:DLR-NLL-W,eq:DLR-KL-mu,eq:DLR-KL-Ups,eq:DLR-KL-W}
into \cref{eq:bbb} gives the \bbb-\dlr update
\begin{align}
\vmu_{t,i} & =\vmu_{t,i-1}+\alpha\left(\vUpsilon_{t|t-1}+\vW_{t|t-1}\vW_{t|t-1}^{\trans}\right)\left(\vmu_{t|t-1}-\vmu_{t,i-1}\right)+\alpha\vg_{t}
\label{eq:bbb-dlr-mu}\\
\vUpsilon_{t,i} 
    &= \vUpsilon_{t,i-1} \nonumber \\
    &\quad +\frac{\alpha}{2}\diag\left(\vSigma_{t,i-1}\left(\vUpsilon_{t|t-1}-\vUpsilon_{t,i-1}+\vW_{t|t-1}\vW_{t|t-1}^{\trans}-\vW_{t,i-1}\vW_{t,i-1}^{\trans}-\vG_{t,i}\right)\vSigma_{t,i-1}\right)
    \label{eq:bbb-dlr-Ups}\\
\vW_{t,i} 
    & =\vW_{t,i-1}\nonumber \\
    &\quad +\alpha\vSigma_{t,i-1}\left(\vUpsilon_{t|t-1}-\vUpsilon_{t,i-1}+\vW_{t|t-1}\vW_{t|t-1}^{\trans}-\vW_{t,i-1}\vW_{t,i-1}^{\trans}-\vG_{t,i}\right)\vSigma_{t,i-1}\vW_{t,i-1}\label{eq:bbb-dlr-W}
\end{align}
The previous covariance can be written using Woodbury as
\begin{equation}
\vSigma_{t,i-1}=\vUpsilon_{t,i-1}^{-1}-\vUpsilon_{t,i-1}^{-1}\vW_{t,i-1}\left(\vI_{\rank}+\vW_{t,i-1}^{\trans}\vUpsilon_{t,i-1}^{-1}\vW_{t,i-1}\right)^{-1}\vW_{t,i-1}^{\trans}\vUpsilon_{t,i-1}^{-1}\label{eq:bbb-dlr-Woodbury}
\end{equation}

The \bbb-\efmc-\dlr update can be computed efficiently
by expanding terms in \cref{eq:bbb-dlr-Ups,eq:bbb-dlr-W,eq:bbb-dlr-Woodbury}.
For example the terms involving $\vG_{t,i}$ can be calculated as
\begin{align}
\diag \left( -\vSigma_{t,i-1} \GMCEF_{t,i} \vSigma_{t,i-1} \right) 
    &= \frac{1}{\nsample} \diag \left( \begin{array}{l}
        \vUpsilon_{t,i-1}^{-1} \gradmat_{t,i} \gradmatT_{t,i} \vUpsilon_{t,i-1}^{-1} \\
        - \vUpsilon_{t,i-1}^{-1} \gradmat_{t,i} \vB_{t,i}^{\trans} \\
        - \vB_{t,i} \gradmatT_{t,i} \vUpsilon_{t,i-1}^{-1}
        + \vB_{t,i} \vB_{t,i}^{\trans}
    \end{array} \right)
    \label{eq:bbb-dlr-diagSigGSig}\\
-\vSigma_{t,i-1} \GMCEF_{t,i} \vSigma_{t,i-1} \vW_{t,i-1} 
    &= \frac{1}{\nsample} \vUpsilon_{t,i-1}^{-1} \gradmat_{t,i} \gradmatT_{t,i} \vUpsilon_{t,i-1}^{-1} \vW_{t,i-1}
    \nonumber \\
    &\quad -\frac{1}{\nsample} \vUpsilon_{t,i-1}^{-1} \gradmat_{t,i} \vB_{t,i}^{\trans} \vW_{t,i-1}
    \nonumber \\
    &\quad 
    -\frac{1}{\nsample} \vB_{t,i} \gradmatT_{t,i} \vUpsilon_{t,i-1}^{-1} \vW_{t,i-1}
    +\frac{1}{\nsample} \vB_{t,i} \vB_{t,i}^{\trans} \vW_{t,i-1} 
    \label{eq:bbb-dlr-SigGSigW}\\
\vB_{t,i} 
    &= \vUpsilon_{t,i-1}^{-1} \vW_{t,i-1}
    \left(\vI_{\rank}+\vW_{t,i-1}^{\trans}\vUpsilon_{t,i-1}^{-1}\vW_{t,i-1}\right)^{-1}
    \nonumber \\
    &\quad \times
    \vW_{t,i-1}^{\trans} \vUpsilon_{t,i-1}^{-1} \gradmat_{t,i}
    \label{eq:bbb-dlr-B}
\end{align}
Using this strategy the update takes $O((\rank+\nsample)\rank\nparam+\nsample\rank^{2}+\rank^{3})$.

The \bbb-\linef-\dlr update comes from replacing $\gradmat_t$ with $\gL_t$ 
and dropping the $M^{-1}$ factors
in \cref{eq:bbb-dlr-diagSigGSig,eq:bbb-dlr-SigGSigW,eq:bbb-dlr-B}.
This update takes $O(\rank^2 \nparam + \rank^{3})$.

\subsubsection{\BBB-\hesslin DLR}

Applying \cref{thm:lbong} to \cref{eq:bbb-dlr-W,eq:bbb-dlr-mu,eq:bbb-dlr-Ups} gives the \bbb-\hesslin-\dlr update
\begin{align}
\vmu_{t,i} & =\vmu_{t,i-1}+\alpha\left(\vUpsilon_{t|t-1}+\vW_{t|t-1}\vW_{t|t-1}^{\trans}\right)\left(\vmu_{t|t-1}-\vmu_{t,i-1}\right)+\alpha\vH_{t,i}^{\trans}\vR_{t,i}^{-1}(\vy_{t}-\hat{\vy}_{t,i})\\
\vUpsilon_{t,i} & =\vUpsilon_{t,i-1}+\frac{\alpha}{2}\diag\left(\vSigma_{t,i-1}\left(\begin{array}{l}\vUpsilon_{t|t-1}-\vUpsilon_{t,i-1}+\vW_{t|t-1}\vW_{t|t-1}^{\trans}\\-\vW_{t,i-1}\vW_{t,i-1}^{\trans}+\vH_{t,i}^{\trans}\vR_{t,i}^{-1}\vH_{t,i}\end{array}\right)\vSigma_{t,i-1}\right)\\
\vW_{t,i} & =\vW_{t,i-1}+\alpha\vSigma_{t,i-1}\left(\begin{array}{l}\vUpsilon_{t|t-1}-\vUpsilon_{t,i-1}+\vW_{t|t-1}\vW_{t|t-1}^{\trans}\\-\vW_{t,i-1}\vW_{t,i-1}^{\trans}+\vH_{t,i}^{\trans}\vR_{t,i}^{-1}\vH_{t,i}\end{array}\right)\vSigma_{t,i-1}\vW_{t,i-1}
\end{align}
This can be computed in $O((\nout+\rank)^{2}\nparam+\nout\rank^{2}+\rank^{3})$ using
\cref{eq:bbb-dlr-Woodbury} and following a computational approach similar to \cref{eq:bbb-dlr-diagSigGSig,eq:bbb-dlr-SigGSigW,eq:bbb-dlr-B}.

%% file: sections/BLR-batch.tex
\subsection{Batch \blr}
\label{sec:blr-batch}

It is interesting to translate the \blr updates derived here back to 
the batch setting where \blr was developed \citep{BLR},
by replacing
$\gauss\left(\vmu_{t|t-1},\vSigma_{t|t-1}\right)$
with a centered spherical prior $\gauss\left(\bm{0},\lambda^{-1}\vI_{\nparam}\right)$.

The batch \blr-\fc update becomes
\begin{align}
\vmu_{i} 
    &= \vmu_{i-1} 
    + \alpha \vSigma_{i} \left( \vg_{i} - \lambda \vmu_{i-1} \right) \\
\vSigma_{i}^{-1} 
    &= (1-\alpha) \vSigma_{i-1}^{-1}
    + \alpha \left( \lambda \vI_\nparam - \vG_{i} \right)
\end{align}

The batch \blr-\hesslin-\fc update becomes
\begin{align}
\vmu_{i} 
    &= \vmu_{i-1}
    + \alpha \vSigma_{i} \left(
        \vH_{i}^{\trans} \vR_{i}^{-1} (\vy_{t}-\hat{\vy}_{i})
        - \lambda \vmu_{i-1} 
        \right) \\
\vSigma_{i}^{-1} 
    &= (1-\alpha) \vSigma_{i-1}^{-1}
    + \alpha \left( 
        \lambda \vI_\nparam
        + \vH_{i}^{\trans} \vR_{i}^{-1} \vH_{i}
    \right)
\end{align}

The batch \blr-\fcmom update becomes
\begin{align}
\vmu_{i} 
    &= \vmu_{i-1} 
    + \alpha \vSigma_{i-1} \left( \vg_{i} - \lambda \vmu_{i-1} \right) \\
\vSigma_{i} 
    &= (1+\alpha) \vSigma_{i-1}
    + \alpha \vSigma_{i-1} \left( \vG_{i} - \lambda\vI_{\nparam} \right) \vSigma_{i-1}
\end{align}

The batch \blr-\hesslin-\fcmom update becomes
\begin{align}
\vmu_{i} 
    &= \vmu_{i-1} 
    + \alpha \vSigma_{i-1} \left(
        \vH_{i}^{\trans} \vR_{i}^{-1} (\vy_{t}-\hat{\vy}_{i})
        - \lambda \vmu_{i-1}
    \right) \\
\vSigma_{i} 
    &= (1 + \alpha) \vSigma_{i-1}
    - \alpha \vSigma_{i-1} \left( \lambda \vI_{\nparam} + \vH_{i}^{\trans} \vR_{i}^{-1} \vH_{i} \right) \vSigma_{i-1}
\end{align}

The batch \blr-\dg update becomes
\begin{align}
\vmu_{i} 
    &= \vmu_{i-1}
    + \alpha \vsigma_{i}^{2} \left( \vg_{i} - \lambda \vmu_{i-1} \right) \\
\vsigma_{i}^{-2} 
    &= (1-\alpha) \vsigma_{i-1}^{-2}
    + \alpha\,\diag \left( \lambda \vI_\nparam - \vG_{i} \right)
\end{align}
The \mchess version of this update is equivalent to \VON \citep{VON} if we use MC approximation with $\nsample=1$. 
The \mcef version with $M=1$ is equivalent to \VOGN \citep{VON}.

The batch \blr-\hesslin-\dg update becomes
\begin{align}
\vmu_{i} 
    &= \vmu_{i-1}
    + \alpha \vsigma_{i}^{2} 
    \left( 
        \vH_{i}^{\trans} \vR_{i}^{-1} (\vy_{t}-\hat{\vy}_{i}) 
        - \lambda \vmu_{i-1}
    \right)\\
\vsigma_{i}^{-2} 
    &= (1-\alpha) \vsigma_{i-1}^{-2}
    + \alpha \, \diag 
    \left( \lambda \vI_\nparam + \vH_{i}^{\trans} \vR_{i}^{-1} \vH_{i} \right)
\end{align}

The batch \blr-\diagmom update becomes
\begin{align}
\vmu_{i} 
    &= \vmu_{i-1}
    + \alpha \vsigma_{i-1}^{2} \left( \vg_{i} - \lambda \vmu_{i-1} \right)\\
\vsigma_{i}^{2} 
    &= (1 + \alpha) \vsigma_{i-1}^{2}
    + \alpha \vsigma_{i-1}^{4} \diag \left( \vG_{i} - \lambda \vI_\nparam \right)
\end{align}

The batch \blr-\hesslin-\diagmom update becomes
\begin{align}
\vmu_{i} 
    &= \vmu_{i-1}
    + \alpha \vsigma_{i-1}^{2} 
    \left( 
        \vH_{i}^{\trans} \vR_{i}^{-1} (\vy_{t}-\hat{\vy}_{i}) 
        - \lambda \vmu_{i-1}
    \right)\\
\vsigma_{i}^{2} 
    &= (1 + \alpha) \vsigma_{i-1}^{2}
    - \alpha \vsigma_{i-1}^{4} 
    \diag \left( \lambda \vI_\nparam + \vH_{i}^{\trans} \vR_{i}^{-1} \vH_{i} \right)
\end{align}

The batch \blr-\dlr update becomes 
\begin{align}
\vmu_{i} 
    &= \vmu_{i-1}
    + \alpha \left(
        \tilde{\vUpsilon}_{i}^{-1} 
        -\tilde{\vUpsilon}_{i}^{-1} \tilde{\vW}_{i} 
        \left(
            \vI_{\rank + \nsample}
            + \tilde{\vW}_{i}^{\trans} \tilde{\vUpsilon}_{i}^{-1} \tilde{\vW}_{i}
        \right)^{-1}
        \tilde{\vW}_{i}^{\trans} \tilde{\vUpsilon}_{i}^{-1}
    \right)
    \nonumber \\
    & \qquad
    \times \left(
        \vg_t
        - \lambda
        \vmu_{i-1}
    \right)
    \label{eq:slang-mu}\\
\vW_{i} 
    &= \vU_{i}\left[:,{:\rank}\right] \vLambda_{i}\left[{:\rank},{:\rank}\right]\\
\vUpsilon_{i} 
    &= \tilde{\vUpsilon}_{i}
    + \diag \left(
        \tilde{\vW}_{i} \tilde{\vW}_{i}^{\trans}
        - \vW_{i} \vW_{i}^{\trans}
    \right)\\
\left( \vU_{i}, \vLambda_{i}, \_ \right) 
    &= {\rm SVD} \left( \tilde{\vW}_{i} \right)\\
\tilde{\vW}_{i} 
    &= \left[
        \sqrt{1-\alpha} \vW_{i-1},
        \sqrt{\frac{\alpha}{\nsample}} \gradmat_{i}
    \right] 
    \label{eq:slang-Wtilde}\\
\vA_{i} 
    &= {\rm chol} \left( \vR_{i}^{-1} \right) \\
\tilde{\vUpsilon}_{i} 
    &= (1-\alpha) \vUpsilon_{i-1}
    + \alpha \lambda \vI_\nparam
\end{align}
This is equivalent to \SLANG except for the following differences. \Slang processes a minibatch of $\nsample$ examples at each iteration, using a single sample $\hat{\vtheta} \sim q_{\vpsi_{i-1}}$ for each minibatch. It uses a different SVD routine which is slightly faster but stochastic, taken from \citep{halko2011finding}. 
Most significantly, \SLANG applies the SVD before the mean update, meaning $\vmu_i$ is calculated using the rank-$\rank$ $\vW_{i}$ and $\vUpsilon_i$ instead of the rank-$(\rank + \nsample)$ $\tilde{\vW}_{i}$ and $\tilde{\vUpsilon}_i$, thus ignoring the non-diagonal information in the $\nsample$ discarded singular vectors from $\tilde{\vW}_i$.

The batch \blr-\hesslin-\dlr update becomes
\begin{align}
\vmu_{i} 
    &= \vmu_{i-1}
    + \alpha \left(
        \tilde{\vUpsilon}_{i}^{-1} 
        -\tilde{\vUpsilon}_{i}^{-1} \tilde{\vW}_{i} 
        \left(
            \vI_{\rank + \nout}
            + \tilde{\vW}_{i}^{\trans} \tilde{\vUpsilon}_{i}^{-1} \tilde{\vW}_{i}
        \right)^{-1}
        \tilde{\vW}_{i}^{\trans} \tilde{\vUpsilon}_{i}^{-1}
    \right)
    \nonumber \\
    & \qquad
    \times \left(
        \vH_{i}^{\trans} \vR_{i}^{-1} (\vy_{t}-\hat{\vy}_{i})
        -\lambda \vmu_{i-1}
    \right)\\
\vW_{i} 
    &= \vU_{i}\left[:,{:\rank}\right] \vLambda_{i}\left[{:\rank},{:\rank}\right]\\
\vUpsilon_{i} 
    &= \tilde{\vUpsilon}_{i}
    + \diag \left(
        \tilde{\vW}_{i} \tilde{\vW}_{i}^{\trans}
        - \vW_{i} \vW_{i}^{\trans}
    \right)\\
\left( \vU_{i}, \vLambda_{i}, \_ \right) 
    &= {\rm SVD} \left( \tilde{\vW}_{i} \right)\\
\tilde{\vW}_{i} 
    &= \left[
        \sqrt{1-\alpha} \vW_{i-1},
        \sqrt{\alpha} \vH_{i}^{\trans} \vA_{i}^{\trans}
    \right]\\
\vA_{i} 
    &= {\rm chol} \left( \vR_{i}^{-1} \right) \\
\tilde{\vUpsilon}_{i} 
    &= (1-\alpha) \vUpsilon_{i-1}
    + \alpha \lambda \vI_\nparam
\end{align}
This algorithm could be called \SLANG-\hesslin and would be deterministic and faster than \SLANG since it does not need MC sampling.
We can also define \SLANG-\linef which would be even faster,
by replacing $\sqrt{\frac{\alpha}{\nsample}} \gradmat_{i}$
with $\sqrt{\alpha} \gL_{i}$ in \cref{eq:slang-Wtilde}
and $\vI_{\rank+\nsample}$ with $\vI_\rank$ in \cref{eq:slang-mu}.

%% file: bong-neurips24-final.bbl
\begin{thebibliography}{43}
\providecommand{\natexlab}[1]{#1}
\providecommand{\url}[1]{\texttt{#1}}
\expandafter\ifx\csname urlstyle\endcsname\relax
  \providecommand{\doi}[1]{doi: #1}\else
  \providecommand{\doi}{doi: \begingroup \urlstyle{rm}\Url}\fi

\bibitem[Amari(1998)]{Amari98}
S~Amari.
\newblock Natural gradient works efficiently in learning.
\newblock \emph{Neural Comput.}, 10\penalty0 (2):\penalty0 251--276, 1998.
\newblock URL \url{http://dx.doi.org/10.1162/089976698300017746}.

\bibitem[Bencomo et~al.(2023)Bencomo, Snell, and Griffiths]{bencomo2023implicit}
Gianluca~M Bencomo, Jake~C Snell, and Thomas~L Griffiths.
\newblock Implicit maximum a posteriori filtering via adaptive optimization.
\newblock \emph{arXiv preprint arXiv:2311.10580}, 2023.

\bibitem[Blundell et~al.(2015)Blundell, Cornebise, Kavukcuoglu, and Wierstra]{BBB}
Charles Blundell, Julien Cornebise, Koray Kavukcuoglu, and Daan Wierstra.
\newblock Weight uncertainty in neural networks.
\newblock In \emph{{ICML}}, 2015.
\newblock URL \url{http://arxiv.org/abs/1505.05424}.

\bibitem[Bonnet(1964)]{bonnet1964}
Georges Bonnet.
\newblock Transformations des signaux al{\'e}atoires a travers les systemes non lin{\'e}aires sans m{\'e}moire.
\newblock In \emph{Annales des T{\'e}l{\'e}communications}, volume~19, pages 203--220. Springer, 1964.

\bibitem[Chang et~al.(2022)Chang, Murphy, and Jones]{diaglofi}
Peter~G Chang, Kevin~Patrick Murphy, and Matt Jones.
\newblock On diagonal approximations to the extended kalman filter for online training of bayesian neural networks.
\newblock In \emph{Continual Lifelong Learning Workshop at {ACML} 2022}, December 2022.
\newblock URL \url{https://openreview.net/forum?id=asgeEt25kk}.

\bibitem[Chang et~al.(2023)Chang, Dur{\'a}n-Mart{\'\i}n, Shestopaloff, Jones, and Murphy]{lofi}
Peter~G Chang, Gerardo Dur{\'a}n-Mart{\'\i}n, Alexander~Y Shestopaloff, Matt Jones, and Kevin Murphy.
\newblock {Low-rank extended Kalman filtering for online learning of neural networks from streaming data}.
\newblock In \emph{{COLLAS}}, May 2023.
\newblock URL \url{http://arxiv.org/abs/2305.19535}.

\bibitem[Ch{\'e}rief-Abdellatif et~al.(2019)Ch{\'e}rief-Abdellatif, Alquier, and Khan]{cherief2019generalization}
Badr-Eddine Ch{\'e}rief-Abdellatif, Pierre Alquier, and Mohammad~Emtiyaz Khan.
\newblock A generalization bound for online variational inference.
\newblock In \emph{Asian conference on machine learning}, pages 662--677. PMLR, 2019.

\bibitem[Duran-Martin et~al.(2024)Duran-Martin, Altamirano, Shestopaloff, S{\'a}nchez-Betancourt, Knoblauch, Jones, Briol, and Murphy]{duran2024outlier}
Gerardo Duran-Martin, Matias Altamirano, Alexander~Y Shestopaloff, Leandro S{\'a}nchez-Betancourt, Jeremias Knoblauch, Matt Jones, Fran{\c{c}}ois-Xavier Briol, and Kevin Murphy.
\newblock Outlier-robust kalman filtering through generalised bayes.
\newblock \emph{arXiv preprint arXiv:2405.05646}, 2024.

\bibitem[Gama et~al.(2013)Gama, Sebasti{\~a}o, and Rodrigues]{Gama2013}
Jo{\~a}o Gama, Raquel Sebasti{\~a}o, and Pedro~Pereira Rodrigues.
\newblock On evaluating stream learning algorithms.
\newblock \emph{MLJ}, 90\penalty0 (3):\penalty0 317--346, March 2013.
\newblock URL \url{https://tinyurl.com/mrxfk4ww}.

\bibitem[Halko et~al.(2011)Halko, Martinsson, and Tropp]{halko2011finding}
Nathan Halko, Per-Gunnar Martinsson, and Joel~A Tropp.
\newblock Finding structure with randomness: Probabilistic algorithms for constructing approximate matrix decompositions.
\newblock \emph{SIAM review}, 53\penalty0 (2):\penalty0 217--288, 2011.

\bibitem[Hoeven et~al.(2018)Hoeven, Erven, and Kot{\l}owski]{hoeven2018many}
Dirk Hoeven, Tim Erven, and Wojciech Kot{\l}owski.
\newblock The many faces of exponential weights in online learning.
\newblock In \emph{Conference On Learning Theory}, pages 2067--2092. PMLR, 2018.

\bibitem[Hutchinson(1989)]{hutchinson1989stochastic}
Michael~F Hutchinson.
\newblock A stochastic estimator of the trace of the influence matrix for laplacian smoothing splines.
\newblock \emph{Communications in Statistics-Simulation and Computation}, 18\penalty0 (3):\penalty0 1059--1076, 1989.

\bibitem[Immer et~al.(2021{\natexlab{a}})Immer, Korzepa, and Bauer]{Immer2021linear}
Alexander Immer, Maciej Korzepa, and Matthias Bauer.
\newblock Improving predictions of bayesian neural nets via local linearization.
\newblock In Arindam Banerjee and Kenji Fukumizu, editors, \emph{AISTATS}, volume 130 of \emph{Proceedings of Machine Learning Research}, pages 703--711. PMLR, 2021{\natexlab{a}}.
\newblock URL \url{https://proceedings.mlr.press/v130/immer21a.html}.

\bibitem[Immer et~al.(2021{\natexlab{b}})Immer, Korzepa, and Bauer]{Immer21a}
Alexander Immer, Maciej Korzepa, and Matthias Bauer.
\newblock Improving predictions of bayesian neural nets via local linearization.
\newblock In \emph{Proceedings of The 24th International Conference on Artificial Intelligence and Statistics}, pages 703--711. PMLR, 2021{\natexlab{b}}.
\newblock URL \url{https://proceedings.mlr.press/v130/immer21a.html}.

\bibitem[Jordan et~al.(1999)Jordan, Ghahramani, Jaakkola, and Saul]{jordan1999introduction}
Michael~I Jordan, Zoubin Ghahramani, Tommi~S Jaakkola, and Lawrence~K Saul.
\newblock An introduction to variational methods for graphical models.
\newblock \emph{Machine learning}, 37:\penalty0 183--233, 1999.

\bibitem[Khan and Lin(2017)]{khan2017conjugate}
Mohammad Khan and Wu~Lin.
\newblock Conjugate-computation variational inference: Converting variational inference in non-conjugate models to inferences in conjugate models.
\newblock In \emph{Artificial Intelligence and Statistics}, pages 878--887. PMLR, 2017.

\bibitem[Khan et~al.(2018{\natexlab{a}})Khan, Nielsen, Tangkaratt, Lin, Gal, and Srivastava]{khan2018fast}
Mohammad Khan, Didrik Nielsen, Voot Tangkaratt, Wu~Lin, Yarin Gal, and Akash Srivastava.
\newblock Fast and scalable bayesian deep learning by weight-perturbation in adam.
\newblock In \emph{International conference on machine learning}, pages 2611--2620. PMLR, 2018{\natexlab{a}}.

\bibitem[Khan and Rue(2023)]{BLR}
Mohammad~Emtiyaz Khan and H{\aa}vard Rue.
\newblock The bayesian learning rule.
\newblock \emph{J. Mach. Learn. Res.}, 2023.
\newblock URL \url{http://arxiv.org/abs/2107.04562}.

\bibitem[Khan et~al.(2018{\natexlab{b}})Khan, Nielsen, Tangkaratt, Lin, Gal, and Srivastava]{VON}
Mohammad~Emtiyaz Khan, Didrik Nielsen, Voot Tangkaratt, Wu~Lin, Yarin Gal, and Akash Srivastava.
\newblock Fast and scalable bayesian deep learning by {Weight-Perturbation} in adam.
\newblock In \emph{{ICML}}, 2018{\natexlab{b}}.
\newblock URL \url{http://arxiv.org/abs/1806.04854}.

\bibitem[Knoblauch et~al.(2022)Knoblauch, Jewson, and Damoulas]{knoblauch2022optimization}
Jeremias Knoblauch, Jack Jewson, and Theodoros Damoulas.
\newblock An optimization-centric view on bayes' rule: Reviewing and generalizing variational inference.
\newblock \emph{Journal of Machine Learning Research}, 23\penalty0 (132):\penalty0 1--109, 2022.

\bibitem[Kunstner et~al.(2020)Kunstner, Balles, and Hennig]{kunstner2020limitations}
Frederik Kunstner, Lukas Balles, and Philipp Hennig.
\newblock Limitations of the empirical fisher approximation for natural gradient descent, 2020.

\bibitem[Kurle et~al.(2020)Kurle, Cseke, Klushyn, van~der Smagt, and Günnemann]{Kurle2020}
Richard Kurle, Botond Cseke, Alexej Klushyn, Patrick van~der Smagt, and Stephan Günnemann.
\newblock Continual learning with bayesian neural networks for non-stationary data.
\newblock In \emph{ICLR}, March 2020.
\newblock URL \url{https://openreview.net/forum?id=SJlsFpVtDB}.

\bibitem[Lambert et~al.(2021)Lambert, Bonnabel, and Bach]{RVGA}
Marc Lambert, Silv{\`e}re Bonnabel, and Francis Bach.
\newblock The recursive variational gaussian approximation ({R-VGA}).
\newblock \emph{Stat. Comput.}, 32\penalty0 (1):\penalty0 10, December 2021.
\newblock URL \url{https://hal.inria.fr/hal-03086627/document}.

\bibitem[Lambert et~al.(2023)Lambert, Bonnabel, and Bach]{LRVGA}
Marc Lambert, Silv{\`e}re Bonnabel, and Francis Bach.
\newblock The limited-memory recursive variational gaussian approximation (l-rvga).
\newblock \emph{Statistics and Computing}, 33\penalty0 (3):\penalty0 70, 2023.

\bibitem[LeCun et~al.(2010)LeCun, Cortes, and Burges]{lecun2010mnist}
Yann LeCun, Corinna Cortes, and CJ~Burges.
\newblock Mnist handwritten digit database.
\newblock \emph{ATT Labs [Online]. Available: http://yann.lecun.com/exdb/mnist}, 2, 2010.

\bibitem[Lin et~al.(2024)Lin, Dangel, Eschenhagen, Bae, Turner, and Makhzani]{lin2024sqrtfree}
Wu~Lin, Felix Dangel, Runa Eschenhagen, Juhan Bae, Richard~E Turner, and Alireza Makhzani.
\newblock Can we remove the square-root in adaptive gradient methods? a second-order perspective.
\newblock \emph{arXiv preprint arXiv:2402.03496}, 2024.

\bibitem[Littlestone and Warmuth(1994)]{littlestone1994weighted}
Nick Littlestone and Manfred~K Warmuth.
\newblock The weighted majority algorithm.
\newblock \emph{Information and computation}, 108\penalty0 (2):\penalty0 212--261, 1994.

\bibitem[Lyu and Tsang(2021)]{lyu2021black}
Yueming Lyu and Ivor~W Tsang.
\newblock Black-box optimizer with stochastic implicit natural gradient.
\newblock In \emph{Machine Learning and Knowledge Discovery in Databases. Research Track: European Conference, ECML PKDD 2021, Bilbao, Spain, September 13--17, 2021, Proceedings, Part III 21}, pages 217--232. Springer, 2021.

\bibitem[Martens(2020)]{martens2020new}
James Martens.
\newblock New insights and perspectives on the natural gradient method.
\newblock \emph{Journal of Machine Learning Research}, 21\penalty0 (146):\penalty0 1--76, 2020.

\bibitem[Mishkin et~al.(2018)Mishkin, Kunstner, Nielsen, Schmidt, and Khan]{SLANG}
Aaron Mishkin, Frederik Kunstner, Didrik Nielsen, Mark Schmidt, and Mohammad~Emtiyaz Khan.
\newblock {SLANG}: Fast structured covariance approximations for bayesian deep learning with natural gradient.
\newblock In \emph{{NIPS}}, pages 6245--6255. Curran Associates, Inc., 2018.

\bibitem[Ollivier(2018)]{Ollivier2018}
Yann Ollivier.
\newblock Online natural gradient as a kalman filter.
\newblock \emph{Electron. J. Stat.}, 12\penalty0 (2):\penalty0 2930--2961, 2018.
\newblock URL \url{https://projecteuclid.org/euclid.ejs/1537257630}.

\bibitem[Price(1958)]{price1958useful}
Robert Price.
\newblock A useful theorem for nonlinear devices having gaussian inputs.
\newblock \emph{IRE Transactions on Information Theory}, 4\penalty0 (2):\penalty0 69--72, 1958.

\bibitem[Puskorius and Feldkamp(1991)]{Puskorius1991}
G~V Puskorius and L~A Feldkamp.
\newblock Decoupled extended kalman filter training of feedforward layered networks.
\newblock In \emph{International Joint Conference on Neural Networks}, volume~i, pages 771--777 vol.1, 1991.
\newblock URL \url{http://dx.doi.org/10.1109/IJCNN.1991.155276}.

\bibitem[Rasmussen and Williams(2006)]{Rasmussen06}
Carl~Edward Rasmussen and Christopher K.~I. Williams.
\newblock \emph{Gaussian Processes for Machine Learning}.
\newblock MIT Press, 2006.

\bibitem[Sarkka and Svensson(2023)]{Sarkka23}
Simo Sarkka and Lennart Svensson.
\newblock \emph{{Bayesian Filtering and Smoothing (2nd edition)}}.
\newblock Cambridge University Press, 2023.

\bibitem[Shen et~al.(2024)Shen, Daheim, Cong, Nickl, Marconi, Bazan, Yokota, Gurevych, Cremers, Khan, and M{\"o}llenhoff]{shen2024IVON}
Yuesong Shen, Nico Daheim, Bai Cong, Peter Nickl, Gian~Maria Marconi, Clement Bazan, Rio Yokota, Iryna Gurevych, Daniel Cremers, Mohammad~Emtiyaz Khan, and Thomas M{\"o}llenhoff.
\newblock Variational learning is effective for large deep networks.
\newblock \emph{arXiv preprint arXiv:2402.17641}, 2024.

\bibitem[Singhal and Wu(1989)]{Singhal1988}
Sharad Singhal and Lance Wu.
\newblock Training multilayer perceptrons with the extended kalman algorithm.
\newblock In \emph{{NIPS}}, volume~1, 1989.

\bibitem[Titsias et~al.(2024)Titsias, Galashov, Rannen-Triki, Pascanu, Teh, and Bornschein]{titsias2023kalman}
Michalis~K Titsias, Alexandre Galashov, Amal Rannen-Triki, Razvan Pascanu, Yee~Whye Teh, and Jorg Bornschein.
\newblock Kalman filter for online classification of non-stationary data.
\newblock In \emph{ICLR}, 2024.

\bibitem[Tomczak et~al.(2020)Tomczak, Swaroop, and Turner]{Tomczak2020}
Marcin~B Tomczak, Siddharth Swaroop, and Richard~E Turner.
\newblock Efficient low rank gaussian variational inference for neural networks.
\newblock In \emph{{NIPS}}, 2020.
\newblock URL \url{https://proceedings.neurips.cc/paper/2020/file/310cc7ca5a76a446f85c1a0d641ba96d-Paper.pdf}.

\bibitem[Tronarp et~al.(2018)Tronarp, Garc{\'\i}a-Fern{\'a}ndez, and S{\"a}rkk{\"a}]{Tronarp2018}
Filip Tronarp, {\'A}ngel~F Garc{\'\i}a-Fern{\'a}ndez, and Simo S{\"a}rkk{\"a}.
\newblock Iterative filtering and smoothing in nonlinear and {Non-Gaussian} systems using conditional moments.
\newblock \emph{IEEE Signal Process. Lett.}, 25\penalty0 (3):\penalty0 408--412, 2018.
\newblock URL \url{https://acris.aalto.fi/ws/portalfiles/portal/17669270/cm_parapub.pdf}.

\bibitem[Yao et~al.(2021)Yao, Gholami, Shen, Mustafa, Keutzer, and Mahoney]{adahessian}
Zhewei Yao, Amir Gholami, Sheng Shen, Mustafa Mustafa, Kurt Keutzer, and Michael~W Mahoney.
\newblock {ADAHESSIAN}: An adaptive second order optimizer for machine learning.
\newblock In \emph{{AAAI}}, 2021.
\newblock URL \url{http://arxiv.org/abs/2006.00719}.

\bibitem[Zellner(1988)]{zellner1988optimal}
Arnold Zellner.
\newblock Optimal information processing and bayes's theorem.
\newblock \emph{The American Statistician}, 42\penalty0 (4):\penalty0 278--280, 1988.

\bibitem[Zhang et~al.(2024)Zhang, Mohamed, Ghanem, Torr, Bibi, and Elhoseiny]{Zhang2024CL}
Wenxuan Zhang, Youssef Mohamed, Bernard Ghanem, Philip Torr, Adel Bibi, and Mohamed Elhoseiny.
\newblock Continual learning on a diet: Learning from sparsely labeled streams under constrained computation.
\newblock In \emph{{ICLR}}, 2024.
\newblock URL \url{https://openreview.net/pdf?id=Xvfz8NHmCj}.

\end{thebibliography}
